\documentclass{article}

\usepackage[T1]{fontenc}
\usepackage{microtype}
\usepackage{graphicx}
\usepackage{booktabs}
\usepackage{array}
\usepackage{longtable}
\usepackage{hyperref}

\usepackage[preprint]{icml2027}
\hypersetup{pdfsubject={Machine learning preprint}}

\usepackage{amsmath,amssymb,amsthm,mathtools}
\usepackage[capitalize,noabbrev]{cleveref}

\theoremstyle{plain}
\newtheorem{theorem}{Theorem}[section]
\newtheorem{proposition}[theorem]{Proposition}
\newtheorem{lemma}[theorem]{Lemma}
\newtheorem{corollary}[theorem]{Corollary}
\newtheorem{conjecture}[theorem]{Conjecture}
\theoremstyle{definition}

\theoremstyle{remark}

\newcommand{\cX}{\mathcal{X}}
\newcommand{\cF}{\mathcal{F}}

\newcommand{\E}{\mathbb{E}}
\newcommand{\Prob}{\mathbb{P}}
\newcommand{\ind}{\mathbf{1}}
\newcommand{\eps}{\varepsilon}
\DeclareMathOperator{\proj}{proj}
\DeclareMathOperator{\TC}{TC}
\DeclareMathOperator{\KL}{KL}
\DeclareMathOperator{\TV}{TV}
\DeclareMathOperator{\cond}{cond}
\newcommand{\before}[1]{G_{<#1}}
\newcommand{\Sched}{S_{\nu}}
\newcommand{\posthoc}{\textsuperscript{\textdagger}}

\newcommand{\smp}[1]{\mathsf{#1}}

\newcommand{\codeplace}{the supplementary code}
\newcommand{\codeplaceSentence}{The supplementary material contains}
\newcommand{\rhoExampleLsixNtwo}{1, 4/5, 7/10, 8/15, 1/3, 0}
\newcommand{\llmBlockLength}{32}
\newcommand{\llmTokensPerStep}{4}

\newcommand{\llmPairFactor}{28/31}
\newcommand{\llmPairFactorDec}{0.903}

\newcommand{\linBoundRatioMax}{9.8\times10^{-4}}
\newcommand{\tcBoundRatioMax}{6.6\times10^{-3}}
\newcommand{\tcSBoundRatioMax}{1.0\times10^{-6}}

\newcommand{\entropyNegLMin}{14}
\newcommand{\entropyNegLMax}{24}
\newcommand{\entropyLargeAlphabetM}{5{,}000}

\newcommand{\deltaFactorExample}{6.6\times10^{20}}
\newcommand{\tcRiseRoundZero}{0.02702010328}
\newcommand{\tcRiseRoundOne}{0.02677695044}
\newcommand{\tcRiseRoundTwo}{0.0270535568}
\newcommand{\tcRiseChange}{2.766\times10^{-4}}
\newcommand{\ampTCBefore}{0.7327}
\newcommand{\ampTCAfter}{4.1589}

\newcommand{\extrapCoefLeight}{1/56}

\newcommand{\smRndTwoRFour}{0.6825}
\newcommand{\smRndTwoRFourSE}{0.0119}
\newcommand{\smRndTwoRFourPerf}{0.6815}
\newcommand{\smRndFourRFour}{0.2550}
\newcommand{\smRndFourRFourSE}{0.0120}
\newcommand{\smRndFourRFourPerf}{0.2776}
\newcommand{\smRndLamRFour}{0.7880}
\newcommand{\smRndLamRFourSE}{0.0019}
\newcommand{\smRndLamRFourPerf}{0.7862}
\newcommand{\smRandomMaxGap}{0.023}
\newcommand{\smFixROne}{0.4935}
\newcommand{\smFixRFour}{0.4925}
\newcommand{\smFixPerf}{0.4842}
\newcommand{\smDataTC}{0.390}
\newcommand{\smLcTCRFour}{2.5\times10^{-5}}
\newcommand{\smSeqCovDev}{0.027}
\newcommand{\smSeqMargDev}{0.014}
\newcommand{\smEosRndTwoEf}{6.503}
\newcommand{\smEosRndTwoEfPerf}{5.977}
\newcommand{\smEosRndThreeEf}{5.312}
\newcommand{\smEosRndThreeEfPerf}{4.776}

\newcommand{\cwConfRFour}{0.410}
\newcommand{\cwConfRFourSE}{0.033}
\newcommand{\cwConfPerf}{0.673}
\newcommand{\cwThrRFour}{0.307}
\newcommand{\cwThrRFourSE}{0.004}
\newcommand{\cwThrPerf}{0.446}
\newcommand{\clHalfPlateauLo}{0.762}
\newcommand{\clHalfPlateauHi}{0.768}
\newcommand{\clHalfPerf}{0.836}
\newcommand{\clFifthRTwelve}{0.529}
\newcommand{\clFifthRTwelveSE}{0.006}
\newcommand{\clFifthPerf}{0.738}
\newcommand{\clThrHalfRFour}{0.703}
\newcommand{\clThrHalfRFourSE}{0.006}
\newcommand{\clThrHalfPerf}{0.723}
\newcommand{\ampLcZeroBound}{10^{-8}}
\newcommand{\ampDataRare}{0.01005}

\newcommand{\ampFixTrigROne}{0.5001}

\newcommand{\ampRndTrigROne}{0.0299}
\newcommand{\ampRndTrigROneSE}{0.0026}
\newcommand{\ampRndTrigROnePerf}{0.0362}

\newcommand{\ampPairedTrig}{+0.0113}
\newcommand{\ampPairedTrigSE}{0.0109}
\newcommand{\ampDriftTrig}{+0.0360}

\newcommand{\ampThetaLowCount}{8}
\newcommand{\ampThetaLowMin}{0.0252}
\newcommand{\ampThetaLowMax}{0.2425}
\newcommand{\ampThetaHighMin}{0.5772}
\newcommand{\ampThetaHighMax}{0.5802}
\newcommand{\llRndEfBase}{199.4}
\newcommand{\llRndTailBase}{0.273}
\newcommand{\llRndPassBase}{0.135}
\newcommand{\llLcPassBase}{0.612}
\newcommand{\llLcBleuBase}{0.624}
\newcommand{\llDepBase}{9.99}
\newcommand{\llHWBase}{2.40}
\newcommand{\llRndEfFour}{47.0}
\newcommand{\llRndTailFour}{1.000}
\newcommand{\llRndPassFour}{0.030}
\newcommand{\llLcPassFour}{0.501}
\newcommand{\llLcBleuFour}{0.734}
\newcommand{\llDepFour}{2.55}
\newcommand{\llHWFour}{4.52}
\newcommand{\llDepDiff}{-7.44}
\newcommand{\llDepDiffSE}{1.06}
\newcommand{\dmRndEfBase}{169.6}
\newcommand{\dmRndTailBase}{0.452}
\newcommand{\dmRndPassBase}{0.058}
\newcommand{\dmLcPassBase}{0.415}
\newcommand{\dmLcBleuBase}{0.429}
\newcommand{\dmDepBase}{10.23}
\newcommand{\dmHWBase}{4.36}
\newcommand{\dmRndEfFour}{52.9}

\newcommand{\dmRndPassFour}{0.023}
\newcommand{\dmLcPassFour}{0.303}
\newcommand{\dmLcBleuFour}{0.624}
\newcommand{\dmDepFour}{3.10}
\newcommand{\dmHWFour}{5.40}

\newcommand{\llRndRepEfFour}{31.9}
\newcommand{\llRndRepDepFour}{3.53}
\newcommand{\llLcRepBleuFour}{0.722}
\newcommand{\llLcRepPassFour}{0.395}

\newcommand{\parViolCount}{5}

\newcommand{\parPreCount}{10}
\newcommand{\parFourCount}{3}

\newcommand{\parAllFourOTE}{-0.065}
\newcommand{\parAllFourOTECI}{[-0.121,\,-0.007]}
\newcommand{\parAllTwoNS}{-0.027}
\newcommand{\parAllTwoNSCI}{[-0.052,\,-0.002]}

\newcommand{\parDecompR}{+1.914}
\newcommand{\parDecompNA}{-0.872}
\newcommand{\parDecompC}{+2.907}
\newcommand{\parDecompD}{-0.002}
\newcommand{\parDecompNB}{-0.119}
\newcommand{\parDecompW}{-0.143}

\newcommand{\parDecompDCI}{[-0.180,\,+0.155]}
\newcommand{\parDecompDOne}{-0.092}
\newcommand{\parDecompDOneCI}{[-0.198,\,-0.007]}

\newcommand{\llEfLearn}{-61.7}
\newcommand{\llEfLearnCI}{[-75.6,\,-48.0]}
\newcommand{\llEfStepOne}{-36.8}
\newcommand{\llEfStepOneCI}{[-50.1,\,-22.1]}

\newcommand{\dmEfLearn}{-25.1}
\newcommand{\dmEfLearnCI}{[-38.9,\,-11.6]}
\newcommand{\dmEfStepOne}{-21.6}
\newcommand{\dmEfStepOneCI}{[-34.3,\,-8.8]}

\newcommand{\llmTrainPrompts}{2{,}000}
\newcommand{\llmEvalPrompts}{250}
\newcommand{\llmEvalSamples}{4}
\newcommand{\llmGenLength}{256}
\newcommand{\llmLoraRank}{16}
\newcommand{\llmLoraAlpha}{32}
\newcommand{\llmLoraSteps}{250}
\newcommand{\llmLoraBatch}{16}
\newcommand{\smLayers}{4}
\newcommand{\smWidth}{256}
\newcommand{\smStepCap}{40{,}000}
\newcommand{\clFifthREight}{0.547}
\newcommand{\parAnswersPerN}{1{,}000}
\newcommand{\parQLo}{50}
\newcommand{\parQHi}{249}
\newcommand{\speedQHi}{99}
\newcommand{\speedSamples}{5}
\newcommand{\poolQHi}{199}
\newcommand{\depStates}{50}
\newcommand{\depWindow}{8}

\newcommand{\dmSeqTailCount}{5}
\newcommand{\dmSeqTailOf}{200}

\newcommand{\lcFullWindowQuestions}{137}
\newcommand{\lcFullWindowBase}{0.807}
\newcommand{\lcFullWindowFour}{0.369}

\newcommand{\monoCounterPOne}{0.5000}
\newcommand{\monoCounterPTwo}{0.5625}

\newcommand{\lcMargEntStart}{0.6529}
\newcommand{\lcMargEntKOne}{0.0775}
\newcommand{\lcNearDetKOne}{7}
\newcommand{\lcMargEntKTwo}{0.1446}
\newcommand{\lcNearDetKTwo}{6}
\newcommand{\lcDepTCEighty}{0.002104}

\newcommand{\limitStressMaxTC}{2.2\times10^{-15}}
\newcommand{\limitStressStarts}{180}
\newcommand{\limitStressRises}{1}
\newcommand{\entropyExactLMin}{14}
\newcommand{\entropyExactLMax}{24}

\newcommand{\tcRiseMinEntry}{1.1\times10^{-58}}

\newcommand{\entropyLargeAlphabetS}{18.1}
\newcommand{\entropyLargeAlphabetExact}{-2.33\times10^{-5}, -5.81\times10^{-6}}
\newcommand{\entropyLargeAlphabetPred}{-2.32\times10^{-5}, -5.81\times10^{-6}}

\newcommand{\discUniform}{0.255}
\newcommand{\discOther}{0.0102}
\newcommand{\extrapClosedFormCases}{7}
\newcommand{\extrapLimitMaxDev}{1.8\times10^{-9}}

\newcommand{\smHeads}{4}
\newcommand{\smFF}{1{,}024}
\newcommand{\smBatch}{512}
\newcommand{\smPatience}{20}
\newcommand{\smValEvery}{500}
\newcommand{\smSixRndTwo}{0.6558}
\newcommand{\smSixRndTwoSE}{0.0008}
\newcommand{\smSixRndTwoPerf}{0.6772}
\newcommand{\smDistTwoRndTwo}{0.626}
\newcommand{\smDistTwoRndTwoSE}{0.007}
\newcommand{\smDistTwoRndTwoPerf}{0.557}
\newcommand{\smDistThreeRndTwo}{0.524}
\newcommand{\smDistThreeRndTwoSE}{0.048}
\newcommand{\smDistThreeRndTwoPerf}{0.428}
\newcommand{\smDistTwoRndFour}{0.228}
\newcommand{\smDistTwoRndFourSE}{0.029}
\newcommand{\smDistTwoRndFourPerf}{0.165}
\newcommand{\smDistThreeRndFour}{-0.032}
\newcommand{\smDistThreeRndFourSE}{0.022}
\newcommand{\smDistThreeRndFourPerf}{0.068}
\newcommand{\smFixDistTwoROne}{0.514}
\newcommand{\smFixDistTwoRFour}{0.371}
\newcommand{\smFixDistTwoRFourSE}{0.056}
\newcommand{\smFixDistTwoPerf}{0.520}
\newcommand{\smSixFixDistTwoROne}{0.559}
\newcommand{\smSixFixDistTwoRFour}{0.581}
\newcommand{\smDecompMaxBC}{0.0020}
\newcommand{\smDecompMaxBCSeed}{0.0039}
\newcommand{\smDecompALo}{-0.062}
\newcommand{\smDecompAHi}{-0.013}
\newcommand{\smEosRndTwoV}{0.435}
\newcommand{\smEosRndTwoVPerf}{0.472}
\newcommand{\smEosRndThreeV}{0.610}
\newcommand{\smEosRndThreeVPerf}{0.650}
\newcommand{\smEosTailRatioLo}{0.53}
\newcommand{\smEosTailRatioHi}{0.59}
\newcommand{\smEosLcSurvZero}{0.540}
\newcommand{\smEosLcSurvOne}{0.753}
\newcommand{\smEosLcSurvFour}{0.759}
\newcommand{\smEosLcSurvPerf}{0.750}
\newcommand{\smEosRMin}{0.171}
\newcommand{\ampOmegaSmall}{0.0192}
\newcommand{\ampOmegaLarge}{0.0247}
\newcommand{\cwConfDecA}{-0.0193}

\newcommand{\cwConfDecB}{-0.0117}
\newcommand{\cwConfDecBSE}{0.0016}
\newcommand{\cwConfDecBSel}{-0.0081}
\newcommand{\cwConfDecBSelSE}{0.0031}
\newcommand{\cwConfDecBVal}{-0.0036}
\newcommand{\cwConfDecBValSE}{0.0015}
\newcommand{\cwConfDecC}{-0.0038}
\newcommand{\cwConfDecCSE}{0.0009}
\newcommand{\cwThrDecA}{-0.0327}

\newcommand{\cwThrDecB}{+0.0011}
\newcommand{\cwThrDecBSE}{0.0003}

\newcommand{\cwThrDecC}{-0.0093}
\newcommand{\cwThrDecCSE}{0.0003}
\newcommand{\warmRndTwo}{1.42}
\newcommand{\warmRndTwoSE}{0.08}
\newcommand{\warmPeakRndTwo}{2.04}
\newcommand{\warmPeakRndTwoSE}{0.19}
\newcommand{\warmEndRndTwo}{1.11}
\newcommand{\warmEndRndTwoSE}{0.06}
\newcommand{\warmRndThree}{1.47}
\newcommand{\warmRndThreeSE}{0.16}
\newcommand{\warmPeakRndThree}{1.45}
\newcommand{\warmPeakRndThreeSE}{0.23}
\newcommand{\warmEndRndThree}{1.11}
\newcommand{\warmEndRndThreeSE}{0.06}
\newcommand{\warmPerLo}{1.22}
\newcommand{\warmPerHi}{1.78}
\newcommand{\warmScratchLo}{0.82}
\newcommand{\warmScratchHi}{1.10}
\newcommand{\llVocabRows}{126{,}464}
\newcommand{\llVocabTok}{126{,}349}
\newcommand{\llVocabPad}{115}
\newcommand{\dmVocabRows}{152{,}064}
\newcommand{\dmVocabTok}{151{,}667}
\newcommand{\dmVocabPad}{397}
\newcommand{\llPadScanSamples}{75{,}600}
\newcommand{\llPadScanHits}{18}
\newcommand{\llPadScanLcMain}{23{,}000}

\newcommand{\padOvershootNopad}{2.33}
\newcommand{\padOvershootNopadAlt}{2.96}
\newcommand{\padOvershootPadAlt}{2.58}
\newcommand{\padOvershootCkpt}{2.71}

\newcommand{\padDATF}{-0.382}
\newcommand{\padDATFCI}{[-0.430,\,-0.338]}

\newcommand{\padDBTF}{+0.082}
\newcommand{\padDBTFCI}{[+0.060,\,+0.104]}
\newcommand{\padShareLo}{0.13}
\newcommand{\padShareHi}{0.29}
\newcommand{\padStreamLo}{0.15}
\newcommand{\padStreamHi}{0.35}
\newcommand{\calBlockMOne}{1.65}
\newcommand{\calBlockMZero}{0.64}
\newcommand{\calTrainMOne}{3.04}
\newcommand{\calTrainMZero}{0.81}

\newcommand{\calBlockFourLo}{1.13}
\newcommand{\calBlockFourHi}{1.66}
\newcommand{\calTrainFourLo}{2.56}
\newcommand{\calTrainFourHi}{3.26}

\newcommand{\learnRHRatio}{8.6}
\newcommand{\lcFullWindowRep}{0.496}
\newcommand{\dmTestEf}{-116.647}
\newcommand{\dmTestEfCI}{[-122.387,\,-110.907]}
\newcommand{\dmTestTail}{+0.548}
\newcommand{\dmTestTailCI}{[+0.515,\,+0.581]}
\newcommand{\dmTestNtail}{+68.68}
\newcommand{\dmTestNtailCI}{[+65.31,\,+72.05]}
\newcommand{\dmTestBleu}{+0.195}
\newcommand{\dmTestBleuCI}{[+0.169,\,+0.220]}
\newcommand{\dmTestDep}{-7.13}
\newcommand{\dmTestDepCI}{[-8.60,\,-5.66]}

\newcommand{\dmLcTailBase}{0.099}
\newcommand{\dmLcTailFour}{0.284}

\newcommand{\dmOpenCells}{26}
\newcommand{\dmCompCells}{13}
\newcommand{\dmCompSame}{11}

\newcommand{\llRndRepDepDiff}{-6.46}
\newcommand{\llDepROneLo}{-1.44}
\newcommand{\llDepROneHi}{-0.66}
\newcommand{\clFifthDiff}{-0.018}
\newcommand{\clFifthDiffSE}{0.010}
\newcommand{\clFifthBalance}{0.52}
\newcommand{\ampBRndThree}{-0.0064}
\newcommand{\ampBRndThreeSE}{0.0031}
\newcommand{\ampBSeqThree}{+0.0167}
\newcommand{\ampBSeqThreeSE}{0.0050}

\newcommand{\llLcDepDiff}{-1.57}
\newcommand{\llLcDepDiffSE}{0.74}
\newcommand{\llLcFiltDepDiff}{-0.79}
\newcommand{\llLcFiltDepDiffSE}{0.76}
\newcommand{\llLcRepDepDiff}{-1.75}
\newcommand{\llLcRepDepDiffSE}{0.77}
\newcommand{\llBaseReevalDep}{+1.11}
\newcommand{\llBaseReevalDepSE}{0.61}

\newcommand{\cwPredConfLo}{0.61}
\newcommand{\cwPredConfHi}{0.65}
\newcommand{\cwPredThrLo}{0.40}
\newcommand{\cwPredThrHi}{0.43}
\newcommand{\gpuLLaDA}{149.56}
\newcommand{\gpuSmall}{57.28}
\newcommand{\gpuDream}{38.20}
\newcommand{\gpuTotal}{245.04}
\newcommand{\gpuLladaMain}{60.60}

\newcommand{\gpuLladaLcrep}{12.20}

\newcommand{\gpuLladaRndrep}{12.21}

\newcommand{\gpuLladaFixsched}{9.50}
\newcommand{\gpuLladaFixalt}{7.90}
\newcommand{\gpuLladaLam}{13.51}

\newcommand{\gpuDreamMain}{33.62}

\newcommand{\fxRefStep}{-45.9}

\newcommand{\fxStepMin}{-36.8}
\newcommand{\fxStepMax}{-29.3}

\newcommand{\fxSmallerCount}{1}
\newcommand{\fxReadouts}{4}

\newcommand{\fxLearnG}{-31.1}

\newcommand{\fxLearnAlt}{-33.1}

\newcommand{\fxLearnRnd}{-56.6}

\newcommand{\fxDataEfG}{215.4}

\newcommand{\fxDataEfAlt}{197.4}

\newcommand{\fxDataEfRnd}{200.6}

\newcommand{\fxDepG}{0.686}

\newcommand{\fxDepAlt}{0.593}

\newcommand{\fxStates}{100}
\newcommand{\fxTrainN}{2{,}000}
\newcommand{\fxQHi}{399}

\newcommand{\fxMatch}{+3.2}
\newcommand{\fxMatchCI}{[-3.3,\,+9.6]}
\newcommand{\fxMatchPerfect}{+36.2}
\newcommand{\fxCrossModel}{-31.6}

\newcommand{\fxSameDecoderDiff}{+2.3}
\newcommand{\fxSameDecoderDiffCI}{[-11.6,\,+16.7]}
\newcommand{\fxDecRndG}{+14.3}
\newcommand{\fxDecRndGCI}{[+5.5,\,+23.1]}

\newcommand{\efqLlLearn}{-51.6}
\newcommand{\efqLlLearnCI}{[-59.9,\,-43.0]}
\newcommand{\efqLlStep}{-48.3}
\newcommand{\efqLlStepCI}{[-55.4,\,-41.3]}

\newcommand{\efqDmLearn}{-24.4}
\newcommand{\efqDmLearnCI}{[-33.2,\,-15.5]}
\newcommand{\efqDmStep}{-26.1}
\newcommand{\efqDmStepCI}{[-33.4,\,-18.6]}

\newcommand{\dqQHi}{99}
\newcommand{\dqSamples}{5}
\newcommand{\flRndEfFour}{48.0}

\newcommand{\flRndPassFour}{0.217}
\newcommand{\flRndTruncFour}{0.016}
\newcommand{\llRndTruncBase}{0.129}
\newcommand{\llRndTruncFour}{0.024}
\newcommand{\flLcPassFour}{0.652}
\newcommand{\flLcBleuFour}{0.800}

\newcommand{\llRndContentBase}{210.3}

\newcommand{\llRndContentFour}{190.0}

\newcommand{\dmRndContentBase}{187.5}

\newcommand{\dmRndContentFour}{139.6}

\newcommand{\llRndRepContentFour}{126.0}

\newcommand{\dmRndTruncBase}{0.049}
\newcommand{\dmRndTruncFour}{0.021}

\newcommand{\foRndTwoRFour}{0.6830}
\newcommand{\foRndFourRFour}{0.2798}
\newcommand{\foRndLamRFour}{0.7895}
\newcommand{\foMaxGapWeak}{0.0033}
\newcommand{\foMaxGapStrong}{0.0135}
\newcommand{\lamLambda}{0.5}
\newcommand{\lamRealN}{1{,}000}
\newcommand{\lamRealEf}{235.5}
\newcommand{\lamRealTail}{0.007}
\newcommand{\lamRealAcc}{0.567}

\newcommand{\lamGenEfThree}{79.4}
\newcommand{\lamGenEfFour}{92.0}
\newcommand{\lamTrainEfFour}{163.7}
\newcommand{\lamTrainTailFour}{0.490}

\newcommand{\lamEfFour}{60.8}
\newcommand{\lamTailFour}{0.997}
\newcommand{\lamDepFour}{7.30}
\newcommand{\lamPassFour}{0.032}
\newcommand{\lamTruncFour}{0.018}

\newcommand{\lamHWFour}{3.36}
\newcommand{\lamEfDiffOrig}{+13.7}
\newcommand{\lamEfDiffOrigCI}{[+11.5,\,+16.0]}
\newcommand{\lamEfDiffRep}{+28.9}
\newcommand{\lamEfDiffRepCI}{[+26.5,\,+31.2]}
\newcommand{\lamTailDiff}{-0.003}
\newcommand{\lamTailDiffCI}{[-0.0064,\,+0.0004]}
\newcommand{\lamStepFour}{-29.7}
\newcommand{\lamStepFourCI}{[-33.5,\,-26.0]}
\newcommand{\lamStepThree}{+12.7}
\newcommand{\lamDepDiffFour}{+4.75}
\newcommand{\lamDepDiffFourSE}{0.59}

\newcommand{\lamHWDiffFour}{-1.16}
\newcommand{\lamHWDiffFourCI}{[-1.38,\,-0.93]}
\newcommand{\lamDepMinEarly}{8.83}
\newcommand{\lamDepMaxEarly}{9.24}
\newcommand{\lamRealQHi}{999}
\newcommand{\lamGenQLo}{1000}
\newcommand{\lamGenQHi}{1999}

\newcommand{\lamChainsGapFour}{15.1}
\newcommand{\lamEfDiffOrigTwo}{+0.4}
\newcommand{\lamEfDiffOrigTwoCI}{[-3.1,\,+3.8]}

\newcommand{\lamChainsDepGapFour}{+0.98}
\newcommand{\lamChainsDepGapFourSE}{0.27}

\icmltitlerunning{Sampler Drift and Learning Error in Self-Distilled Masked Diffusion Language Models}

\begin{document}

\twocolumn[
  \icmltitle{Sampler Drift and Learning Error in Self-Distilled\\
    Masked Diffusion Language Models}

  \begin{icmlauthorlist}
    \icmlauthor{Zhong Han}{nus}
  \end{icmlauthorlist}

  \icmlaffiliation{nus}{Department of Biomedical Engineering, College of Design and Engineering, National University of Singapore, Singapore}
  \icmlcorrespondingauthor{Zhong Han}{zhong.han.hz@u.nus.edu}

  \icmlkeywords{masked diffusion language models, self-distillation, parallel decoding, model collapse, total correlation}

  \vskip 0.3in
]

\printAffiliationsAndNotice{}

\begin{abstract}
Masked diffusion language models decode several tokens per step, sampling them independently given the decoded context, and a growing set of methods trains them on their own outputs. Such a loop drifts even when learning is perfect, because a parallel sampler $S$ does not reproduce the distribution it samples from; a self-distilled model may be degraded by its sampler or by its training, and its outputs do not show which. We study one round as the map $q\mapsto\lambda p+(1-\lambda)S(q)$, with a fraction $\lambda$ of real data $p$, and give tools that tell the two sources apart. Under perfect learning, schedules that do not look at the state multiply each interaction among positions, near independence, by an explicit factor per round, and real data set the fixed point; far from independence, one round can make a rare event frequent; with $n$ tokens per step and an end-token probability that ignores content, a round lowers the probability of no end token before a block end at most to its $n$-th power, a speed limit. In small models with exact conditionals, neighbouring dependence follows the map under such schedules and keeps drifting under deterministic confidence rules, where the map stops. In four rounds of self-distilling LLaDA-8B and Dream-7B with random remasking, the mean position of the first end token moves from $\llRndEfBase$ to $\llRndEfFour$ and from $\dmRndEfBase$ to $\dmRndEfFour$ while text continues after it; in the first round of both models, learning error moves it about as far as the trained model's parallel step, which stays within the speed limit; LLaDA's round exceeds the limit through learning error. With low-confidence remasking, the rule used in practice, answers grow more alike and, unless filtered, accuracy falls.
\end{abstract}

\section{Introduction}\label{sec:intro}

\begin{figure*}[t]
\centering
\includegraphics[width=\textwidth]{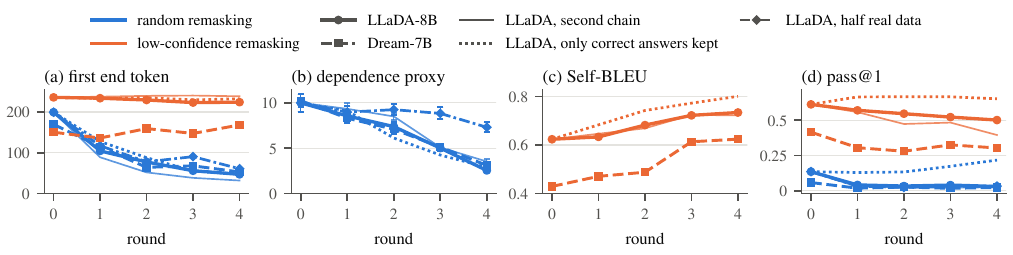}
\caption{Four rounds of self-distillation of two pretrained masked diffusion language models on their own GSM8K answers ($\llmTokensPerStep$ tokens per step in blocks of $\llmBlockLength$). (a) Mean position of the first end token. (b) Dependence proxy (estimated total correlation of $\depWindow$ positions) on fixed states (random chains; $\pm1$ standard error). (c) Self-BLEU on the text before the first end token (low-confidence chains). (d) pass@1, scored on the whole answer window. Half real data: \cref{sec:exp-llm}.}
\label{fig:rounds}
\end{figure*}

Masked diffusion language models \citep{austin2021structured,lou2023discrete,sahoo2024simple,shi2024simplified}, such as LLaDA \citep{nie2025llada} and Dream \citep{ye2025dream}, generate text by filling in masked positions, several per step. The positions filled in one step are sampled independently given what has been decoded, so the dependence among them is lost. This single-pass error is characterized through the total correlation among tokens and related quantities \citep{li2025breaking,chen2025optimal,zhao2026adaptation,cai2026confidence}, solved exactly for Markov data \citep{baule2026rsa}, decays in the worst case only like the inverse number of steps \citep[Thm.~1]{hayakawa2025di4c}, and degrades quality on real tasks \citep{kang2025parallelbench}.

A growing family of methods trains these models on their own outputs or trajectories, mostly to make few steps suffice: distilling two sampling steps into one over seven rounds \citep{deschenaux2025sdtt}; distilling a many-step teacher into a few-step student \citep{hayakawa2025di4c}; retraining on couplings generated by the previous model \citep{yoo2025redi}; fine-tuning LLaDA on its own correct answers \citep{chen2025dparallel} or on selected tokens of its own rollouts \citep{kim2026pivotsd}; and training on the teacher's decoding trajectories \citep{qian2026d3llm,zhang2026t3d}. Each applies one or a few rounds, but the next round can start where the last one ended: policy-gradient fine-tuning already samples its completions from the current model at every update \citep{zhao2025d1}, on-policy self-distillation conditions on the model's own answers \citep{luo2026dopsd}, and recursive self-distillation of autoregressive models has failure modes of its own \citep{abdelsadek2026progressive}.

For other generative models, training on generated data erodes the tails of the data and can end in collapse \citep{shumailov2023curse,alemohammad2023mad}; a retrained language model keeps producing correct outputs while their diversity collapses \citep{briesch2023selfconsuming}; real data that are mixed in or accumulated can keep the loop stable \citep{bertrand2024stability,gerstgrasser2024accumulate}; and curation turns the loop into an optimizer of the curator's preferences \citep{ferbach2024curated}. In most of these analyses the drift comes from learning (finite samples, misspecification).

Masked diffusion models have a second source: with exact learning and unlimited data, a parallel sampler still does not reproduce the distribution it samples from. Termination is our main readout: given a long answer window, instruction-tuned LLaDA and Dream can end an answer after a few tokens, traced to end tokens doubling as padding \citep{kim2025rainbow,liu2026voidpadding}.

This paper studies where the loop leads (\cref{fig:rounds}) and how to tell its two sources of drift apart. The sampler's part is fixed by the decoding rule and the data; it can be computed or bounded, and it remains when the model fits its training data exactly. The rest depends on the training recipe. Neither a degraded model nor the usual control, decoding one token per step (\cref{sec:split}), shows which part is at work. Our reference is \emph{perfect learning}, where the next model's conditionals are those of its training distribution and one round is the map $q\mapsto\lambda p+(1-\lambda)S(q)$ on distributions over sequences, with a fraction $\lambda$ of real data $p$. The drift of this map is the sampler's; what a trained chain does beyond it is learning error. Our contributions:
\begin{itemize}\setlength{\itemsep}{1pt}\setlength{\parsep}{0pt}\setlength{\topsep}{2pt}
\item \emph{The sampler under perfect learning} (\cref{sec:sampler}). Near independence, a schedule that does not look at the state multiplies each interaction among positions by an explicit factor per round, and real data set the fixed point. Away from it, one round can make a rare event frequent. We prove a \emph{speed limit} for termination: with at most $n$ tokens per step and an end-token hazard that ignores content, one round multiplies $-\log\Prob(\text{no end token before a block end})$ by at most $n$.
\item \emph{Telling the sampler from training} (\cref{sec:split}). An exact three-term split of a learned chain shows that the same-seed comparison with a sequential control can hide most of the drift (\cref{fig:small}(c)); within one round, sequential samples of the trained model separate its learning error from its parallel step.
\item \emph{Small models with exact conditionals} (\cref{sec:exp-small}; \cref{tab:scorecard}). Dependence between neighbouring positions follows perfect learning under state-independent schedules and keeps drifting under confidence rules.
\item \emph{A controlled study of two pretrained models on one task} (\cref{sec:exp-llm,sec:exp-split}). Four rounds of self-distillation with random remasking move the first end token of LLaDA-8B and Dream-7B far toward the start while text continues after it, for LLaDA also when only correct answers are kept. In the first round, learning error moves it about as far as the parallel step of the trained model; that step stays within the speed limit in both models, and the excess of LLaDA's round is learning error. Trained for one round on data decoded with a fixed grouping, LLaDA keeps a parallel step that perfect learning would remove. Half real data keep most of LLaDA's dependence proxy but do not stop its end token. Under low-confidence remasking answers grow more alike and, unless filtered, accuracy falls.
\end{itemize}

\section{Setup}\label{sec:setup}

\paragraph{Parallel decoding.}
Let $x=(x_1,\dots,x_L)\in\cX^L$ with $|\cX|=m$, and $x_A=(x_i)_{i\in A}$.
At a state with decoded positions $D$ and values $x_D$, a masked diffusion language model returns a distribution for every masked position; in the theory these are the conditionals $q(x_i\mid x_D)$ of one joint distribution $q$.
A \emph{schedule} is an ordered partition $G=(G_1,\dots,G_K)$ of the positions, with $\before{k}=G_1\cup\dots\cup G_{k-1}$.
Decoding with $G$ samples, at step $k$, every position of $G_k$ independently given the values decoded so far, so the output has the distribution
\begin{equation}\label{eq:projection}
  \proj_G(q)(x)=\prod_{k=1}^{K}\prod_{i\in G_k} q\bigl(x_i\mid x_{\before{k}}\bigr).
\end{equation}
With singleton groups $\proj_G(q)=q$: sequential decoding in any order reproduces $q$.
A step with several positions replaces their joint conditional by the product of its marginals: two perfectly correlated bits decoded in one step come out independent.

\paragraph{Samplers.}
(i) \emph{State-independent schedules} $\Sched(q)=\E_{G\sim\nu}\proj_G(q)$, for a law $\nu$ that does not depend on $q$: fixed partitions, random remasking with $n$ tokens per step (uniform over ordered partitions into groups of size $n$), and the \emph{block sampler} of our language-model experiments, which decodes blocks of $\ell$ positions from left to right \citep[App.~B.4]{nie2025llada} and $n$ uniformly random positions per step inside the block \citep{llada2025code}.
(ii) \emph{Adaptive rules}, which choose the next group from the current conditionals but not from the values about to be sampled: the $k$ positions of highest \emph{confidence} $\max_aq(a\mid x_D)$, all positions above a confidence threshold \citep{wu2025fastdllm}, or an entropy budget \citep[Alg.~1]{cai2026confidence}.
(iii) \emph{Low-confidence remasking} \citep[Sec.~2.4 and Alg.~5]{nie2025llada}, in the sampled form of the official code \citep{llada2025code}, which looks at the sampled values: it draws $v_i\sim q(\cdot\mid x_D)$ at every masked position and commits the $k$ positions with the largest $q(v_i\mid x_D)$. \Cref{app:notation} lists the notation and what is proved for each class.

\paragraph{Perfect learning.}
In round $t$ the current model generates a training set with its sampler $S$, and the next model is trained on it, mixed with a fraction $\lambda\in[0,1)$ of data from the data distribution $p$.
Under \emph{perfect learning} the conditionals of the next model are those of its training distribution, so from $q_0=p$,
\begin{equation}\label{eq:loop}
  q_{t+1}=\lambda\,p+(1-\lambda)\,S(q_t).
\end{equation}
A learned model returns a conditional table $Q$ that need not be the table $\cond(q)$ of any joint distribution; $S[Q]$ denotes the sampler run with $Q$, so $S(q)=S[\cond(q)]$.

\paragraph{A fixed schedule stops after one round.}
For fixed $G$, the map $\proj_G$ is a projection: it is idempotent, and its fixed points are the distributions under which the positions of each group are conditionally independent given the earlier groups.
In the loop \eqref{eq:loop} a fixed schedule therefore acts only once: $q_t=\lambda p+(1-\lambda)\proj_G(p)$ for every $t\ge1$ (\cref{prop:fixed-schedule}).
Dynamics over several rounds need another source: a schedule drawn afresh in every round, a rule that looks at the sampled values or is applied to a learned table (\cref{sec:sampler}), or training itself (\cref{sec:split}).

\section{What the sampler does under perfect learning}\label{sec:sampler}

\subsection{Contraction toward independence}\label{sec:contraction}

Fix a full-support product distribution $\mu=\bigotimes_i\mu_i$ on $\cX^L$ and write a nearby distribution as $q=\mu\,(1+\eps u)$ with $\E_\mu u=0$ and $U=\|u\|_\infty$.
We use the ANOVA (Hoeffding) decomposition of $u$ under $\mu$: with $u^V=\E_\mu[u\mid x_V]$ and $u_B=\sum_{C\subseteq B}(-1)^{|B\setminus C|}u^C$, one has $u^V=\sum_{B\subseteq V}u_B$.
Components with $|B|=1$ move marginals; components with $|B|\ge2$ are interactions.
For $B\neq\emptyset$ and an ordered partition $G$, let $E_B(G)$ be the event that the last group of $G$ that meets $B$ contains exactly one element of $B$, and let
$\rho_B(\nu)=\Prob_{G\sim\nu}\bigl(E_B(G)\bigr)$.
For a pair, $\rho_{\{i,j\}}=1-\pi_{ij}$, where $\pi_{ij}$ is the probability that $i$ and $j$ are decoded in the same step.

\begin{theorem}[Linearization]\label{thm:linearization}
Let $\nu$ be a state-independent schedule law and $|\eps|\le 1/(4LU)$. Then
\[
  \Sched(q)=\mu\Bigl(1+\eps\sum_{B\neq\emptyset}\rho_B(\nu)\,u_B+R\Bigr),
\]
where $\|R\|_\infty\le 8L^2U^2\eps^2$.
\end{theorem}

To first order, one round is diagonal in the ANOVA basis: the interaction on $B$ is multiplied by $\rho_B$, and marginals ($\rho_{\{i\}}=1$) move only at second order (proofs: \cref{app:proofs-contraction}).
For a single pass, this first-order statement, the formula for $\rho_k$ below and its pair value under block decoding were obtained independently in concurrent work \citep{li2026retention}, where $\rho_B$ is the retention of the interaction.
For random remasking, pairs survive longest.

\begin{lemma}[Pairs are the slowest]\label{lem:pairs-slowest}
Let $\nu$ be uniform over ordered partitions of $[L]$ into groups of size $n\ge2$, with $L=Kn$. Then $\rho_B$ depends only on $k=|B|$, $\rho_k=n\sum_{b=1}^{K}\binom{(b-1)n}{k-1}\big/\binom{L}{k}$, and $\rho_{k+1}\le\rho_k$ for $k\ge1$. Hence $\max_{|B|\ge2}\rho_B=\rho_2=1-\pi$ with $\pi=(n-1)/(L-1)$.
\end{lemma}

For the block sampler, $\rho_B$ is the lemma's $\rho_k$ with the block length $\ell$ in place of $L$ and $k$ the number of positions of $B$ in the last block that meets $B$ ($k=1$: not damped). In our language-model chains ($\ell=\llmBlockLength$, $n=\llmTokensPerStep$) the pair factor inside a block is $\llmPairFactor\approx\llmPairFactorDec$; far from independence it indicates only the direction of the drift.

\begin{theorem}[Local contraction]\label{thm:local-contraction}
Let $\TC(q)=\sum_iH(q_i)-H(q)$ be the total correlation in nats. Under the assumptions of \cref{thm:linearization} with $|\eps|\le1/(8LU)$,
\[
  \TC\bigl(\Sched(q)\bigr)\le\rho_*^2\,\TC(q)+56\,L^4U^3|\eps|^3,
\]
where $\rho_*=\max_{|B|\ge2}\rho_B(\nu)$.
\end{theorem}

\begin{corollary}[First-order fixed point with real data]\label{cor:lambda-fixed-point}
Let $p=\mu(1+\eps u)$ and run \eqref{eq:loop} with a state-independent $\nu$ and $\lambda\in(0,1)$. For every $t$, $q_t=\mu\bigl(1+\eps\sum_Bu^{(t)}_B+O_t(\eps^2)\bigr)$, where $u^{(0)}=u$ and $u^{(t+1)}_B=\lambda u_B+(1-\lambda)\rho_B\,u^{(t)}_B$. As $t\to\infty$ these coefficients converge to $u^{*}_B=\lambda\,u_B/\bigl(\lambda+(1-\lambda)(1-\rho_B)\bigr)$; for a pair under random remasking the factor is $\lambda/(\lambda+\pi(1-\lambda))$.
\end{corollary}

Without real data and with $\rho_*<1$, an iteration that starts close enough to a full-support product distribution converges to one (\cref{thm:local-convergence}; the neighbourhood is tiny).
Neither the total correlation (far from independence) nor the entropy (even near it) is monotone along the loop (\cref{app:beyond-neighbourhood}), and that $\TC(q_t)\to0$ for such laws is a conjecture.

\subsection{Rules that look at the state or at the sampled values}\label{sec:lowconf}

\paragraph{Adaptive rules freeze.}
At a product distribution $\mu$ the conditionals do not depend on the decoded values, so an adaptive rule reduces to a state-independent law $\nu_\mu$, its \emph{frozen schedule}, and near $\mu$ a rule that is Lipschitz or locally constant in the conditionals acts as $\nu_\mu$ to first order (\cref{thm:frozen}).
Top-$k$ and threshold rules are a single partition at generic $\mu$ (no ties, no confidence at the threshold), so they stop after one round near $\mu$ (\cref{sec:setup}; \cref{cor:deterministic-rules}).

\paragraph{Learned confidence rules ratchet.}
On a learned chain the rule is applied to a learned table, so the partition $G_t$ of round $t$ can change near ties.
To first order, round $t$ multiplies the interaction on $B$ by $\ind[E_B(G_t)]$ and adds the error of that round's sampled values (\cref{prop:ratchet}), so dependence removed by one round's partition does not come back, and the set of removed interactions, reached at once under perfect learning, can keep growing.

\paragraph{Low-confidence remasking sharpens.}
With $k<L$ commits per step this rule maps each marginal $q_i$ of a product distribution to $q'_i(v)=q_i(v)\,W_i(q_i(v))$ with $W_i$ nondecreasing, so $q'_i$ majorizes $q_i$: the mode probability does not decrease and the entropy does not increase (\cref{thm:lc-reweighting}; for one commit per step this is the majorization step of \citealp[Thm.~1]{zhang2026differences}, and \citealp{jeon2026lcr} treat identical positions).
The rule can also create dependence between independent positions; we conjecture that its iteration ends in a product with at most $k$ uncertain positions.

\subsection{Rare contexts and extrapolation}\label{sec:rare}

Near independence, state-independent schedules and adaptive rules at generic points only remove dependence, to first order.
Far from it, a parallel step can create dependence: two positions decoded together land in a combination the data almost never produce, and everything decoded afterwards is conditioned on that rare context.
Consider the \emph{copy-plus-rare-trigger} family $C_{L,m}(\eta)$, $0<\eta<1$: $x_1$ is uniform on $m$ values; with probability $1-\eta$, $x_2=x_1$, and otherwise $x_2$ is uniform on the other $m-1$ values; each of the $T=L-2$ trigger positions equals $\ind[x_1\neq x_2]$.

\begin{theorem}[Amplification]\label{thm:amplification}
Let $p=C_{L,m}(\eta)$ and let $h$ be the binary entropy.
\begin{enumerate}
\item[(a)] For every fixed schedule whose first group is $\{1,2\}$, $S(p)$ makes $x_1,x_2$ independent and uniform and sets every trigger to $\ind[x_1\neq x_2]$, whatever the convention at contexts of probability zero. The probability of $x_1\neq x_2$ and every trigger marginal jump from $\eta$ to $1-1/m$, and for $m=2$ the total correlation rises by $(T-1)\bigl(\ln2-h(\eta)\bigr)>0$ when $L\ge4$ and $\eta\neq\frac12$.
\item[(b)] For random remasking with $n\ge2$ tokens per step, $\Prob_{S(p)}(x_1\neq x_2)\ge\pi(1-1/m)(1-\eta)^{L-2}$ with $\pi=(n-1)/(L-1)$; for $n=2$ every trigger marginal is at least $(1-1/m)\cdot2/(L(L-1))$, whatever the convention.
\end{enumerate}
\end{theorem}

What happens after a rare combination depends on conditionals at contexts that $p$ never produces. Perfect learning does not determine them (\cref{prop:discontinuity}); a trained network fills them by extrapolation.
For state-independent schedules the extrapolation enters each binary marginal only through a visit-weighted average $\bar\theta$ of the learned table over those contexts (\cref{thm:structure}). For the amplification family the limit $\eta\to0$ is explicit (\cref{thm:extrapolation}): $\Prob(x_1\neq x_2)$ does not depend on the extrapolation, while the mean trigger marginal tends to $a+b\,\bar\theta$, with $a=1/(L(L-1))$ and $b=(L-4)/(4L(L-1))$ for random pairs ($n=m=2$).

\subsection{Termination: a speed limit}\label{sec:termination}

Language models end an answer with an end token (EOS or EOT; we write E). When the base LLaDA model decodes one token at a time, $\lamRealTail$ of its answers to $\lamRealN$ training questions have content after the first E: the data are close to \emph{monotone}, once E always E.
Parallel decoding can break this structure, because a position decoded early does not see the E that a later decision places before it.
In this subsection positions are indexed $0,\dots,L-1$. Let $e_j=1$ if position $j$ is E and $0$ if it is content, $f=\min\{j:e_j=1\}$ the position of the first E ($f=L$ if there is none), $n_E$ the number of E, and $n_{\mathrm{tail}}$ the number of content positions after the first E, so that $f+n_E+n_{\mathrm{tail}}=L$.
The \emph{E hazard ignores content} if, in every context (also of probability zero), the conditional law of $e_j$ depends only on which decoded positions are E and which are content, not on the content tokens; a binary model always does.

\begin{theorem}[Speed limit]\label{thm:speed-limit}
Let $D$ be a set of positions that are decoded before all others (for the block sampler, the positions $0,\dots,e-1$ before a block end $e$). Suppose every group has at most $n$ positions, groups are chosen without looking at the values about to be sampled, and either $q$ is binary or its E hazard ignores content. Then
\[
  \Prob_{S(q)}(\text{no E in }D)\ \ge\ \Prob_q(\text{no E in }D)^{\,n}.
\]
\end{theorem}

Hence one round of perfect learning without real data multiplies $-\log\Prob(f\ge e)$ at a block end by at most $n$, as long as the current distribution satisfies the hypothesis.
For a fixed partition the argument is short. Let $g_k$ be the probability under $q$ that the group $G_k$ is all content given that the earlier groups are; the $g_k$ multiply to $\Prob_q(\text{no E in }D)$. The sampler draws the positions of $G_k$ independently, each content with probability at least $g_k$, so the group comes out all content with probability at least $g_k^{|G_k|}\ge g_k^n$ (\cref{app:proofs-termination} treats random and adaptive groups).

The condition on the hazard cannot be dropped: if positions $0$ and $1$ each hold one of two content tokens, uniformly, and differ with small probability $\eta$, and position $2$ is E exactly when they differ, decoding $\{0,1\}$ together and then $\{2\}$ gives $\Prob(\text{no E})=\frac12<(1-\eta)^2$, the amplification of \cref{sec:rare} with E as the trigger.

The limit applies prompt by prompt. Let $P_{1,z}(e)$ be the probability of $f\ge e$ under sequential decoding for prompt $z$, and $P_n(e)$ that under the block sampler with $n$ random positions per step for a random prompt. If for every prompt the model's conditionals are those of one joint distribution satisfying the hypothesis, then
\begin{equation}\label{eq:per-question}
  P_n(e)\ge\E_z\bigl[P_{1,z}(e)^n\bigr]\ge\bigl(\E_z P_{1,z}(e)\bigr)^n .
\end{equation}
Middle: \emph{per-question form}; right: weaker \emph{pooled form}.

\section{Telling the sampler from training}\label{sec:split}

\begin{figure*}[t]
\centering
\includegraphics[width=\textwidth]{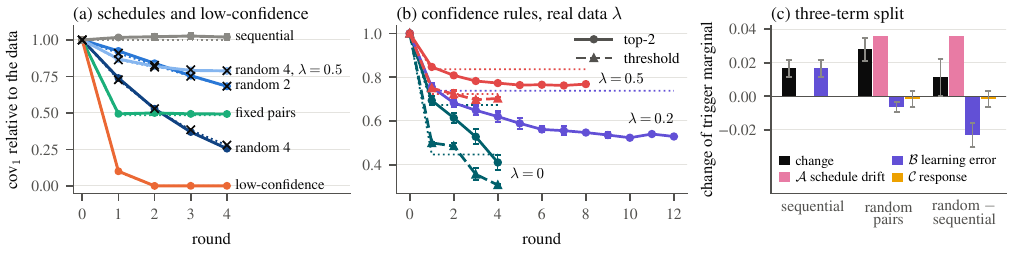}
\caption{Small models with exact conditionals (mean $\pm$ one standard error; three seeds, five in (c)). (a), (b) Nearest-neighbour covariance cov$_1$ of the Ising chain ($L=12$, $J=0.3$) relative to the data: learned chains and perfect learning (dotted, exact). (a) State-independent schedules (random $n$: $n$ tokens per step), low-confidence remasking (two per step) and sequential decoding; crosses: the first-order law $c_{t+1}=\lambda+(1-\lambda)(1-\pi)c_t$ of \cref{sec:contraction}. (b) Top-$2$ by confidence and the threshold rule ($\tau=0.662$) with real-data fraction $\lambda$. (c)\posthoc{} Copy-plus-rare-trigger family ($\eta=10^{-2}$): change of the trigger marginal after three rounds, split as in \eqref{eq:three-term}; right: random minus sequential arm, same seeds.}
\label{fig:small}
\end{figure*}

\paragraph{A three-term split.}
Let $\Phi(q)=\lambda p+(1-\lambda)S[\cond(q)]$ be one round of perfect learning, and let a learned chain follow $q_{s+1}=\lambda p+(1-\lambda)S[Q_s]$, where $Q_s$ is the table learned from $q_s$. For any functional $\psi$ (a marginal, a covariance, the total correlation),
\begin{equation}\label{eq:three-term}
  \psi(q_t)-\psi(q_0)=\mathcal A_t+\mathcal B_t+\mathcal C_t ,
\end{equation}
with the \emph{schedule drift} (the sampler's drift under perfect learning) $\mathcal A_t=\psi(\Phi^tq_0)-\psi(q_0)$, the \emph{learning error} $\mathcal B_t=\sum_{s<t}\bigl[\psi(q_{s+1})-\psi(\Phi q_s)\bigr]$, each term measured in the round where it is injected, and the \emph{response} $\mathcal C_t$ of the later rounds to the injected errors, which is the remainder (\cref{app:proofs-rare}).
Without real data, sequential decoding has $\mathcal A_t=\mathcal C_t=0$: all drift of a sequential control is learning error.
Hence the common comparison of a parallel arm with a sequential arm trained with the same seed measures
$\mathcal A^{\nu}_t+\bigl(\mathcal B^{\nu}_t-\mathcal B^{\mathrm{seq}}_t\bigr)+\mathcal C^{\nu}_t$,
not the schedule drift $\mathcal A^\nu_t$, and can hide most of it when the two learning errors have opposite signs (\cref{fig:small}(c)).

\paragraph{One round of a pretrained model.}
If a model's conditionals are consistent, decoding one token at a time samples its own joint distribution. Learned conditionals are not exactly consistent \citep{young2022inconsistencies,asadov2026marginal}; our reference is the model's own sampler with one token per step.
Let $\smp{D}_0$ be the training data of a round (parallel samples of the current model $m_0$), $m_1$ the model trained on them, $\smp{Q}_1$ its sequential and $\smp{D}_1$ its parallel samples. For every readout, $\smp{D}_1-\smp{D}_0=(\smp{Q}_1-\smp{D}_0)+(\smp{D}_1-\smp{Q}_1)$: the \emph{learning error}, read before the sampler acts (unlike $\mathcal B_t$), plus the \emph{parallel step} of $m_1$.
Under perfect learning the first term vanishes, and the second is constrained by \cref{sec:termination}: with $P_X(e)$ the probability of $f\ge e$ in the sample $X$, the log margin $\log P_{\smp{D}_1}-n\log P_{\smp{Q}_1}$ of the parallel step is nonnegative under the hypotheses of \cref{thm:speed-limit}. The margin of the whole round, $\log P_{\smp{D}_1}-n\log P_{\smp{D}_0}$, is this margin plus the learning-error term $n(\log P_{\smp{Q}_1}-\log P_{\smp{D}_0})$; the per-question form replaces each $P_X^n$ by $\E_z[P_{X,z}^n]$.

\paragraph{A projection test.}
If the training data are decoded with a fixed schedule $G$, then under perfect learning the trained model is a fixed point of $\proj_G$ (\cref{sec:setup}): decoding it with $G$ in parallel reproduces its sequential samples, and positions decoded together are conditionally independent given earlier groups.

\section{Experiments}\label{sec:experiments}

\Cref{tab:scorecard} sets the predictions of \cref{sec:sampler}, except the amplification of \cref{sec:rare}, against trained models. For small models with exact conditionals the map is a quantitative prediction; for the two pretrained models it is a reference: what a round does beyond the sampler's step is learning error, and the speed limit bounds that step.
\Cref{app:prespecified} lists the readouts and decision rules fixed before the data were seen; the main later analyses are marked~\posthoc.

\begin{table*}[t]
\caption{Predictions of the perfect-learning map against trained models. Small models, unless stated: nearest-neighbour covariance relative to the data after four rounds ($J=0.3$), learned against perfect learning. Pretrained: LLaDA, then Dream; chains after four rounds.}
\label{tab:scorecard}
\centering
\footnotesize
\begin{tabular}{@{}>{\raggedright\arraybackslash}p{0.28\textwidth}>{\raggedright\arraybackslash}p{0.32\textwidth}>{\raggedright\arraybackslash}p{0.345\textwidth}@{}}
\toprule
Under perfect learning & Small models with exact conditionals (\cref{sec:exp-small}, \cref{app:small-models}) & LLaDA-8B and Dream-7B (\cref{sec:exp-llm,sec:exp-split}) \\
\midrule
Random remasking: neighbouring dependence times $1-\pi$ per round (\cref{lem:pairs-slowest}) & followed: $\smRndTwoRFour\pm\smRndTwoRFourSE$ against $\smRndTwoRFourPerf$ ($n=2$), $\smRndFourRFour\pm\smRndFourRFourSE$ against $\smRndFourRFourPerf$ ($n=4$) & direction: the dependence proxy falls, $\llDepBase\to\llDepFour$ and $\dmDepBase\to\dmDepFour$ \\\addlinespace[2pt]
Real data set the fixed point (\cref{cor:lambda-fixed-point}) & followed: $\smRndLamRFour\pm\smRndLamRFourSE$ against $\smRndLamRFourPerf$ at $\lambda=0.5$, $n=4$ & direction: proxy $\lamDepFour$ with half real data, $\llDepFour$ and $\llRndRepDepFour$ without\posthoc{} (LLaDA) \\\addlinespace[2pt]
A fixed schedule acts once (\cref{sec:setup,sec:split}) & followed for neighbours: $\smFixROne$ and $\smFixRFour$ in rounds 1 and 4, against $\smFixPerf$ & not met: a parallel step of $\fxStepMin$ to $\fxStepMax$ tokens remains (LLaDA) \\\addlinespace[2pt]
Deterministic confidence rules stop after one round (\cref{sec:lowconf}) & not met: $\cwConfRFour$ against $\cwConfPerf$ under top-$2$ & not tested \\\addlinespace[2pt]
Low-confidence remasking sharpens every position (\cref{sec:lowconf}) & followed (\cref{fig:small}(a)): total correlation $\smDataTC\to\smLcTCRFour$ & direction: Self-BLEU rises, $\llLcBleuBase\to\llLcBleuFour$ and $\dmLcBleuBase\to\dmLcBleuFour$ \\\addlinespace[2pt]
A parallel step keeps the speed limit (\cref{thm:speed-limit}) & kept in every round of the end-token chain & kept by $m_1$ of LLaDA and of Dream; base LLaDA violates it in $\parViolCount$ of $\parPreCount$ cells with $n\neq4$ \\\addlinespace[2pt]
No learning error & small in the first three rows; more than half of the schedule drift under top-$2$ & about as large as the parallel step in the first round\posthoc \\
\bottomrule
\end{tabular}
\end{table*}

\subsection{Small models with exact conditionals}\label{sec:exp-small}

We use binary sequences of length $L=12$ from an Ising chain (couplings $J=0.3$ and $0.6$) and from an end-token chain, and the copy-plus-rare-trigger family at $L=8$ with a noise floor of weight $10^{-4}$.
In every round a small masked diffusion transformer is trained from scratch on samples of $q_t$; from its conditional table on all partial assignments we compute $q_{t+1}=\lambda p+(1-\lambda)S[Q_t]$ exactly, and likewise the perfect-learning chain (\cref{app:small-models}).

\paragraph{State-independent schedules follow the map.}
Under random remasking the nearest-neighbour covariance of the Ising chain follows perfect learning (\cref{fig:small}(a), \cref{tab:scorecard}): over both couplings and step sizes, and with real data, the gap is at most $\smRandomMaxGap$, and the sequential control stays at the data.
The first-order law of \cref{sec:contraction} stays within $\foMaxGapWeak$ ($J=0.3$) and $\foMaxGapStrong$ ($J=0.6$) of the exact chain for neighbouring positions, although these chains are not weakly coupled.
Fixed pairs $(\{1,2\},\{3,4\},\dots)$ stop after one round, though at distance two the chain for $J=0.3$ keeps falling, which we cannot explain (\cref{app:small-models}).

\paragraph{Confidence rules keep drifting.}
Under perfect learning, top-$2$ by confidence and a threshold rule stop after one round on this chain. Learned chains do not (\cref{fig:small}(b)): after four rounds the ratio is $\cwConfRFour\pm\cwConfRFourSE$ under top-$2$ (perfect learning $\cwConfPerf$) and $\cwThrRFour\pm\cwThrRFourSE$ under the threshold rule ($\cwThrPerf$), far below what a model of errors in the learned table, fixed before the run, had predicted ($\cwPredConfLo$--$\cwPredConfHi$ and $\cwPredThrLo$--$\cwPredThrHi$). The ratchet of \cref{sec:lowconf}\posthoc{} gives a first-order reading. Real data limit the drift: with $\lambda=0.5$ the top-$2$ chain stays between $\clHalfPlateauLo$ and $\clHalfPlateauHi$ from round five to eight, below the perfect-learning fixed point $\clHalfPerf$.

\paragraph{Rare contexts, and what a sequential control hides.}
On the copy-plus-rare-trigger family with $\eta=10^{-2}$ (\cref{fig:rare-full} in \cref{app:small-models}), fixed pairs that decode the copy pair first lift the trigger marginal from $\ampDataRare$ to $\ampFixTrigROne$ in one round, as \cref{thm:amplification}(a) predicts.
Under random pairs, $\Prob(x_1\neq x_2)$ follows perfect learning, but the trigger marginal stays below it ($\ampRndTrigROne\pm\ampRndTrigROneSE$ against $\ampRndTrigROnePerf$): the first does not depend on the extrapolation, the second does (\cref{sec:rare}; \cref{app:small-models} gives the implied $\bar\theta$\posthoc).
In the three-term split\posthoc{} of this marginal (\cref{fig:small}(c)), the sequential arm drifts too, through learning error alone and with the opposite sign: $\mathcal B^{\mathrm{seq}}_3=\ampBSeqThree\pm\ampBSeqThreeSE$ against $\mathcal B^{\mathrm{rnd}}_3=\ampBRndThree\pm\ampBRndThreeSE$. The same-seed difference of the two arms after three rounds is thus only $\ampPairedTrig\pm\ampPairedTrigSE$, while the schedule drift is $\mathcal A_3=\ampDriftTrig$; at $\eta=10^{-3}$ the difference recovers most of the drift (\cref{app:small-models}).

\subsection{Four rounds in two pretrained models}\label{sec:exp-llm}

\begin{figure*}[t]
\centering
\includegraphics[width=\textwidth]{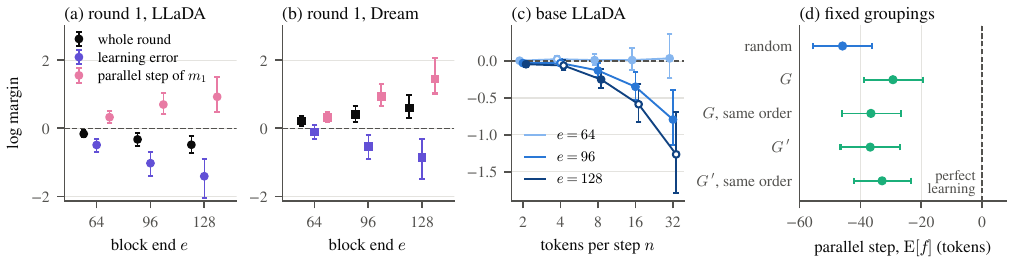}
\caption{(a)--(c) Per-question log margins to the speed limit (\cref{thm:speed-limit}); values below zero violate it. (a), (b) First round of the random chains, split as in \cref{sec:split}. (c) Base LLaDA with $n$ random tokens per step, questions $\parQLo$--$\parQHi$ (open markers: cells outside the pre-specified readout). (d) Parallel step, in $\E[f]$, of LLaDA models trained for one round on randomly remasked data (blue) or on data decoded with a fixed grouping $G$ or $G'$ (green), each decoded as its training data; ``same order'': against sequential samples in the order of the grouping. Bars: 95\% intervals from resampling questions. Values: \cref{tab:round1,tab:base-margins,tab:fixed-grouping}.}
\label{fig:termination}
\end{figure*}

\paragraph{Setup.}
We self-distill LLaDA-8B-Instruct \citep{nie2025llada} and Dream-v0-Instruct-7B \citep{ye2025dream} on GSM8K \citep{cobbe2021gsm8k}. In each round the current model answers $\llmTrainPrompts$ training questions once with its own sampler (a window of $\llmGenLength$ tokens in blocks of $\llmBlockLength$, $\llmTokensPerStep$ tokens per step, temperature $1$), and the next model continues training a LoRA adapter \citep{hu2021lora} on these answers for $\llmLoraSteps$ steps. Each model has a chain of four rounds with random and one with low-confidence remasking; for LLaDA we also ran both chains keeping only the answers scored as correct, as in dParallel \citep{chen2025dparallel}, a second chain of each unfiltered kind, a random chain with real data in every round, and further trainings of the first round (\cref{app:llada,app:dream}).
Random remasking is state independent (\cref{sec:setup}) and is our controlled setting; with four tokens per step it is a poor sampler (pass@1 of the base models $\llRndPassBase$ and $\dmRndPassBase$, against $\llLcPassBase$ and $\dmLcPassBase$ with low-confidence remasking, the rule used in practice).

Models are evaluated with their own sampler on $\llmEvalPrompts$ test questions with $\llmEvalSamples$ samples each. \emph{pass@1} scores the whole window with special tokens removed: the answer is the last number, including any content after the first end token (a tail), and the filter uses the same rule. The \emph{truncated accuracy} scores only the text before the first end token.
The \emph{dependence proxy} estimates, on $\depStates$ fixed partially decoded states of the base model, the total correlation of $\depWindow$ positions of the next block (\cref{app:llada}).

\paragraph{Random remasking breaks the termination structure.}
With random remasking (\cref{fig:rounds}(a)), the mean position of the first end token moves from $\llRndEfBase$ to $\llRndEfFour$ for LLaDA ($\llRndRepEfFour$ in the second chain) and from $\dmRndEfBase$ to $\dmRndEfFour$ for Dream, and the fraction of answers with a tail grows from $\llRndTailBase$ and $\dmRndTailBase$ to $\llRndTailFour$ in both.
Content shrinks less: after four rounds LLaDA's windows still hold $\llRndContentFour$ content tokens ($\llRndRepContentFour$ in the second chain; $\llRndContentBase$ before) and Dream's $\dmRndContentFour$ ($\dmRndContentBase$ before), most of them after the first end token.
pass@1 falls to $\llRndPassFour$ and $\dmRndPassFour$ (\cref{fig:rounds}(d)), and the truncated accuracy from $\llRndTruncBase$ and $\dmRndTruncBase$ to $\llRndTruncFour$ and $\dmRndTruncFour$.
The dependence proxy falls in every round (\cref{fig:rounds}(b)), the direction of \cref{sec:contraction}, although the models become flatter, which also lowers the proxy.
All five tests pre-specified for Dream came out as predicted (\cref{app:dream}).

\paragraph{Filtering and real data do not stop the end token.}
Keeping only the answers scored as correct (LLaDA, one chain; \cref{fig:rounds}) raises pass@1 from $\llRndPassBase$ to $\flRndPassFour$, but the first end token still moves to position $\flRndEfFour$ and the truncated accuracy falls to $\flRndTruncFour$: the number scored as correct almost always stands after the first end token, where a filter that reads the whole window still finds it (we did not filter on the truncated text).
With sequential answers of the base model as half of every training set ($\lambda=\lamLambda$; one chain, \cref{app:llada-real}), most of the dependence proxy is kept (\cref{tab:scorecard}), the direction of \cref{cor:lambda-fixed-point}, but the first end token is at $\lamEfFour$, against $\llRndEfFour$ and $\llRndRepEfFour$ in the two chains without real data ($\lamChainsGapFour$ apart through training noise alone), and still moving ($\lamStepFour$ from round three, after $\lamStepThree$); $\lamTailFour$ of the answers have a tail, and pass@1 is $\lamPassFour$.

\paragraph{Low-confidence remasking concentrates.}
With low-confidence remasking Self-BLEU rises in every chain (\cref{fig:rounds}(c), \cref{tab:scorecard}; $\llLcRepBleuFour$ in LLaDA's second chain), the direction of the sharpening of \cref{sec:lowconf}, and pass@1 falls (LLaDA $\llLcPassBase\to\llLcPassFour$, second chain $\llLcRepPassFour$; Dream $\dmLcPassBase\to\dmLcPassFour$), which our results do not explain.
In LLaDA's filtered chain pass@1 does not fall ($\flLcPassFour$) while Self-BLEU rises further, to $\flLcBleuFour$ (one chain), as in curated loops, where the reward rises while its variance collapses \citep{ferbach2024curated}.

\subsection{Sampler versus training}\label{sec:exp-split}

\paragraph{The trained model's parallel step keeps the speed limit; the excess is learning error.}
The split of \cref{fig:small}(c) needs the full conditional table, infeasible for a pretrained model; the one-round split of \cref{sec:split} needs only samples. \Cref{fig:termination}(a,b) applies it to the first round of the random chains, on training questions 0--$\speedQHi$ with $\speedSamples$ answers per question and decoding mode.
In both models the parallel step of $m_1$ is within the limit at the three block ends $e=64,96,128$ (every interval above zero).
LLaDA's whole round nevertheless moves the first end token faster than the limit allows (all three intervals below zero): the excess is in the learning-error term, because $m_1$, decoding one token at a time, already ends earlier than the data it was trained on.
Dream's learning-error term has the same sign (intervals below zero at $e=96$ and $128$, containing zero at $e=64$) but does not carry the round beyond the limit.
In tokens, the round moves $\E[f]$ by $\efqLlLearn$ through the learning error and $\efqLlStep$ through the parallel step for LLaDA, and by $\efqDmLearn$ and $\efqDmStep$ for Dream\posthoc: training contributes about as much as the sampler.
Each split rests on one training; four further trainings of LLaDA's first round also end earlier than their training data when decoded sequentially (\cref{app:llada-learning-error}).

Why $m_1$ ends earlier than its training data is open. Where no end token is visible, LLaDA's $m_1$ predicts $\calBlockMOne$ times the end tokens of held-out samples like its training data\posthoc{} ($\calBlockFourLo$--$\calBlockFourHi$ in four other trainings; base model: $\calBlockMZero$; \cref{app:llada-learning-error}). Small models trained from scratch do not end earlier; fine-tuned from a trained start they do, less than LLaDA (\cref{app:small-models}); and padding the targets with end tokens \citep{kim2025rainbow,liu2026voidpadding} is at most a minor contributor: trainings without it also end earlier.

\paragraph{Two tests whose premise fails.}
\emph{Content.} For the base LLaDA model with random remasking (\cref{fig:termination}(c); test questions $\parQLo$--$\parQHi$, the pre-specified set of \cref{app:prespecified}), $\parViolCount$ of the $\parPreCount$ pre-specified cells $(n,e)$ with $n=2,8,16,32$ violate the per-question limit, the largest at $n=32$. For $n=4$, the setting of our chains, no interval is below zero\posthoc{} (on all $250$ questions the one at $e=128$ is; \cref{app:llada}).
The data point to the premise on content: at the start of the second to fourth block the model expects more end tokens after prefixes written in parallel than after sequentially written prefixes of the same question\posthoc, the amplification of \cref{sec:rare} with the end token as the trigger.

\emph{Perfect learning.} We trained LLaDA for one round on answers decoded with a fixed grouping $G$ of each block (a random permutation of its $\llmBlockLength$ positions cut into eight groups of four), and separately with a second grouping $G'$. Under perfect learning each trained model, decoded with its own grouping, would reproduce its sequential samples (\cref{sec:split}); instead its parallel step moves $\E[f]$ by between $\fxStepMin$ and $\fxStepMax$ tokens across the two groupings and two orders of sequential decoding (\cref{fig:termination}(d)), every interval far from zero. Only $\fxSmallerCount$ of these $\fxReadouts$ steps is distinguishably smaller than that of the model trained on randomly remasked data ($\fxRefStep$), and this is an effect of the decoder\posthoc: decoded with $G$, that model's step is $\fxCrossModel$.
Averaged over the two trained models, the step under a model's own grouping cannot be told apart from its step under the other grouping (difference $\fxMatch$ $\fxMatchCI$\posthoc), and the dependence proxy of four positions decoded together falls only to $\fxDepG$ and $\fxDepAlt$ of the base model's (perfect learning: zero), about as for other positions.
One round of this fine-tuning thus does not bring LLaDA into the fixed-point family of its sampler (one training per grouping; \cref{app:llada-fixed}).

\section{Related work}\label{sec:related}

\paragraph{One pass of parallel decoding.}
Beyond the analyses cited in \cref{sec:intro}, one-pass theory covers the dependence of the step count on the metric \citep{feng2025theoretical}, lower bounds for deterministic adaptive rules \citep{wen2026serial} and end-to-end guarantees with a learned denoiser \citep{wu2026transformer,sheriff2026endtoend}. Closest to \cref{sec:contraction} is the concurrent one-pass result of \citet{li2026retention} discussed there. None of these works retrains on its samples; we iterate the pass and separate the two errors in trained models by measurement.

\paragraph{Training on generated data.}
Theories of self-consuming loops attribute the drift mostly to learning \citep{shumailov2023curse,bertrand2024stability,gerstgrasser2024accumulate,khelifa2026error}. For continuous diffusion, \citet{khelifa2026limiting} show that truncating the reverse process drives recursive training to a unique limiting distribution even with exact scores; parallel decoding plays this role in masked diffusion models.

\section{Limitations and conclusion}\label{sec:conclusion}

\paragraph{Limitations.}
Perfect learning is an idealization: the theorems do not predict the learning error, which was more than half of the schedule drift for top-$2$ by confidence and comparable to the parallel step in the first round of both pretrained models.
The rates are first-order statements near independence, with explicit remainder bounds only in very small neighbourhoods; global convergence of the total correlation and the collapse under low-confidence remasking are conjectures.
The speed limit excludes low-confidence remasking, the rule used in practice, and needs an end-token hazard that ignores content, which fails for the base LLaDA model.
The pretrained models were studied on one task with one decoding and LoRA recipe, once per Dream round, filtered chain, real-data chain (one $\lambda$), fixed grouping and first-round split.

\paragraph{Conclusion.}
Under perfect learning, training a masked diffusion language model on its own parallel samples iterates one map, the sampler, whose first-order rates, fixed points and speed limit are explicit.
Trained chains leave this map under confidence rules, in rare contexts and in both pretrained models, and sequential samples of the trained model show how much of a round is the sampler's.
Why fine-tuning a pretrained model on its own parallel samples makes it end earlier than those samples remains open.

\section*{Impact Statement}
This paper analyses what happens when masked diffusion language models are trained repeatedly on their own parallel samples, a practice used to make generation faster. It identifies mechanisms by which such loops lose diversity, amplify rare events and shorten answers, and it gives checks that practitioners can apply: comparing the parallel step of a trained model with the speed limit, and not reading the same-seed difference from a sequential control as the drift caused by the schedule. The experiments fine-tune two publicly available models on a public dataset of grade-school mathematics problems. We do not see risks specific to this work beyond those of the models studied.

\section*{AI Disclosure Statement}
Large language models of the Claude family (Anthropic) were used throughout this work under the author's direction: to survey the literature and propose candidate questions, derive the theoretical results and proofs, design and implement the experiments, analyse the results and produce the figures and tables, and draft the manuscript. Their outputs were checked by independent model instances, by scripts with exact arithmetic, and by recomputing the experiment summaries from per-sample tables. The author takes full responsibility for the content of the paper.

\bibliography{refs}
\bibliographystyle{icml2027}

\newpage
\appendix
\onecolumn
\raggedbottom
\setcounter{topnumber}{3}\setcounter{totalnumber}{4}\renewcommand{\topfraction}{0.85}
\section{Notation and summary of results}\label{app:notation}

\begin{table}[h]
\centering
\small
\caption{Notation.}
\label{tab:notation}
\begin{tabular}{@{}ll@{}}
\toprule
Symbol & Meaning \\
\midrule
$L$, $\cX$, $m=|\cX|$ & sequence length, alphabet, alphabet size \\
$p$, $q$, $q_t$ & data distribution, a distribution on $\cX^L$, the distribution after $t$ rounds \\
$\lambda$ & fraction of real data mixed in each round, \eqref{eq:loop} \\
$G=(G_1,\dots,G_K)$, $\before{k}$ & ordered partition (schedule), positions decoded before step $k$ \\
$\proj_G$ & decoding with the fixed schedule $G$, \eqref{eq:projection} \\
$\nu$, $\Sched$ & law on ordered partitions, the state-independent sampler $\E_{G\sim\nu}\proj_G$ \\
$n$, $\pi=(n-1)/(L-1)$ & tokens per step, probability that two given positions share a step \\
$\ell$ & block length of the block sampler \\
$\mu$, $u$, $u_B$, $\eps$, $U$ & product base point, $q=\mu(1+\eps u)$, ANOVA components, $U=\|u\|_\infty$ \\
$E_B(G)$, $\rho_B(\nu)$ & last group meeting $B$ contains one element of $B$; its probability \\
$\rho_*$ & $\max_{|B|\ge2}\rho_B$ \\
$\TC$, $H$, $\KL$, $\TV$ & total correlation, entropy, Kullback--Leibler divergence, total variation (nats) \\
$\kappa$, $\nu_\mu$, $G_\mu$ & adaptive policy, its frozen schedule at $\mu$, the frozen partition of a deterministic rule \\
$k$ & commits per step of low-confidence remasking (and of top-$k$) \\
$Q$, $\cond(q)$, $S[Q]$ & learned conditional table, table of a joint $q$, sampler run with $Q$ \\
$\Phi$ & one round of perfect learning, $\Phi(q)=\lambda p+(1-\lambda)S[\cond(q)]$ \\
$\psi$; $\mathcal A_t$, $\mathcal B_t$, $\mathcal C_t$ & a functional; schedule drift, learning error, response, \eqref{eq:three-term} \\
$F$, $\mathrm{vis}$, $\bar\theta$ & upward-closed set of contexts (the floor), visit probability, visit-weighted extrapolation \\
$e_j$, $f$, $n_E$, $n_{\mathrm{tail}}$ & end-token indicator, length before the first E, number of E, content after the first E \\
$m_t$ & model after $t$ rounds of self-distillation ($m_0$ the base model) \\
\bottomrule
\end{tabular}
\end{table}

\Cref{tab:fate} summarizes the multi-round behaviour of the three sampler classes under perfect learning, and what is known for learned chains. ``Proved'' refers to a statement proved in this paper (in the main text or the appendices); ``identity'' to an exact decomposition that holds by construction; ``numerical'' to exact computations or experiments; ``conjecture'' to statements supported only numerically.

\begin{table}[h]
\centering
\small
\caption{Multi-round behaviour of the three sampler classes (perfect learning unless stated otherwise).}
\label{tab:fate}
\begin{tabular}{@{}>{\raggedright\arraybackslash}p{0.17\textwidth}>{\raggedright\arraybackslash}p{0.25\textwidth}>{\raggedright\arraybackslash}p{0.25\textwidth}>{\raggedright\arraybackslash}p{0.25\textwidth}@{}}
\toprule
 & (i) State-independent schedules & (ii) Adaptive rules (look only at current conditionals) & (iii) Low-confidence remasking (looks at sampled values) \\
\midrule
Product distributions & fixed (proved) & fixed (proved) & every marginal majorized, $H(q_i')\le H(q_i)$ (proved); dependence can be created (proved) \\\addlinespace[3pt]
Marginals near independence & unchanged to first order (proved) & as (i) with the frozen schedule (proved) & move toward the mode even at product points (proved) \\\addlinespace[3pt]
Interactions near independence & multiplied by $\rho_B$ per round; for random remasking the slowest are pairs, $1-\pi$ (proved) & as (i) with $\nu_\mu$; deterministic rules stop after one round (proved) & --- \\\addlinespace[3pt]
Total correlation & local contraction at rate $\rho_*^2$ (proved); can rise after round $1$ (numerical); global convergence (conjecture) & as (i) near generic points (proved) & collapse to at most $k$ uncertain positions (numerical, conjecture) \\\addlinespace[3pt]
Real data $\lambda>0$ & first-order fixed point $\lambda/(\lambda+(1-\lambda)(1-\rho_B))$ (proved) & near generic product points, one round to the fixed point for deterministic rules (proved) & --- \\\addlinespace[3pt]
Rare contexts & amplified (proved); extrapolation enters each marginal only through a visit-weighted average $\bar\theta$ (proved) & the same entry through $\bar\theta$ for rules that choose groups from the rows of the current context (proved) & copy relation cut and triggers switched off (numerical) \\\addlinespace[3pt]
Termination & speed limit, factor $\le n$ per round (proved); from a monotone start, more tokens per step end earlier (proved) & speed limit holds if groups ignore the sampled values (proved) & the limit does not apply \\\addlinespace[3pt]
Learned chains & three-term split (identity) & first-order ratchet (proved); learned chains keep drifting (numerical) & --- \\
\bottomrule
\end{tabular}
\end{table}

\section{\texorpdfstring{Proofs for Sections~\ref{sec:setup} and~\ref{sec:contraction}; adaptive rules}{Proofs for Sections 2 and 3.1; adaptive rules}}\label{app:proofs-contraction}

Throughout, conditionals at contexts of probability zero are given by a fixed convention, the same function of the context in every round.

\subsection{One round with a fixed schedule}

\begin{proposition}[Fixed schedule]\label{prop:fixed-schedule}
Let $G=(G_1,\dots,G_K)$ be an ordered partition of $[L]$.
\begin{enumerate}
\item[(a)] $\proj_G$ is idempotent, and $\proj_G(q)=q$ if and only if $q\in\cF_G$, the family of distributions under which, for each $k$, the coordinates in $G_k$ are conditionally independent given $x_{\before{k}}$.
\item[(b)] For every $r\in\cF_G$,
$\KL(q\,\|\,r)=\KL(q\,\|\,\proj_G q)+\sum_k\sum_{i\in G_k}\E_q\bigl[\KL\bigl(q(\cdot\mid x_{\before{k}})\,\|\,r(\cdot\mid x_{\before{k}})\bigr)\bigr]$ (the inner divergence is between the conditionals of $x_i$), so $\proj_G(q)$ minimizes $\KL(q\,\|\,\cdot)$ over $\cF_G$; moreover, with $\TC(r)=\sum_iH(r_i)-H(r)$ the total correlation of a distribution $r$ with marginals $r_i$, applied to the conditional law of a group given the realized context,
\begin{equation}\label{eq:one-round-kl}
  \KL\bigl(q\,\|\,\proj_G q\bigr)=\sum_{k=1}^{K}\E_{x\sim q}\Bigl[\TC\bigl(q(x_{G_k}\mid x_{\before{k}})\bigr)\Bigr].
\end{equation}
\item[(c)] For $\lambda\in[0,1)$, the loop $q_{t+1}=\lambda p+(1-\lambda)\proj_G(q_t)$ with $q_0=p$ satisfies $q_t=q_1=\lambda p+(1-\lambda)\proj_G(p)$ for all $t\ge1$, and $\KL(p\,\|\,q_1)\le(1-\lambda)\KL(p\,\|\,\proj_G p)$.
\end{enumerate}
\end{proposition}

The identity \eqref{eq:one-round-kl} for the error of one pass is known (\citealp[Eq.~(20)]{cai2026confidence}; \citealp[Eq.~(5)]{baule2026rsa}).

\begin{proof}
(a) Under $\proj_G(q)$, the conditional law of $x_{G_k}$ given $x_{\before{k}}$ is $\prod_{i\in G_k}q(x_i\mid x_{\before{k}})$; at contexts that are null under $q$ it is the conventional one, which is the same function of the context. Marginalizing the later groups shows that $\proj_G(q)(x_i\mid x_{\before{k}})=q(x_i\mid x_{\before{k}})$ for $i\in G_k$, hence $\proj_G(\proj_G q)=\proj_G q$. If $q\in\cF_G$, the chain rule over groups gives $q=\proj_G q$; conversely $\proj_G q\in\cF_G$ by construction.

(b) If $q(x)>0$, every context $x_{\before{k}}$ has positive $q$-probability, so $\proj_G q(x)=\prod_k\prod_{i\in G_k}q(x_i\mid x_{\before{k}})>0$. If some $x$ with $q(x)>0$ has $r(x)=0$, take the first $k$ with $r(x_{\before{k+1}})=0<r(x_{\before{k}})$; then some $r(x_i\mid x_{\before{k}})$ with $i\in G_k$ vanishes while $q(x_i\mid x_{\before{k}})>0$, so both sides of the identity are $+\infty$. Otherwise, writing $q(x)=\prod_k q(x_{G_k}\mid x_{\before{k}})$ and $r(x)=\prod_k\prod_{i\in G_k}r(x_i\mid x_{\before{k}})$ on the support of $q$,
\[
\log\frac{q(x)}{r(x)}=\sum_k\log\frac{q(x_{G_k}\mid x_{\before{k}})}{\prod_{i\in G_k}q(x_i\mid x_{\before{k}})}+\sum_k\sum_{i\in G_k}\log\frac{q(x_i\mid x_{\before{k}})}{r(x_i\mid x_{\before{k}})} .
\]
Taking $\E_q$, the first sum gives $\KL(q\,\|\,\proj_G q)=\sum_k\E_q\bigl[\KL\bigl(q(x_{G_k}\mid x_{\before{k}})\,\|\,\prod_{i\in G_k}q(x_i\mid x_{\before{k}})\bigr)\bigr]=\sum_k\E_q[\TC(x_{G_k}\mid x_{\before{k}})]$, and the second gives the stated nonnegative sum.

(c) Under both $p$ and $\proj_G(p)$, the conditional law of $x_i$ given $x_{\before{k}}$ ($i\in G_k$) is $p(x_i\mid x_{\before{k}})$ (with the convention at $p$-null contexts). A mixture of two distributions with the same conditional of $x_i$ given $x_{\before{k}}$ has that conditional, so $q_1$ has the same within-group conditionals as $p$ and $\proj_G(q_1)=\proj_G(p)$; hence $q_2=q_1$, and by induction $q_t=q_1$. The divergence bound is convexity of $\KL(p\,\|\,\cdot)$.
\end{proof}

\subsection{\texorpdfstring{Linearization (\cref{thm:linearization})}{Linearization}}

\begin{proof}[Proof of \cref{thm:linearization}]
\emph{Marginals and conditionals.} Since $\mu$ is a product, summing $\mu(1+\eps u)$ over $x_{[L]\setminus V}$ gives $q(x_V)=\mu(x_V)(1+\eps u^V(x_V))$ with $|u^V|\le U$. For $i\notin V$,
\[
q(x_i\mid x_V)=\mu_i(x_i)\,\frac{1+\eps u^{V\cup\{i\}}}{1+\eps u^V},
\]
and $1+\eps u^V\ge1-\frac1{4L}>0$. Therefore $\proj_G(q)=\mu\,\Psi_G$ with
\[
\Psi_G=\prod_k\prod_{i\in G_k}\frac{1+\eps a_{k,i}}{1+\eps b_k},\qquad a_{k,i}=u^{\before{k}\cup\{i\}},\quad b_k=u^{\before{k}},\quad |a_{k,i}|,|b_k|\le U.
\]
\emph{The combinatorial identity.} Let $D_G=\sum_k\sum_{i\in G_k}(a_{k,i}-b_k)$. Since $u^{V\cup\{i\}}-u^V=\sum_{B\ni i,\,B\subseteq V\cup\{i\}}u_B$,
\[
D_G=\sum_{B\neq\emptyset}u_B\cdot\#\bigl\{(k,i):\ i\in B\cap G_k,\ B\setminus\{i\}\subseteq\before{k}\bigr\}.
\]
For a pair $(k,i)$ in this set, $G_k$ contains $i\in B$ and all other elements of $B$ lie in earlier groups, so $G_k$ is the last group meeting $B$ and $B\cap G_k=\{i\}$. Thus the count is $1$ on $E_B(G)$ and $0$ otherwise: $D_G=\sum_{B\neq\emptyset}\ind[E_B(G)]\,u_B$.

\emph{Remainder.} For $|z|\le\frac12$, $|\log(1+z)-z|\le z^2$. The logarithm of $\Psi_G$ is a sum of $L$ terms $\log(1+\eps a_{k,i})$ and $L$ terms $-\log(1+\eps b_k)$, each with $|\eps a|,|\eps b|\le\frac1{4L}\le\frac12$, so $\log\Psi_G=\eps D_G+r_G$ with $|r_G|\le2L\eps^2U^2$. Let $y=\eps D_G+r_G$. Since $|D_G|\le2LU$ and $|\eps|U\le\frac1{4L}$, $|y|\le2LU|\eps|(1+U|\eps|)\le2.5\,LU|\eps|\le\frac58$, and
\[
|\Psi_G-1-\eps D_G|\le|e^y-1-y|+|r_G|\le\tfrac12y^2e^{5/8}+2L\eps^2U^2\le(3.125\,e^{5/8}+2)L^2U^2\eps^2<8L^2U^2\eps^2 .
\]
\emph{Averaging.} $\Sched(q)=\mu\,\E_\nu\Psi_G=\mu\bigl(1+\eps\,\E_\nu D_G+R\bigr)$ with $R=\E_\nu[\Psi_G-1-\eps D_G]$, and $\E_\nu D_G=\sum_B\Prob_\nu(E_B)u_B=\sum_B\rho_Bu_B$.
\end{proof}

Two consequences used below: (i) the first-order map $u\mapsto u'=\sum_B\rho_Bu_B=\E_\nu D_G$ satisfies $\|u'\|_\infty\le2LU$; (ii) $E_{\{i\}}(G)$ always holds, so $\rho_{\{i\}}=1$ and marginals change only at order $\eps^2$.
On random distributions and schedule laws ($L\in\{3,4,6\}$, $m\in\{2,3\}$, $|\eps|\in\{1/(8LU),1/(32LU),1/(128LU)\}$), the largest ratio of the actual remainder to the bound was $\linBoundRatioMax$; the constant $8$ is loose.

\subsection{\texorpdfstring{Pairs are the slowest (\cref{lem:pairs-slowest})}{Pairs are the slowest}}

For $L=6$ and $n=2$, for example, $\rho_1,\dots,\rho_6$ are $\rhoExampleLsixNtwo$.

\begin{proof}
\emph{Formula.} A uniform ordered partition into groups of size $n$ is a uniformly random assignment of the $L$ positions to $L$ slots, slots $(b-1)n+1,\dots,bn$ forming group $b$. A set $B$ with $|B|=k$ occupies a uniformly random $k$-subset of slots. $E_B$ holds if and only if the other $k-1$ elements of $B$ lie in groups before the group of the largest occupied slot. If that slot is one of the $n$ slots of group $b$, the remaining $k-1$ slots must be chosen among the $(b-1)n$ slots of earlier groups; summing over $b$ gives the formula.

\emph{An identity.} Fix $B$ with $|B|=k+1\ge2$, let $c$ be the number of elements of $B$ in the last group meeting $B$, and $c'$ the number in the second-to-last group meeting $B$ (this group exists when $c=1$). Remove a uniformly random element of $B$; by exchangeability, the remaining set $B'$ satisfies $\Prob(E_{B'})=\rho_k$. (For $|B|=2$ the event $\{c=1,\,c'\ge2\}$ below is empty, and the conclusion is immediate.) If $c=1$, $E_{B'}$ holds unless the removed element is the one in the last group (probability $\frac1{k+1}$), in which case it holds if and only if $c'=1$. If $c=2$, $E_{B'}$ holds exactly when one of the two elements in the last group is removed (probability $\frac2{k+1}$). If $c\ge3$, it never holds. Since $\rho_{k+1}=\Prob(c=1)$,
\[
\rho_k-\rho_{k+1}=\frac{2\Prob(c=2)-\Prob(c=1,\,c'\ge2)}{k+1}.
\]
\emph{An exchange.} Let $\mathrm{Src}=\{c=1,c'\ge2\}$ and $\mathrm{Tgt}=\{c=2\}$. From a partition in $\mathrm{Src}$, swap an element of $B$ in the second-to-last group meeting $B$ with an element outside $B$ in the last group meeting $B$; the result is in $\mathrm{Tgt}$, because group sizes are preserved, the last group meeting $B$ now contains two elements of $B$, and the second-to-last one still meets $B$. Each source has $c'(n-1)\ge2(n-1)$ images. A target with a second-to-last group meeting $B$ has at most $2(n-c'_{\mathrm{t}})\le2(n-1)$ preimages, where $c'_{\mathrm{t}}\ge1$ is the number of elements of $B$ in that group: one of the two elements of $B$ in the last group is moved back, in exchange for an element outside $B$ of that group; a target without such a group has none. Counting the pairs (source, image) under the uniform law gives $\Prob(\mathrm{Src})\le\Prob(\mathrm{Tgt})$, hence $\rho_k-\rho_{k+1}\ge\Prob(c=2)/(k+1)\ge0$.
\end{proof}

The formula agrees with brute-force enumeration for $(L,n)\in\{(4,2),(6,2),(6,3),(8,2),(8,4),(9,3)\}$, and the computed $\rho_k$ are nonincreasing in $k$ for all $(L,n)$ with $L\le120$ (script \texttt{rho\_monotone.py} in \codeplace). For a general schedule law the maximum need not be at pairs: with $L=3$ there is a law under which $\rho_{\{1,2,3\}}=1$.

\subsection{\texorpdfstring{Local contraction (\cref{thm:local-contraction})}{Local contraction}}

\begin{lemma}[Second-order expansion of total correlation]\label{lem:tc-expansion}
Let $\mu$ be a full-support product, $r=\mu(1+v)$ with $\E_\mu v=0$ and $\|v\|_\infty\le\gamma\le\frac12$. Then
$\bigl|\TC(r)-\tfrac12\sum_{|A|\ge2}\|v_A\|^2_{L^2(\mu)}\bigr|\le(L+1)\gamma^3/3$.
\end{lemma}

\begin{proof}
Because $\mu$ is a product, $\KL(r\,\|\,\mu)=\TC(r)+\sum_i\KL(r_i\,\|\,\mu_i)$, and $r_i=\mu_i(1+v_{\{i\}})$. Let $f(z)=(1+z)\log(1+z)-z-z^2/2$. For $0\le z\le\frac12$, $|f(z)|\le z^3/6$ from the integral form of the Taylor remainder with $|f'''(s)|=(1+s)^{-2}\le1$; for $-\frac12\le z<0$, $f(z)=\sum_{j\ge3}|z|^j/(j(j-1))\le\frac{|z|^3}6\sum_{i\ge0}2^{-i}=\frac{|z|^3}3$. Hence $\KL(r\,\|\,\mu)=\E_\mu[(1+v)\log(1+v)]=\frac12\|v\|^2+e_0$ with $|e_0|\le\gamma^3/3$, and likewise $\KL(r_i\,\|\,\mu_i)=\frac12\|v_{\{i\}}\|^2+e_i$ with $|e_i|\le\gamma^3/3$ (since $|v_{\{i\}}|\le\gamma$). By orthogonality $\|v\|^2=\sum_{B\neq\emptyset}\|v_B\|^2$, and subtracting gives the claim.
\end{proof}

\begin{proof}[Proof of \cref{thm:local-contraction}]
With $v=\eps u$ and $\gamma=U|\eps|\le\frac1{8L}$, \cref{lem:tc-expansion} gives $\TC(q)=\frac{\eps^2}2\sum_{|A|\ge2}\|u_A\|^2+e$ with $|e|\le(L+1)U^3|\eps|^3/3$.
By \cref{thm:linearization}, $\Sched(q)=\mu(1+v')$ with $v'=\eps u'+R$, $u'=\sum_B\rho_Bu_B$, $\|u'\|_\infty\le2LU$ and $\|R\|_\infty\le8L^2U^2\eps^2\le LU|\eps|$; also $\E_\mu v'=0$. Hence $\gamma'=\|v'\|_\infty\le3LU|\eps|\le\frac38$, and \cref{lem:tc-expansion} gives $\TC(\Sched q)=\frac12\sum_{|A|\ge2}\|v'_A\|^2+e_1$ with $|e_1|\le(L+1)\gamma'^3/3\le9(L+1)L^3U^3|\eps|^3$.
By orthogonality, with $u'_{\ge2}$ and $R_{\ge2}$ the parts with $|A|\ge2$,
\[
\sum_{|A|\ge2}\|v'_A\|^2=\eps^2\sum_{|A|\ge2}\rho_A^2\|u_A\|^2+2\eps\langle u'_{\ge2},R_{\ge2}\rangle+\|R_{\ge2}\|^2 .
\]
By Cauchy--Schwarz the middle term is at most $2|\eps|\cdot2LU\cdot8L^2U^2\eps^2=32L^3U^3|\eps|^3$, and $\|R_{\ge2}\|^2\le64L^4U^4\eps^4\le8L^3U^3|\eps|^3$. Both enter $\TC(\Sched q)$ with the factor $\frac12$, so altogether $\TC(\Sched q)=\frac{\eps^2}2\sum_{|A|\ge2}\rho_A^2\|u_A\|^2+e'$ with $|e'|\le\bigl(9L^4+(16+4+9)L^3\bigr)U^3|\eps|^3\le38L^4U^3|\eps|^3$. Therefore
\[
\TC(\Sched q)\le\rho_*^2\Bigl(\TC(q)-e\Bigr)+|e'|\le\rho_*^2\,\TC(q)+\Bigl(38L^4+\tfrac{L+1}3\Bigr)U^3|\eps|^3\le\rho_*^2\,\TC(q)+56L^4U^3|\eps|^3 .\qedhere
\]
\end{proof}

To read the bound of \cref{thm:local-contraction} as a rate, take $\mu$ to be the product of the marginals of $q$, fix $u$ with a nonzero interaction part and let $\eps\to0$: then $\TC(\Sched q)/\TC(q)\le\rho_*^2+O(\eps)$. For random remasking $\rho_*=1-\pi$ (\cref{lem:pairs-slowest}), and if $u$ has only pairwise interactions the ratio tends to $(1-\pi)^2$, so the bound is tight as $\eps\to0$.

For a fixed number of rounds $t$, iterating \cref{thm:linearization} gives $q_t=\mu(1+\eps\sum_B\rho_B^tu_B+R_t)$ with $\|R_t\|_\infty\le C_t\eps^2$, hence $\TC(q_t)=\frac{\eps^2}2\sum_{|A|\ge2}\rho_A^{2t}\|u_A\|^2+O_t(\eps^3)$.
In the same numerical trials as above, the largest ratios of the actual error to its bound were $\tcBoundRatioMax$ for $\TC(q)$ (bound $(L+1)U^3|\eps|^3/3$) and $\tcSBoundRatioMax$ for $\TC(\Sched q)$ (bound $38L^4U^3|\eps|^3$).

\paragraph{Fixed point with real data (\cref{cor:lambda-fixed-point}).}
Write $q_t=\mu(1+\eps u^{(t)}+O(\eps^2))$. By \cref{thm:linearization}, $\Sched(q_t)=\mu(1+\eps\sum_B\rho_Bu^{(t)}_B+O(\eps^2))$, and \eqref{eq:loop} is linear in the distribution, so $u^{(t+1)}_B=\lambda u_B+(1-\lambda)\rho_Bu^{(t)}_B$ to first order. The fixed point solves $u^*_B=\lambda u_B+(1-\lambda)\rho_Bu^*_B$. The remainder is not uniform in $t$, so the corollary concerns the first-order coefficients.

\subsection{Local convergence}

\begin{theorem}[Local convergence]\label{thm:local-convergence}
Let $\nu$ be a state-independent schedule law with $\rho_*=\max_{|B|\ge2}\rho_B<1$, and run \eqref{eq:loop} with $\lambda=0$ from a full-support $q_0$. Let $\mu_t$ be the product of the marginals of $q_t$, $q_t=\mu_t(1+w_t)$, $a_t=\|w_t\|_{L^2(\mu_t)}$ (so $a_t^2=\chi^2(q_t\,\|\,\mu_t)$), $\underline\mu=\min_{i,x_i}\mu_{0,i}(x_i)$, $\theta=(1+\rho_*)/2$ and $\Gamma=8L^2(\underline\mu/2)^{-L}$. If
\[
a_0^2\le\min\Bigl\{\frac1{4L\Gamma},\frac{1-\theta^2}{2\Gamma}\Bigr\},\qquad(1+4L)\Gamma a_0\le\frac{1-\rho_*}4,\qquad(1+\Gamma a_0^2)^{L/2}\le\frac{2(1+\rho_*)}{1+3\rho_*},
\]
then $a_t\le\theta^ta_0$ and all marginal probabilities stay at least $\underline\mu/2$; $\mu_t$ converges to a product $\mu_\infty$ with $|\mu_{\infty,i}/\mu_{0,i}-1|\le2\Gamma a_0^2/(1-\theta^2)$; $\|q_t-\mu_t\|_1\le a_t$; $\TC(q_t)=\KL(q_t\,\|\,\mu_t)\le a_t^2\to0$; and $\limsup_t a_{t+1}/a_t\le\rho_*$ and, if $\TC(q_t)>0$ for all $t$, $\limsup_t\TC(q_{t+1})/\TC(q_t)\le\rho_*^2$.
\end{theorem}

\begin{proof}
Let $\underline\mu_t$ be the smallest marginal probability of $q_t$; we show by induction on $t$ that $\underline\mu_s\ge\underline\mu/2$ and $a_s\le\theta^sa_0$ for all $s\le t$ (step (5) sums over all earlier rounds).
(1) Since $\mu_t(x)\ge\underline\mu_t^L$, $\|w_t\|_\infty\le\underline\mu_t^{-L/2}a_t\le(\underline\mu/2)^{-L/2}a_0$, and the first condition gives $\|w_t\|_\infty\le1/(4L)$. \Cref{thm:linearization} with base point $\mu_t$, $\eps=1$ and $u=w_t$ applies: $S(q_t)=\mu_t(1+\mathcal Tw_t+R_t)$ with $\mathcal Tw=\sum_{|B|\ge2}\rho_Bw_B$ ($w_t$ has no components with $|B|\le1$ because $\mu_t$ is the product of the marginals of $q_t$) and $e_t:=\|R_t\|_\infty\le8L^2\|w_t\|_\infty^2\le\Gamma a_t^2$.
(2) The new marginals are $\mu_{t+1,i}=\mu_{t,i}(1+r_i)$ with $r_i=\E_{\mu_t}[R_t\mid x_i]$, $|r_i|\le e_t$, because $\mathcal Tw_t$ has no component with $|B|\le1$. Hence $w_{t+1}=(1+\mathcal Tw_t+R_t)/\prod_i(1+r_i)-1$.
(3) Since $Le_t\le L\Gamma a_0^2\le\frac14$, $|1/\prod_i(1+r_i)-1|\le2Le_t$; and $\|\mathcal Tw_t\|_\infty\le2L\|w_t\|_\infty\le\frac12$. Hence $\|w_{t+1}-\mathcal Tw_t\|_\infty\le(1+4L)e_t$.
(4) Orthogonality gives $\|\mathcal Tw_t\|_{L^2(\mu_t)}\le\rho_*a_t$. Changing the measure from $\mu_t$ to $\mu_{t+1}$ costs a factor at most $(1+e_t)^{L/2}$, so
\[
a_{t+1}\le\bigl(\rho_*a_t+(1+4L)e_t\bigr)(1+e_t)^{L/2}\le\bigl(\rho_*+(1+4L)\Gamma a_t\bigr)(1+\Gamma a_t^2)^{L/2}a_t\le\theta a_t
\]
by the second and third conditions.
(5) $\underline\mu_{t+1}\ge\underline\mu_t(1-e_t)$ and $\sum_{s\le t}e_s\le\Gamma\sum_sa_s^2\le\Gamma a_0^2/(1-\theta^2)\le\frac12$, so $\underline\mu_{t+1}\ge\underline\mu/2$; the same sum gives the convergence of $\mu_t$ and the bound on $\mu_\infty$.
(6) $\|q_t-\mu_t\|_1=\E_{\mu_t}|w_t|\le a_t$, and $\TC(q_t)=\KL(q_t\,\|\,\mu_t)\le\log(1+\chi^2(q_t\,\|\,\mu_t))\le a_t^2$. As $a_t\to0$ the factor in step (4) tends to $\rho_*$, and \cref{lem:tc-expansion} gives $\TC(q_t)=\frac12a_t^2+O(a_t^3)$, so when all $\TC(q_t)>0$ the ratio of consecutive total correlations has $\limsup$ at most $\rho_*^2$.
\end{proof}

The factor $(\underline\mu/2)^{-L}$ in $\Gamma$ makes the neighbourhood extremely small (it is $\deltaFactorExample$ for $\underline\mu=0.1$ and $L=16$), so the theorem is qualitative.

\subsection{Adaptive rules: frozen schedules}\label{sec:adaptive}

\paragraph{The rules.} An adaptive rule chooses the next group from the current decoding state, but not from the values about to be sampled. The \emph{confidence} of a masked position $i$ at the state $(D,x_D)$ is $\max_aq(a\mid x_D)$. Top-$k$ by confidence ranks the masked positions and decodes the $k$ most confident; a confidence threshold decodes all positions with confidence at least $\tau$, or the most confident one if there is none; the entropy budget of \citet[Alg.~1]{cai2026confidence} adds positions in a random order until the sum of their conditional entropies exceeds a budget.

An \emph{adaptive policy} $\kappa$ chooses, at a decoding state $(D,x_D)$, a nonempty group $A$ of undecoded positions with probability $\kappa_q(A\mid D,x_D)$, which may depend on $q$; the positions in $A$ are then sampled independently from $q(x_i\mid x_D)$.
The policy does not see the values it is about to sample.
For fixed $x$, the probabilities $P_q(G\mid x)=\prod_k\kappa_q(G_k\mid\before{k},x_{\before{k}})$ define a law on ordered partitions, and the output is $S_\kappa(q)(x)=\sum_G P_q(G\mid x)\,\proj_G(q)(x)$.

\begin{theorem}[Frozen schedules]\label{thm:frozen}
Let $q=\mu(1+\eps u)$ as above with $|\eps|\le1/(4LU)$, and let $\bar D=[L]\setminus D$ denote the undecoded positions.
\begin{enumerate}
\item[(a)] Every adaptive policy fixes products: $S_\kappa(\mu)=\mu$.
\item[(b)] Suppose the policy sees the context only through the conditional law of the undecoded positions, $\kappa_q(A\mid D,x_D)=F\bigl(A,D,q(x_{\bar D}\mid x_D)\bigr)$; confidences, entropies, margins and even the dependence among undecoded positions are of this form. At $\mu$ the policy does not depend on $x_D$, so it reduces to a state-independent law $\nu_\mu$, the \emph{frozen schedule}. If $F$ is $\Lambda$-Lipschitz from $\ell_1$ to total variation, then
\[
  S_\kappa(q)=\mu\Bigl(1+\eps\sum_{B\neq\emptyset}\rho_B(\nu_\mu)\,u_B+R'\Bigr)
\]
with $\|R'\|_\infty\le\bigl(8+\tfrac{32}{3}\Lambda\bigr)L^2U^2\eps^2$. Thus \cref{thm:linearization} holds with $\nu_\mu$ in place of $\nu$ and this constant in place of $8$, and \cref{thm:local-contraction} holds with $\nu_\mu$ and correspondingly larger constants on a smaller range of $\eps$.
\item[(c)] If $F$ is locally constant, $F(\cdot,D,\zeta)=F(\cdot,D,\mu(x_{\bar D}))$ whenever $\|\zeta-\mu(x_{\bar D})\|_1<\delta$, then $S_\kappa(q)=S_{\nu_\mu}(q)$ exactly once $\frac83U|\eps|<\delta$.
\end{enumerate}
\end{theorem}

Part (c) covers the hard rules in common use, at generic base points: a confidence threshold when no confidence equals the threshold and the confidences that are compared are distinct, top-$k$ by confidence when the $k$-th and $(k{+}1)$-st confidences differ, and the entropy budget of \citet[Alg.~1]{cai2026confidence} when the budget is not a partial sum of base entropies.
For the two deterministic rules, $\nu_\mu$ is a single ordered partition $G_\mu$.

\begin{corollary}[Deterministic confidence rules]\label{cor:deterministic-rules}
Near a generic product point, a deterministic confidence rule acts as the fixed schedule $G_\mu$. Hence, as long as the iterates stay in the neighbourhood of \cref{thm:frozen}(c), the loop \eqref{eq:loop} reaches its fixed point $\lambda p+(1-\lambda)\proj_{G_\mu}(p)$ after one round for every $\lambda\in[0,1)$. Random-order adaptive rules, such as the entropy budget, contract interactions, to first order, at the rates $\rho_B(\nu_\mu)$.
\end{corollary}

At non-generic points (ties, a confidence equal to the threshold, or a budget equal to a partial sum) the corollary does not apply.
The one-round stop relies on perfect learning. On a learned chain the rule is applied to a learned table, its partition can change from round to round, and \cref{prop:ratchet} shows that dependence removed by one partition does not come back at first order.

\begin{proof}[Proof of \cref{thm:frozen}]
(a) $\proj_G(\mu)=\mu$ for every $G$ and $\sum_GP_\mu(G\mid x)=1$.

\emph{A general first-order law.} Suppose that $\TV\bigl(P_q(\cdot\mid x),P_\mu(\cdot\mid x)\bigr)\le\Lambda'|\eps|$ for all $x$. The proof of \cref{thm:linearization} gives, for each $G$, $\Psi_G=1+\eps D_G+r'_G$ with $|r'_G|\le8L^2U^2\eps^2$ and $|D_G|\le2LU$. Hence
\[
\frac{S_\kappa(q)}{\mu}=\sum_GP_q(G\mid x)\,\Psi_G=1+\eps\sum_GP_\mu(G\mid x)D_G+\eps\sum_G\bigl(P_q-P_\mu\bigr)(G\mid x)\,D_G+\sum_GP_q(G\mid x)\,r'_G,
\]
so $S_\kappa(q)=\mu\bigl(1+\eps\sum_B\rho_B(x)u_B+R'\bigr)$ with $\rho_B(x)=P_\mu(E_B\mid x)$ and $\|R'\|_\infty\le(8L^2U^2+4\Lambda'LU)\eps^2$. In general $\rho_B(x)$ depends on $x$ and the first-order map is not diagonal.

(b) If the policy sees the context only through $q(x_{\bar D}\mid x_D)$, then at $\mu$ this law equals $\mu(x_{\bar D})$ and does not depend on $x_D$; so $P_\mu(G\mid x)=\nu_\mu(G)$ and $\rho_B(x)=\rho_B(\nu_\mu)$. Away from $\mu$, $q(x_{\bar D}\mid x_D)=\mu(x_{\bar D})\bigl(1+\eps(u-u^{D})/(1+\eps u^D)\bigr)$, so $\|q(x_{\bar D}\mid x_D)-\mu(x_{\bar D})\|_1\le2U|\eps|/(1-U|\eps|)\le\frac83U|\eps|$. A partition is built in at most $L$ steps, so $\Lambda'=\frac83L\Lambda U$, and the general law gives $\|R'\|_\infty\le(8+\frac{32}3\Lambda)L^2U^2\eps^2$.

(c) If $F$ is locally constant with radius $\delta$ and $\frac83U|\eps|<\delta$, then $P_q(G\mid x)=\nu_\mu(G)$ for every $x$ and $G$, so $S_\kappa(q)=S_{\nu_\mu}(q)$.
\end{proof}

\paragraph{Corollary~\ref{cor:deterministic-rules}.} For a deterministic rule $\nu_\mu$ is the point mass at $G_\mu$, and \cref{prop:fixed-schedule}(c) applies as long as the iterates stay within the radius of part (c); shrinking $\eps$ by a factor of order $L$ guarantees this.

\section{Total correlation and entropy are not monotone}\label{app:beyond-neighbourhood}

\emph{Total correlation can rise.} For $L=4$, binary values and uniform random pairs, a start drawn from a Dirichlet distribution with concentration $0.05$ (smallest probability $\tcRiseMinEntry$), iterated in exact rational arithmetic, has total correlation $\tcRiseRoundZero$, $\tcRiseRoundOne$ and $\tcRiseRoundTwo$ in rounds $0$, $1$, $2$: a rise of $\tcRiseChange$ after the first round. Among $\limitStressStarts$ random starts ($L=4$ with $n=2$, $L=6$ with $n=2,3$) this happened for $\limitStressRises$.

\emph{Entropy can fall, even near independence.} Near a product $\mu$ whose marginals are those of $q$, a second-order expansion gives $H(Sq)-H(q)=\frac{\eps^2}2\sum_{|A|\ge2}(1-\rho_A^2)\|u_A\|^2+(\text{change of the marginal entropies})$, where the first term is the loss of total correlation and the second comes from the second-order drift of the marginals, which is bilinear in the interaction components. With one position over $M$ values (one value of probability $\frac12$, the others equally likely), three binary positions, uniform random pairs and $u=u_{23}(1+\beta\tilde\ell_1)$, where $u_{23}$ is the product of the standardized values of positions $2$ and $3$ and $\tilde\ell_1$ the centred log-probability of position $1$, the expansion gives $\eps^2\|u_{23}\|^2\bigl(\frac5{18}+\frac38\beta^2v+\frac16\beta v\bigr)$ with $v=\mathrm{Var}(\log\mu_1)$, which is negative for $\beta=-\frac29$ and $v>15$. For $M=\entropyLargeAlphabetM$ ($v=\entropyLargeAlphabetS$) the exact change is $\entropyLargeAlphabetExact$ at $\eps=0.02$, $0.01$, against $\entropyLargeAlphabetPred$ from the expansion. For binary sequences with equal marginals $p=0.0832$ under random pairs, the second-order form on symmetric perturbations has a negative eigenvalue from $L=\entropyNegLMin$ to $L=\entropyNegLMax$ (the largest length computed), and exact computations of one round along the corresponding eigenvectors confirm a decrease of the entropy at every even $L$ from $\entropyExactLMin$ to $\entropyExactLMax$. In searches over random starts with $L\le8$ and small alphabets we found no decrease under uniform schedules.

\emph{Global convergence (conjecture).} For every state-independent $\nu$ with $\rho_*<1$ and every full-support start, $\TC(q_t)\to0$. Evidence: from the $\limitStressStarts$ random starts above, the total correlation after $200$ rounds was at most $\limitStressMaxTC$. The limit product can differ from the initial marginals by $O(1)$ when the start is far from independence (\cref{sec:rare}); only near independence is the change controlled by \cref{thm:local-convergence}.

\section{\texorpdfstring{Ratchets and sharpening: statements and proofs for Section~\ref{sec:lowconf}}{Ratchets and sharpening: statements and proofs for Section 3.2}}\label{app:proofs-lowconf}

\subsection{The first-order ratchet}

\Cref{cor:deterministic-rules} says that under perfect learning, near a generic product point, a deterministic confidence rule stops after one round. A learned chain applies the rule to a learned table, so the partition it uses can change from round to round, and the values are sampled from the learned table.
Work in the setting of \cref{thm:linearization}: $p=\mu(1+\eps u(p))$ and $q_t=\mu(1+\eps u^{(t)})$ with $U_t=\|u^{(t)}\|_\infty$ and $|\eps|\le1/(4LU_t)$. Round $t$ uses a given ordered partition $G_t$ and samples values from the learned table $Q_t$; the value error of the round is $S_{G_t}[Q_t]-\proj_{G_t}(q_t)=\eps\mu\,w^{(t)}$, and $w^{(t)}_B$ is its $B$-component. With real data, $q_{t+1}=\lambda p+(1-\lambda)S_{G_t}[Q_t]$.

\begin{proposition}[First-order ratchet]\label{prop:ratchet}
For every $B\neq\emptyset$,
\[
  u^{(t+1)}_B=\lambda\,u_B(p)+(1-\lambda)\bigl[\ind[E_B(G_t)]\,u^{(t)}_B+w^{(t)}_B\bigr]+O(\eps),
\]
where the $O(\eps)$ term is bounded by $2^{|B|}\cdot8L^2U_t^2|\eps|$.
\end{proposition}

Without real data ($\lambda=0$), unrolling gives
\[
  u^{(t)}_B=\Bigl[\prod_{s<t}\ind[E_B(G_s)]\Bigr]u_B(p)+\sum_{s<t}\Bigl[\prod_{s<r<t}\ind[E_B(G_r)]\Bigr]w^{(s)}_B+O(t\eps).
\]
Once some round's partition removes $B$ (for a pair: decodes both positions in one step), the part of $p$ on $B$ is gone for good; what remains on $B$ is value error injected after the last removal. The set of removed components is the union of the sets removed in each round. Under perfect learning $G_t$ is constant, so the union is reached in one round; on a learned chain the partition can change near ties, the union can grow, and at first order no partition restores what was removed.
With real data and no value error, the ratio $r^{(t)}_B=u^{(t)}_B/u_B(p)$ (for a $B$ with $u_B(p)\neq0$; pointwise for alphabets larger than two) obeys $r^{(t+1)}_B=\lambda+(1-\lambda)\ind[E_B(G_t)]\,r^{(t)}_B$ to first order and stays in $[\lambda,1]$ for every sequence of partitions; the perfect-learning fixed point is one point of this range, and changing partitions can drive a learned chain below it.
Beyond first order the ratchet is not strict: for three binary spins $s_i=\pm1$ with $q\propto1+\eps s_3(s_1+s_2)$, $s_1$ and $s_2$ are independent, and the partition $(\{3\},\{1,2\})$ gives them the covariance $\eps^2$.
The small Ising chains of \cref{sec:experiments} are not weakly coupled, so there the proposition is an accounting, not a quantitative prediction.

\begin{proof}[Proof of \cref{prop:ratchet}]
\Cref{thm:linearization} with $\nu$ the point mass at $G_t$ gives $\proj_{G_t}(q_t)=\mu\bigl(1+\eps\sum_B\ind[E_B(G_t)]u^{(t)}_B+R\bigr)$ with $\|R\|_\infty\le8L^2U_t^2\eps^2$ (for $|\eps|\le1/(4LU_t)$). Add the value error $\eps\mu\,w^{(t)}$ and mix with $\lambda p$; the map $r\mapsto(r/\mu-1)/\eps$ is linear, so taking the $B$-component gives the recursion. The $B$-component of a function has sup norm at most $2^{|B|}$ times the sup norm of the function, which bounds the remainder.
\end{proof}

\subsection{Low-confidence remasking}

\paragraph{The rule.} Low-confidence remasking \citep[Sec.~2.4 and Alg.~5]{nie2025llada} predicts every masked position, keeps the predictions with the highest confidence and remasks the rest. We study the sampled version used in the official code \citep{llada2025code} and in our experiments: at each step, draw $v_i\sim q(\cdot\mid x_D)$ independently at every masked $i$, and commit the $k$ positions with the largest $q(v_i\mid x_D)$, breaking ties uniformly at random; in the experiments it runs inside the blocks of the block sampler with $k=n$. We write $S_{\mathrm{LC}}$ for this sampler.
Low-confidence remasking chooses which positions to commit after seeing the sampled values. Unlike the samplers of \cref{sec:contraction}, it does not fix all product distributions.

\begin{theorem}[Monotone reweighting]\label{thm:lc-reweighting}
Let $q=\bigotimes_iq_i$ be a product distribution and let low-confidence remasking commit $k<L$ positions per step. The $i$-th marginal of $S_{\mathrm{LC}}(q)$ is
\[
  q'_i(v)=q_i(v)\,W_i\bigl(q_i(v)\bigr),
\]
where $W_i$ is a nondecreasing function on $[0,1]$ (depending on the other positions) with $\sum_vq'_i(v)=1$. Hence $q'_i$ majorizes $q_i$: the mode probability does not decrease and $H(q'_i)\le H(q_i)$, strictly unless $W_i$ is constant on $\{q_i(v):q_i(v)>0\}$. A position that is uniform on its support is not sharpened.
\end{theorem}

\begin{proof}[Proof of \cref{thm:lc-reweighting}]
(1) Under a product distribution the conditionals do not depend on the context, so the proposals at each step are fresh and independent across positions and steps.
(2) Fix $i$. Conditioned on $i$ not being committed at a step, the set committed among the others does not depend on $i$'s proposal: if $c_i$ is below the $k$-th largest confidence among the others, the others' top $k$ is committed; if $c_i$ equals it, then given that $i$ loses the tie, the others' choice is a uniform subset of their tied set, with the same law as in the first case.
(3) Hence, given the rest of the process, $i$ is committed at step $s$ with probability $\alpha_s(c_i)$, equal to $1$ above the threshold, the probability of winning the tie at the threshold, and $0$ below it; $\alpha_s$ is nondecreasing in $c$, and $\alpha_K\equiv1$ at the last step.
(4) The probability that $i$ ends with value $v$ is $q_i(v)\sum_s\beta_s\alpha_s(q_i(v))$, where $\beta_s=\prod_{r<s}\bigl(1-\sum_uq_i(u)\alpha_r(q_i(u))\bigr)$ does not depend on $v$. Taking the expectation over the other positions' process gives $W_i$.
(5) Order the values by decreasing $q_i$; then $W_i(q_i(v))$ is also decreasing, so each partial sum of $q'_i$ over the top values is at least that of $q_i$ (a weighted-average inequality). Entropy is Schur concave.
\end{proof}

For one commit per step the theorem is the majorization step in the proof of \citet[Thm.~1 and App.~C.1]{zhang2026differences}, who show that low-confidence remasking, and a variant with a confidence threshold, cannot increase entropy when the data factorize over positions and the model is exact; for identical positions and one commit per step, \citet[Props.~1--2]{jeon2026lcr} compute the suppression of non-top tokens exactly. To our knowledge the case $k>1$ and the consequences below are new.

\emph{Creating dependence.} With $L=2$, $k=1$, two independent binary positions with mode probability $p_0>\frac12$: both end in the minority value only if the first committed proposal is a minority value whose confidence wins (both proposals are minority, probability $(1-p_0)^2$, and the winner is either) and the second is also minority, so $\Prob(\text{both minority})=(1-p_0)^3$, while each marginal of the minority value is $\frac12(1-p_0)(2-p_0)$, and $(1-p_0)^3<\bigl(\frac12(1-p_0)(2-p_0)\bigr)^2$ is equivalent to $p_0^2>0$. At $p_0=0.8$ the total correlation created is $\lcDepTCEighty$.

\begin{proposition}[Fixed points]\label{prop:lc-fixed-points}
Let $k$ divide $L$. Every product distribution with at most $k$ positions that are not deterministic is a fixed point of $S_{\mathrm{LC}}$.
\end{proposition}

\begin{proof}
Deterministic positions have confidence $1$ and are committed first, $k$ per step. The number of remaining positions is always a multiple of $k$. At the first step at which fewer than $k$ deterministic positions remain, fewer than $2k$ positions remain in total, so exactly $k$ remain: all of them, including every uncertain position, are committed together in this last step, each sampled from its marginal without competing.
\end{proof}

\begin{conjecture}
Start from a product distribution. If no position is uniform on its support, the confidence values of all positions are pairwise distinct, and no position has two confidence values in the same gap between the confidences of the other positions, then iterating $S_{\mathrm{LC}}$ converges to a product with at most $k$ uncertain positions (per block, for block decoding).
\end{conjecture}

\emph{Numerical evidence.} On an Ising chain with $L=8$ and $J=0.3$, iterating $S_{\mathrm{LC}}$ lowers the mean marginal entropy from $\lcMargEntStart$ to $\lcMargEntKOne$ ($k=1$, $\lcNearDetKOne$ positions nearly deterministic after four rounds) and to $\lcMargEntKTwo$ ($k=2$, $\lcNearDetKTwo$ nearly deterministic).

\section{\texorpdfstring{Rare contexts and the three-term split: proofs and details for Sections~\ref{sec:rare} and~\ref{sec:split}}{Rare contexts and the three-term split: proofs and details for Sections 3.3 and 4}}\label{app:proofs-rare}

Positions $1,2$ are the copy pair and $3,\dots,L$ the triggers; $T=L-2$.

\subsection{Amplification}

\begin{proof}[Proof of \cref{thm:amplification}(a)]
The first step samples $x_1$ and $x_2$ independently from their marginals, which are both uniform. Every later context contains $(x_1,x_2)$ and triggers decoded earlier, which by induction equal $\ind[x_1\neq x_2]$; for $\eta>0$ such contexts have positive probability under $p$, and the conditional of the next trigger is the point mass at $\ind[x_1\neq x_2]$. For $m=2$, $\TC(p)=\ln2+(T-1)h(\eta)$ and $\TC(S(p))=T\ln2$.
\end{proof}

For $L=8$, $m=2$ and $\eta=10^{-3}$, one round as in (a) takes the total correlation from $\ampTCBefore$ to $\ampTCAfter$ nats.

\begin{proof}[Proof of \cref{thm:amplification}(b)]
(i) If $1$ and $2$ are in the same group, the context at that step consists of triggers only. With probability at least $(1-\eta)^{L-2}$ these triggers are all $0$; then $x_1$ and $x_2$ are sampled independently from conditionals that are symmetric in the $m$ values, and they differ with probability $1-1/m$. The two positions share a group with probability $\pi$.
(ii) For $n=2$, the first group is exactly $\{1,2\}$ with probability $1/\binom L2$, and then part (a) applies: every trigger equals $\ind[x_1\neq x_2]$, which has probability $1-1/m$. Neither step uses a context of probability zero.
\end{proof}

\begin{proposition}[Perfect learning does not determine rare contexts]\label{prop:discontinuity}
Let $S$ be a state-independent schedule. If $p$ lacks full support and $N$ has full support, $p_\eps=(1-\eps)p+\eps N$, then $S(p_\eps)\to S^{(N)}(p)$ as $\eps\to0$, where $S^{(N)}$ uses the conditionals of $N$ at the null contexts of $p$. Different $N$ can give different limits.
\end{proposition}

\begin{proof}
At contexts $c$ with $p(c)>0$ the conditional of $p_\eps$ tends to that of $p$; at contexts with $p(c)=0<N(c)$, $p_\eps(\cdot\mid c)=N(\cdot\mid c)$ for every $\eps$. The output is a finite sum of products of conditionals.
\end{proof}

\emph{Example (exact computation).} Take $L=4$, binary values, the copy-plus-rare-trigger family with $\eta=0.01$ and the fixed schedule $(\{3\},\{1,2\},\{4\})$. With uniform noise $N$, $\Prob(x_4=1)$ after one round tends to $\discUniform$ as $\eps\to0$; with the full-support noise $N'$ that puts weight $0.999$ uniformly on the states with $x_3=x_4$ and $0.001$ uniformly on all states, it tends to $\discOther$.

\subsection{The structure theorem}

Let $Q$ be a learned conditional table, with rows $Q_i(\cdot\mid c)$ for a position $i$ and a context $c=(D,x_D)$, $i\notin D$, let $S[Q]$ be the sampler run with it, and let $\mathrm{vis}(i,c)$ be the probability that the sampler decodes $i$ at the context $c$. Call a set $F$ of contexts \emph{upward closed} if every extension of a context in $F$ is in $F$, as for the contexts of probability zero.

\begin{theorem}[Structure]\label{thm:structure}
Let $\nu$ be state independent and $F$ upward closed. Until the sampler first enters $F$ it uses only rows of $Q$ outside $F$, and afterwards only rows in $F$. Consequently, for every position $j$ with binary values,
\[
  \Prob_{S[Q]}(x_j=1)=\alpha_j+\omega_j\,\bar\theta_j ,
\]
where $\alpha_j$ (decoded outside $F$ with value $1$) and $\omega_j=\sum_{c\in F}\mathrm{vis}(j,c)$ depend only on $Q$ outside $F$, and $\bar\theta_j=\sum_{c\in F}\mathrm{vis}(j,c)\,Q_j(1\mid c)/\omega_j$ is a visit-weighted average over the rows in $F$ (when $\omega_j>0$).
\end{theorem}

\begin{proof}[Proof of \cref{thm:structure}]
Until the process first enters $F$, every row it uses is at a context outside $F$, including the row used at the entering step (whose context is not yet in $F$). Hence the joint law of the entry time, the partial assignment and the schedule at entry depends only on $Q$ outside $F$. After entry every context extends a context in $F$, so it lies in $F$, and only rows in $F$ are used. For position $j$: it is decoded either before entry (value $1$ with total probability $\alpha_j$) or at a context in $F$ (total probability $\omega_j$), and $\Prob(x_j=1)=\alpha_j+\sum_{c\in F}\mathrm{vis}(j,c)Q_j(1\mid c)$.
\end{proof}

The proof extends to adaptive rules that choose groups using only the rows of the current context: before entry these rows lie outside $F$. It does not cover the general class of \cref{sec:adaptive}, whose rules may look at rows of extended contexts, which can already lie in $F$.

\subsection{Extrapolation decomposition}

For the amplification family we can compute the pieces in the limit.
Let $p_{\eta,\eps}$ be $C_{L,2}(\eta)$ mixed with uniform noise of weight $\eps$, let $F$ be the \emph{floor}, the contexts incompatible with all four states of $C_{L,2}(\eta)$, and let $Q=\cond(p_{\eta,\eps})$ outside $F$ with an arbitrary table $\varphi$ on $F$ that depends on neither $\eta$ nor $\eps$.
For an ordered partition $G$ in which $1$ and $2$ share the group $G_k$, let $z(G)$ be the number of triggers in $G_1,\dots,G_k$.

\begin{theorem}[Extrapolation decomposition]\label{thm:extrapolation}
For every state-independent $\nu$, as $\eta\to0$ and $\eps/\eta\to0$,
\begin{gather*}
  \Prob(x_1\neq x_2)\to\tfrac12\,\Prob_\nu(1,2\text{ in one group}),\\
  \frac1T\sum_{j\ge3}\Prob(x_j=1)\to a(\nu)+b(\nu)\,\bar\theta_\varphi ,
\end{gather*}
with $a(\nu)=\frac12\nu(G_1=\{1,2\})$, $b(\nu)=\E_\nu\bigl[\ind[1,2\text{ in one group},\,z\ge1]\,(T-z)\bigr]/(2T)$, and $\bar\theta_\varphi$ the average of $\varphi$ along the paths that enter the floor, weighted by their visit probabilities (defined below). For random pairs ($n=2$), $a=1/(L(L-1))$ and $b=(L-4)/(4L(L-1))$; for $n\ge3$, $a=0$.
\end{theorem}

The amplification under a random schedule thus splits into a part $a$ that passes only through contexts seen in training, and a part $b\,\bar\theta_\varphi$ that is pure extrapolation. For random pairs the two coefficients are equal at $L=8$ (both $\extrapCoefLeight$), the coefficient of the extrapolation part is larger for $L>8$ ($b/a=(L-4)/4$), and for $n\ge3$ all of the amplification is extrapolation.
In the same limit, $\Prob(x_1\neq x_2)$ does not depend on $\varphi$.

\emph{Definition of $\bar\theta_\varphi$.} For a partition $G$ in which $1,2$ share the group $G_k$ and $z(G)\ge1$, let $N_\varphi(G)$ be the expected number of triggers equal to $1$ among the $T-z$ triggers not yet decoded, when decoding continues with $\varphi$ along the remaining groups of $G$ from the state ``$x_1\neq x_2$ (each orientation with probability $\frac12$), the $z$ decoded triggers equal to $0$''. Then
\[
\bar\theta_\varphi=\frac{\sum_G\nu(G)\,\ind[1,2\text{ in one group},\,z\ge1]\,N_\varphi(G)}{\sum_G\nu(G)\,\ind[1,2\text{ in one group},\,z\ge1]\,(T-z)} .
\]

\begin{proof}[Proof of \cref{thm:extrapolation}]
There are finitely many partitions and steps, so it suffices to show that each step follows the path below with probability $1-O(\eta+\eps/\eta)$; $\varphi$ does not depend on $\eta$, and the part after entry depends continuously on the entry law.
(1) Before both $x_1$ and $x_2$ are decoded, every decoded trigger sees a context compatible with the copy states, where $p_{\eta,\eps}$ gives a trigger the value $1$ with probability $O(\eta+\eps)$; so these triggers are $0$.
(2) If $1$ and $2$ are in different groups, the later one sees the earlier one and some zeros, a context compatible with the copy states, and copies the earlier one with probability $1-O(\eta+\eps)$. All later contexts are compatible with the copy states and the triggers are $0$. The floor is never entered.
(3) If $1$ and $2$ share the group $G_k$, they are sampled independently given zeros only, each with conditional marginal tending to $\frac12$, so they differ with probability tending to $\frac12$ (each orientation $\frac14$); the triggers in the same group are $0$. If they agree, continue as in (2). If they differ: when $z=0$ (that is, $G_1=\{1,2\}$), the context $x_1\neq x_2$ is compatible with the rare states, a trigger is $1$ with conditional probability $(\eta/2+O(\eps))/(\eta/2+O(\eps))\to1$, and the same holds at every later context, so all $T$ triggers are $1$; when $z\ge1$, the context ($x_1\neq x_2$ and $z$ zeros) is in the floor, every later context is in the floor, and the expected number of ones among the remaining triggers is $N_\varphi(G)$.
(4) Collecting terms, the trigger average is $\frac1T\bigl[\sum_{G_1=\{1,2\}}\nu(G)\cdot\frac12T+\sum_{z\ge1}\nu(G)\cdot\frac12N_\varphi(G)\bigr]+o(1)=a+b\,\bar\theta_\varphi+o(1)$, and $\Prob(x_1\neq x_2)$ comes only from step (3).
\end{proof}

\emph{Closed forms.} For random remasking with $n$ tokens per step and $L=Kn$, positions $1,2$ share a group with probability $(n-1)/(L-1)$; given that, the group index $k$ is uniform on $\{1,\dots,K\}$, $z=kn-2$ and $T-z=L-kn$. For $n=2$ this gives $a=1/(L(L-1))$ and $b=(L-4)/(4L(L-1))$; for $n\ge3$, $a=0$ (the first group always contains a trigger) and $b=(n-1)(L-n)/(4(L-1)(L-2))$. For the fixed pairs $(\{1,2\},\{3,4\},\dots)$, $a=\frac12$ and $b=0$, which is \cref{thm:amplification}(a).

\emph{Numerical checks.} Counting partitions reproduces the closed forms for $\extrapClosedFormCases$ uniform schedules ($n=2$ with $L=4,6,8,10$, $n=3$ with $L=6,9$, $n=4$ with $L=8$). At $\eta=10^{-9}$ and $\eps=0$, the exact sampler agrees with $a+b\,\bar\theta_\varphi$ and with the limit of $\Prob(x_1\neq x_2)$ to within $\extrapLimitMaxDev$ for constant and random floor tables ($n=2$ with $L=6,8,10$, $n=3$ with $L=6,9$, $n=4$ with $L=8$, and a random mixture of $15$ partitions with groups of size $1$--$4$).

\subsection{The three-term decomposition}

In \eqref{eq:three-term} the response is
\[
  \mathcal C_t=\sum_{s<t}\Bigl\{\bigl[\psi(\Phi^{t-1-s}q_{s+1})-\psi(\Phi^{t-s}q_s)\bigr]-\bigl[\psi(q_{s+1})-\psi(\Phi q_s)\bigr]\Bigr\},
\]
the effect at round $t$ of the error injected in round $s$, minus that error as it was injected. With real data, $\Phi(q)=\lambda p+(1-\lambda)S[\cond(q)]$ and the identity is unchanged.

\begin{proof}[Proof of \eqref{eq:three-term}]
$\mathcal A_t+\mathcal B_t+\mathcal C_t=\psi(\Phi^tq_0)-\psi(q_0)+\sum_{s<t}\bigl[\psi(\Phi^{t-1-s}q_{s+1})-\psi(\Phi^{t-s}q_s)\bigr]$, and the sum telescopes to $\psi(q_t)-\psi(\Phi^tq_0)$.
\end{proof}

Two cases with $\lambda=0$ in which terms vanish. \emph{Sequential decoding} ($\mathcal A_t=\mathcal C_t=0$). With one position per step, $\proj_G(q)=q$ by the chain rule, and null contexts are visited with probability $0$, so $\Phi$ is the identity. \emph{Fixed schedule} ($\mathcal C_t=0$). For any table $Q$, $S_G[Q]=\prod_k\prod_{i\in G_k}Q(x_i\mid x_{\before{k}})$ lies in $\cF_G$, on which $\proj_G$ is the identity (\cref{prop:fixed-schedule}); hence $\Phi(q_s)=q_s$ for $s\ge1$, and with $\Phi^2=\Phi$ every bracket in $\mathcal C_t$ vanishes.

\section{\texorpdfstring{Termination: proofs and details for Section~\ref{sec:termination}}{Termination: proofs and details for Section 3.4}}\label{app:proofs-termination}

Positions are $0,\dots,L-1$ and blocks have $\ell$ positions. For a monotone $q$ write $\sigma(x)=\Prob_q(f>x)$ for the survival function, with $\sigma(-1)=1$ (no position decoded yet).

\paragraph{The speed limit.}
\begin{proof}[Proof of \cref{thm:speed-limit}]
If $\Prob_q(\text{no E in }D)=0$ there is nothing to prove. Otherwise, for a sequence of groups $G_1,\dots,G_K$ covering $D$, let $a_k=\Prob_q(\text{no E in }D\mid G_1,\dots,G_k\text{ content})$ and $g_k=\Prob_q(G_k\text{ content}\mid G_1,\dots,G_{k-1}\text{ content})$. Then $a_{k-1}=g_ka_k$ and $a_K=1$, so $\prod_kg_k=a_0=\Prob_q(\text{no E in }D)$ for every such sequence. The sampler draws the positions of $G_k$ independently from their conditionals, and each of them is content with probability at least $g_k$; hence $G_k$ comes out all content with probability at least $g_k^{|G_k|}\ge g_k^n$. For a fixed partition the product gives $a_0^n$. If the groups are random or depend on decoded values, induct backwards: on all-content histories $a_k=a_{k-1}/g_k$ is determined by the groups chosen so far, so $\E[\ind[G_k\text{ content}]\,a_k^n\mid\text{history}]\ge\E_{G_k}[g_k^n(a_{k-1}/g_k)^n]=a_{k-1}^n$. For binary $q$ every context on the all-content path has positive probability, so no convention is involved; if instead the E hazard ignores content, the content tokens on the path, even at contexts of probability zero, do not change the factors.
\end{proof}

\paragraph{Record formula.} Under a monotone $q$, if the decoded positions are all content and the largest of them is $M$, then $\Prob_q(e_j=0\mid\cdot)=\sigma(\max(j,M))/\sigma(M)$. Hence, for one block (given no E before it, which leaves the conditional law monotone) and a partition $G$ of it,
\[
  \Prob_{S(q)}(\text{no E in the block})=\E_G\Bigl[\prod_k\prod_{j\in G_k,\,j>M_{k-1}}\frac{\sigma(j)}{\sigma(M_{k-1})}\Bigr],
\]
where $M_{k-1}$ is the largest position in $G_1,\dots,G_{k-1}$ ($M_0=-1$): only positions beyond the current record matter, and all contexts on the path have positive probability.

\paragraph{More tokens per step can only end earlier.}
\begin{proposition}\label{prop:monotone-in-n}
Let $q$ be monotone with an E hazard that ignores content (or binary), and let $S_n$ decode blocks of $\ell$ positions from left to right with uniformly random groups of $n$ positions inside each block. If $n\mid n'\mid\ell$, then $\Prob_{S_{n'}(q)}(f\ge e)\le\Prob_{S_n(q)}(f\ge e)$ at every block end $e$.
\end{proposition}
\begin{proof}
Given no E in earlier blocks, the conditional law within a block is again monotone and, because the hazard ignores content, does not depend on the content of earlier blocks; so $\Prob(f\ge e)$ is a product of per-block probabilities and it suffices to treat one block. Cut a uniformly random permutation of the block into consecutive groups of $n$; this is the choice of $S_n$. Cutting the same permutation into groups of $n'$ gives $S_{n'}$, whose groups are unions of $n'/n$ consecutive $n$-groups. Fix the permutation and merge two consecutive groups $G,G'$ (let $M$ be the record before $G$ and $M'\ge M$ the record after it). In the record formula the factors of $G$ do not change; a position $j\in G'$ with $j>M'$ changes from $\sigma(j)/\sigma(M')$ to $\sigma(j)/\sigma(M)$, which is not larger because $\sigma$ is nonincreasing; a position with $M<j\le M'$ changes from $1$ to $\sigma(j)/\sigma(M)\le1$; positions $j\le M$ keep the factor $1$; the factors of later groups are unchanged; and if $\sigma(M')=0$ both sides vanish. Every factor is not larger, so the product is not larger; average over permutations.
\end{proof}
Monotonicity of $q$ cannot be dropped: for two positions with $\Prob(\text{both content})=\frac12$ and each of ``E then content'' and ``content then E'' with probability $\frac14$, one token per step gives $\Prob(\text{no E})=\monoCounterPOne$ and two give $\monoCounterPTwo$.

\paragraph{The per-question form.} If the speed limit holds for each prompt $z$, then for $1\le\bar n\le n$,
$P_n(e)=\E_zP_{n,z}(e)\ge\E_z[P_{1,z}(e)^n]\ge(\E_z[P_{1,z}(e)^{\bar n}])^{n/\bar n}\ge(\E_zP_{1,z}(e))^n$, the last two steps by the power-mean inequality. With four samples per prompt, $\E_z[P_{1,z}^{\bar n}]$ is estimated without bias for $\bar n\le4$ by averaging $\binom{K_z}{\bar n}/\binom4{\bar n}$ over prompts, where $K_z$ is the number of the four sequential samples with $f\ge e$ ($K_z$ is binomial with success probability $P_{1,z}$, and $\E\binom{K}{\bar n}/\binom4{\bar n}=P^{\bar n}$).

\section{Small exact models: details}\label{app:small-models}

\subsection{Setup}

\paragraph{Models and training.} Every model is a bidirectional transformer over the vocabulary $\{0,1,\text{mask}\}$ with $\smLayers$ layers, width $\smWidth$, $\smHeads$ heads, feed-forward width $\smFF$, pre-layer normalization and no dropout. A training example masks $k$ positions, $k$ uniform on $\{1,\dots,L\}$, and the loss is the cross-entropy at the masked positions. Training uses AdamW (learning rate $10^{-3}$, warm-up $500$ steps, batch $\smBatch$) in fp32, on $10^5$ samples drawn from the exact current distribution $q_t$, with a validation loss every $\smValEvery$ steps, early stopping with patience $\smPatience$, and at most $\smStepCap$ steps; the weights with the best validation loss are kept. Each round trains a new model from scratch.
\paragraph{Exact evaluation.} The conditional table is computed on all partial assignments ($3^{12}$ for $L=12$), and $q_{t+1}=\lambda p+(1-\lambda)S[Q_t]$ is computed exactly: for fixed partitions as a product, for random schedules by dynamic programming over partial assignments, and for low-confidence remasking by enumerating the proposals and the commit rule (ties within $10^{-12}$ broken uniformly). The perfect-learning chain $\Phi^t(p)$ is computed in the same way from the true conditionals, and the three-term split \eqref{eq:three-term} from both chains.
\paragraph{Data.} Ising chains $p(x)\propto\exp\bigl(J\sum_is_is_{i+1}+\sum_ih_is_i\bigr)$, $s=2x-1$, with $J\in\{0.3,0.6\}$ and fields $h_i=0.1+0.02\,(i-1)$, $i=1,\dots,12$; an end-token chain, the binary monotone law (\cref{sec:termination}) with $\Prob(f=j)=h(1-h)^j$ for $j<12$ and $h=0.05$, decoded in two blocks of six positions, mixed with uniform noise of weight $10^{-4}$; and the copy-plus-rare-trigger family at $L=8$, $m=2$ with a noise floor: weight $1-\eta-\eps$ on the two copy sequences, $\eta$ on the two sequences with $x_1\neq x_2$ and all triggers on, and $\eps=10^{-4}$ spread uniformly over all sequences. Three seeds for $L=12$ and five for $L=8$; standard errors are over seeds.
\paragraph{Arms.} Ising: sequential decoding in a uniformly random order, the fixed pairs $(\{1,2\},\{3,4\},\dots)$, random remasking with $2$ and $4$ tokens per step, random remasking with $4$ tokens per step and $\lambda=0.5$, and low-confidence remasking with $k=2$ in two blocks of six. End-token chain: in two blocks of six, sequential decoding and random remasking with $2$ or $3$ tokens per step inside the block, and low-confidence remasking with $k=2$. The confidence rules of \cref{fig:small}(b) use top-$2$ by confidence and a threshold $\tau=0.662$ with the most confident position as fallback, without blocks and with ties broken by position index, on the Ising chain with $J=0.3$.

\subsection{Further results}

\paragraph{Ising chains.} For the nearest-neighbour covariance the first-order law of \cref{sec:contraction}, $c_{t+1}=\lambda+(1-\lambda)(1-\pi)c_t$ with $\pi=(n-1)/11$, is close to the exact perfect-learning chain: after four rounds at $J=0.3$ it gives $\foRndTwoRFour$, $\foRndFourRFour$ and $\foRndLamRFour$ for the three random arms, against $\smRndTwoRFourPerf$, $\smRndFourRFourPerf$ and $\smRndLamRFourPerf$, and over these arms and rounds $1$--$4$ the largest difference is $\foMaxGapWeak$ for $J=0.3$ and $\foMaxGapStrong$ for $J=0.6$. Low-confidence remasking removes almost all dependence (total correlation $\smDataTC$ in the data, $\smLcTCRFour$ after four rounds), and the sequential control stays within $\smSeqCovDev$ of the data in the covariance ratio and within $\smSeqMargDev$ in every marginal (means over seeds). With the stronger coupling the random-pairs ratio after four rounds is $\smSixRndTwo\pm\smSixRndTwoSE$ against $\smSixRndTwoPerf$. At distance two the pair rate does not describe the exact chain ($\smDistTwoRndTwoPerf$ against $\foRndTwoRFour$ after four rounds with two tokens per step). For $J=0.3$, the relative deviations at distances two and three are larger, but the covariances are small: after four rounds the distance-two ratio is $\smDistTwoRndTwo\pm\smDistTwoRndTwoSE$ against $\smDistTwoRndTwoPerf$ under random pairs and $\smDistTwoRndFour\pm\smDistTwoRndFourSE$ against $\smDistTwoRndFourPerf$ with four tokens per step, and the distance-three ratio $\smDistThreeRndTwo\pm\smDistThreeRndTwoSE$ against $\smDistThreeRndTwoPerf$ and $\smDistThreeRndFour\pm\smDistThreeRndFourSE$ against $\smDistThreeRndFourPerf$. One deviation is unexplained: under fixed pairs the distance-two ratio for $J=0.3$ falls from $\smFixDistTwoROne$ after one round to $\smFixDistTwoRFour\pm\smFixDistTwoRFourSE$ after four, whereas perfect learning keeps it at $\smFixDistTwoPerf$ (for $J=0.6$ it moves from $\smSixFixDistTwoROne$ to $\smSixFixDistTwoRFour$). In the three-term split of the nearest-neighbour covariance (in absolute terms), the learning error and the response are at most $\smDecompMaxBC$ in absolute value in seed means (single seeds up to $\smDecompMaxBCSeed$) in every round, for fixed pairs and the random arms, while the schedule drift after four rounds lies between $\smDecompALo$ and $\smDecompAHi$.

\paragraph{End-token chain.} Random remasking breaks the structure ``once E, always E'' in the learned chains as under perfect learning, somewhat more slowly. After four rounds $\E[f]$ is $\smEosRndTwoEf$ against $\smEosRndTwoEfPerf$ with $2$ tokens per step and $\smEosRndThreeEf$ against $\smEosRndThreeEfPerf$ with $3$, and the fraction $V$ of sequences with a tail is $\smEosRndTwoV$ against $\smEosRndTwoVPerf$ and $\smEosRndThreeV$ against $\smEosRndThreeVPerf$. The mean tail length is between $\smEosTailRatioLo$ and $\smEosTailRatioHi$ times its perfect-learning value in every round. No learned round of the random-remasking chains exceeds the speed limit at the block ends $6$ and $12$ (the smallest log margin over these arms, rounds, block ends and seeds is $\smEosRMin$), and none shows the shortening learning error of the pretrained models. Low-confidence remasking raises the probability of filling both blocks without an end token from $\smEosLcSurvZero$ to $\smEosLcSurvOne$ in the first round (perfect learning $\smEosLcSurvPerf$) and then stops ($\smEosLcSurvFour$ after four rounds).

\begin{figure}[h]
\centering
\includegraphics[width=\textwidth]{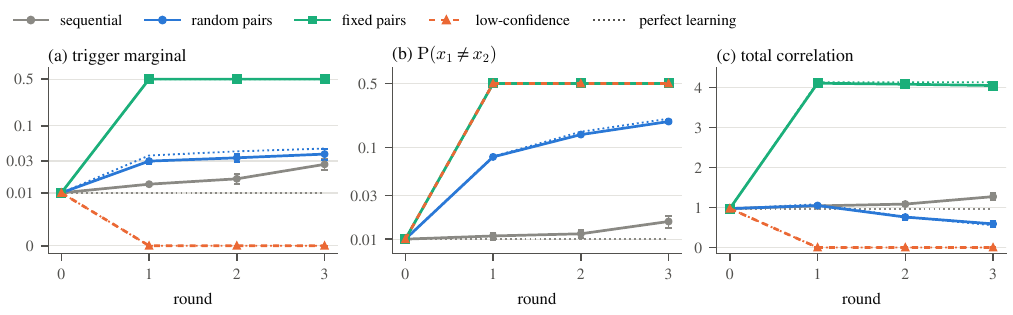}
\caption{Copy-plus-rare-trigger family ($L=8$, $m=2$, $\eta=10^{-2}$, noise floor $10^{-4}$): small masked diffusion models trained from scratch for three rounds (five seeds, mean $\pm$ one standard error) and perfect learning (dotted, computed exactly). Fixed pairs decode the copy pair first; low-confidence remasking commits $k=2$ per step. (a) Trigger marginal (logarithmic axis, linear below $0.004$); the low-confidence arm is below $\ampLcZeroBound$ from round $1$ on. (b) $\Prob(x_1\neq x_2)$ (logarithmic axis); fixed pairs and low-confidence remasking coincide at $\frac12$. (c) Total correlation in nats. \Cref{tab:amp} gives the numbers.}
\label{fig:rare-full}
\end{figure}

\paragraph{Rare contexts.} \Cref{fig:rare-full} shows all readouts of the copy-plus-rare-trigger experiment at $\eta=10^{-2}$: besides the trigger marginal, low-confidence remasking makes the copy pair independent, so that $\Prob(x_1\neq x_2)=\frac12$ as under fixed pairs, but its total correlation falls to zero, as under perfect learning. \Cref{tab:amp} gives the numbers behind the figure and behind the three-term split of \cref{fig:small}(c)\posthoc, where the learning-error bar of the third group is $\mathcal B^{\mathrm{rnd}}-\mathcal B^{\mathrm{seq}}$ and the sequential arm has $\mathcal A=\mathcal C=0$; \cref{tab:amp-small} gives the same for $\eta=10^{-3}$, where the rare rule is unstable in training and the learned trigger marginals under fixed pairs vary more across seeds; there the same-seed difference recovers most of the schedule drift (last row of \cref{tab:amp-small}). In both tables a value that rounds to zero at the precision shown is printed as $0$, and a standard error that rounds to zero is omitted. In the upper parts the zeros are the trigger marginal and the total correlation under low-confidence remasking, which are below $\ampLcZeroBound$ in the learned and in the perfect-learning chains: the sampler leaves a product distribution with all triggers off. In the lower parts, $\mathcal A$ and $\mathcal C$ of the sequential arm and $\mathcal C$ of fixed pairs vanish identically (\cref{app:proofs-rare}), and $\mathcal B$ and $\mathcal C$ of the low-confidence arm round to zero.

\begin{table}[h]
\centering
\caption{Copy-plus-rare-trigger family ($L=8$, $m=2$, $\eta=10^{-2}$, noise floor $10^{-4}$), three rounds. Top: learned chains (five seeds, mean $\pm$ standard error) and perfect learning; fixed pairs decode the copy pair first, low-confidence remasking commits $k=2$ per step. Bottom: change from round $0$ to round $3$ split as in \eqref{eq:three-term} into schedule drift $\mathcal A$, learning error $\mathcal B$ and response $\mathcal C$; the last row is the same-seed difference between the random and the sequential arm, whose $\mathcal B$ column is $\mathcal B^{\mathrm{rnd}}-\mathcal B^{\mathrm{seq}}$.}
\label{tab:amp}
\scriptsize
\setlength{\tabcolsep}{3pt}
\begin{tabular*}{\textwidth}{@{\extracolsep{\fill}}llcccccc@{}}
\toprule
& & \multicolumn{2}{c}{$\Prob(x_1\neq x_2)$} & \multicolumn{2}{c}{trigger marginal} & \multicolumn{2}{c}{total correlation (nats)}\\
\cmidrule(lr){3-4}\cmidrule(lr){5-6}\cmidrule(lr){7-8}
Arm & & round 1 & round 3 & round 1 & round 3 & round 1 & round 3\\
\midrule
Data $p$ (round 0) & & \multicolumn{2}{c}{0.01005} & \multicolumn{2}{c}{0.01005} & \multicolumn{2}{c}{0.973}\\
\addlinespace[2pt]
Sequential & learned & 0.0109\,$\pm$\,0.0010 & 0.0157\,$\pm$\,0.0023 & 0.0135\,$\pm$\,0.0007 & 0.0268\,$\pm$\,0.0050 & 1.039\,$\pm$\,0.017 & 1.271\,$\pm$\,0.082\\
 & perfect & 0.0101 & 0.0101 & 0.0101 & 0.0101 & 0.973 & 0.973\\
\addlinespace[2pt]
Fixed pairs & learned & 0.4999 & 0.4999 & 0.5001\,$\pm$\,0.0009 & 0.5003\,$\pm$\,0.0018 & 4.103\,$\pm$\,0.005 & 4.046\,$\pm$\,0.017\\
 & perfect & 0.5000 & 0.5000 & 0.4985 & 0.4985 & 4.127 & 4.127\\
\addlinespace[2pt]
Random pairs & learned & 0.0792\,$\pm$\,0.0011 & 0.1923\,$\pm$\,0.0056 & 0.0299\,$\pm$\,0.0026 & 0.0381\,$\pm$\,0.0067 & 1.049\,$\pm$\,0.029 & 0.590\,$\pm$\,0.082\\
 & perfect & 0.0800 & 0.2066 & 0.0362 & 0.0460 & 1.079 & 0.555\\
\addlinespace[2pt]
Low-confidence & learned & 0.4999 & 0.5000 & 0 & 0 & 0 & 0\\
 & perfect & 0.5000 & 0.5000 & 0 & 0 & 0 & 0\\
\bottomrule
\end{tabular*}

\vspace{12pt}
\begin{tabular*}{\textwidth}{@{\extracolsep{\fill}}lcccccccc@{}}
\toprule
& \multicolumn{4}{c}{trigger marginal} & \multicolumn{4}{c}{total correlation (nats)}\\
\cmidrule(lr){2-5}\cmidrule(lr){6-9}
Arm & change & $\mathcal A$ & $\mathcal B$ & $\mathcal C$ & change & $\mathcal A$ & $\mathcal B$ & $\mathcal C$\\
\midrule
Sequential & $+$0.0167 & 0 & $+$0.0167\,$\pm$\,0.0050 & 0 & $+$0.2981 & 0 & $+$0.2981\,$\pm$\,0.0823 & 0\\
Fixed pairs & $+$0.4903 & $+$0.4885 & $+$0.0018\,$\pm$\,0.0018 & 0 & $+$3.0726 & $+$3.1535 & $-$0.0809\,$\pm$\,0.0168 & 0\\
Random pairs & $+$0.0280 & $+$0.0360 & $-$0.0064\,$\pm$\,0.0031 & $-$0.0016\,$\pm$\,0.0049 & $-$0.3826 & $-$0.4184 & $-$0.0427\,$\pm$\,0.0254 & $+$0.0785\,$\pm$\,0.0781\\
Low-confidence & $-$0.0100 & $-$0.0100 & 0 & 0 & $-$0.9730 & $-$0.9730 & 0 & 0\\
\addlinespace[2pt]
Random $-$ sequential & $+$0.0113\,$\pm$\,0.0109 & $+$0.0360 & $-$0.0231\,$\pm$\,0.0072 & $-$0.0016\,$\pm$\,0.0049 & $-$0.6807\,$\pm$\,0.1486 & $-$0.4184 & $-$0.3408\,$\pm$\,0.1030 & $+$0.0785\,$\pm$\,0.0781\\
\bottomrule
\end{tabular*}

\end{table}

\begin{table}[h]
\centering
\caption{Copy-plus-rare-trigger family at $\eta=10^{-3}$; otherwise as \cref{tab:amp}.}
\label{tab:amp-small}
\scriptsize
\setlength{\tabcolsep}{3pt}
\begin{tabular*}{\textwidth}{@{\extracolsep{\fill}}llcccccc@{}}
\toprule
& & \multicolumn{2}{c}{$\Prob(x_1\neq x_2)$} & \multicolumn{2}{c}{trigger marginal} & \multicolumn{2}{c}{total correlation (nats)}\\
\cmidrule(lr){3-4}\cmidrule(lr){5-6}\cmidrule(lr){7-8}
Arm & & round 1 & round 3 & round 1 & round 3 & round 1 & round 3\\
\midrule
Data $p$ (round 0) & & \multicolumn{2}{c}{0.00105} & \multicolumn{2}{c}{0.00105} & \multicolumn{2}{c}{0.733}\\
\addlinespace[2pt]
Sequential & learned & 0.0011\,$\pm$\,0.0003 & 0.0014\,$\pm$\,0.0005 & 0.0017\,$\pm$\,0.0005 & 0.0022\,$\pm$\,0.0008 & 0.750\,$\pm$\,0.015 & 0.754\,$\pm$\,0.023\\
 & perfect & 0.0011 & 0.0011 & 0.0011 & 0.0011 & 0.733 & 0.733\\
\addlinespace[2pt]
Fixed pairs & learned & 0.5000 & 0.4999 & 0.4782\,$\pm$\,0.0123 & 0.4799\,$\pm$\,0.0101 & 3.824\,$\pm$\,0.157 & 3.785\,$\pm$\,0.123\\
 & perfect & 0.5000 & 0.5000 & 0.4860 & 0.4860 & 3.923 & 3.923\\
\addlinespace[2pt]
Random pairs & learned & 0.0725\,$\pm$\,0.0001 & 0.1973\,$\pm$\,0.0058 & 0.0204\,$\pm$\,0.0015 & 0.0300\,$\pm$\,0.0041 & 0.926\,$\pm$\,0.004 & 0.523\,$\pm$\,0.011\\
 & perfect & 0.0723 & 0.1994 & 0.0273 & 0.0352 & 0.987 & 0.496\\
\addlinespace[2pt]
Low-confidence & learned & 0.4999 & 0.4999\,$\pm$\,0.0001 & 0 & 0 & 0 & 0\\
 & perfect & 0.5000 & 0.5000 & 0 & 0 & 0 & 0\\
\bottomrule
\end{tabular*}

\vspace{12pt}
\begin{tabular*}{\textwidth}{@{\extracolsep{\fill}}lcccccccc@{}}
\toprule
& \multicolumn{4}{c}{trigger marginal} & \multicolumn{4}{c}{total correlation (nats)}\\
\cmidrule(lr){2-5}\cmidrule(lr){6-9}
Arm & change & $\mathcal A$ & $\mathcal B$ & $\mathcal C$ & change & $\mathcal A$ & $\mathcal B$ & $\mathcal C$\\
\midrule
Sequential & $+$0.0012 & 0 & $+$0.0012\,$\pm$\,0.0008 & 0 & $+$0.0212 & 0 & $+$0.0212\,$\pm$\,0.0232 & 0\\
Fixed pairs & $+$0.4789 & $+$0.4850 & $-$0.0061\,$\pm$\,0.0101 & 0 & $+$3.0522 & $+$3.1902 & $-$0.1380\,$\pm$\,0.1234 & 0\\
Random pairs & $+$0.0290 & $+$0.0341 & $-$0.0040\,$\pm$\,0.0030 & $-$0.0011\,$\pm$\,0.0012 & $-$0.2103 & $-$0.2368 & $-$0.0517\,$\pm$\,0.0070 & $+$0.0782\,$\pm$\,0.0062\\
Low-confidence & $-$0.0010 & $-$0.0010 & 0 & 0 & $-$0.7333 & $-$0.7333 & 0 & 0\\
\addlinespace[2pt]
Random $-$ sequential & $+$0.0278\,$\pm$\,0.0045 & $+$0.0341 & $-$0.0052\,$\pm$\,0.0035 & $-$0.0011\,$\pm$\,0.0012 & $-$0.2315\,$\pm$\,0.0201 & $-$0.2368 & $-$0.0729\,$\pm$\,0.0235 & $+$0.0782\,$\pm$\,0.0062\\
\bottomrule
\end{tabular*}

\end{table}

\emph{Estimating $\bar\theta$.} For each random-pairs model of round $0$, we replace its rows outside the floor by the true conditionals of $p$ and keep its rows on the floor. By \cref{thm:structure}, the trigger marginal of the resulting sampler is $\alpha+\omega\,\bar\theta$, where $\alpha$ and $\omega$ do not depend on the floor rows; we computed $\omega$ exactly by varying a constant floor table ($\omega=\ampOmegaSmall$ at $\eta=10^{-3}$ and $\ampOmegaLarge$ at $\eta=10^{-2}$, against the limit $b=\extrapCoefLeight$) and solved for $\bar\theta$. This gives $\bar\theta$ between $\ampThetaLowMin$ and $\ampThetaLowMax$ in $\ampThetaLowCount$ of the $10$ models (five seeds, both values of $\eta$), and $\ampThetaHighMin$ and $\ampThetaHighMax$ in the other two; the noise floor would give $\frac12$.

\paragraph{Confidence rules.} After four rounds of top-$2$ by confidence ($J=0.3$, $\lambda=0$), the change of the nearest-neighbour covariance splits into a schedule drift $\mathcal A=\cwConfDecA$, a learning error $\mathcal B=\cwConfDecB\pm\cwConfDecBSE$ and a response $\mathcal C=\cwConfDecC\pm\cwConfDecCSE$; the learning error itself splits into the choice of groups from the learned table ($\cwConfDecBSel\pm\cwConfDecBSelSE$) and the sampling of values from it ($\cwConfDecBVal\pm\cwConfDecBValSE$). Under perfect learning only $\mathcal A$ would be present, and it is reached in the first round. For the threshold rule the split is $\mathcal A=\cwThrDecA$, $\mathcal B=\cwThrDecB\pm\cwThrDecBSE$ and $\mathcal C=\cwThrDecC\pm\cwThrDecCSE$: here most of the extra drift is the response of later rounds to small errors. With real data, the top-$2$ chains were run for eight rounds at $\lambda=0.5$ and twelve at $\lambda=0.2$, and the threshold rule for four rounds at $\lambda=0.5$; \cref{fig:small}(b) shows all of them. After four rounds at $\lambda=0.5$ the threshold rule is at $\clThrHalfRFour\pm\clThrHalfRFourSE$ (perfect learning $\clThrHalfPerf$); at $\lambda=0.2$ the top-$2$ ratio is $\clFifthREight$ at round eight and $\clFifthRTwelve\pm\clFifthRTwelveSE$ at round twelve ($\clFifthPerf$). At $\lambda=0.5$ the top-$2$ chain stops below the perfect-learning value by the pre-specified classification of point estimates. At $\lambda=0.2$ the change from round eight to round twelve is $\clFifthDiff\pm\clFifthDiffSE$, which fits slow convergence to a balance near $\clFifthBalance$ and also a chain that has already stopped.

\subsection{Fine-tuning from a trained start}\label{app:warm-start}

To separate fine-tuning from training from scratch, we start from the round-$0$ model of the end-token chain (trained on $p$) and continue training it on the same samples of $q_1=S[Q_0]$ that the from-scratch model of round $1$ used, with a fresh AdamW optimizer, a constant learning rate after a short warm-up, and no early stopping. One configuration copies the LLaDA recipe ($2{,}000$ samples, batch $16$, learning rate $10^{-4}$); two others use all $10^5$ samples with batch $512$ and learning rate $10^{-3}$ or $10^{-4}$. The \emph{overshoot ratio} of $\E[f]$ is $[\E_{\mathrm{ck}}[f]-\E_{Q_0}[f]]/[\E_{q_1}[f]-\E_{Q_0}[f]]$, where $\E_{\mathrm{ck}}$ and $\E_{Q_0}$ are under sequential decoding of the checkpoint and of the starting model: a ratio above $1$ means that the fine-tuned model, decoding one token at a time, moved past its training distribution.
With the LLaDA recipe, after $250$ steps the ratio is $\warmRndTwo\pm\warmRndTwoSE$ for the data of random remasking with $2$ tokens per step and $\warmRndThree\pm\warmRndThreeSE$ with $3$; all six (arm, seed) pairs lie between $\warmPerLo$ and $\warmPerHi$, above the from-scratch models of round $1$ ($\warmScratchLo$--$\warmScratchHi$). With all samples and learning rate $10^{-4}$ the ratio peaks early ($\warmPeakRndTwo\pm\warmPeakRndTwoSE$ and $\warmPeakRndThree\pm\warmPeakRndThreeSE$ at step $125$) and is $\warmEndRndTwo\pm\warmEndRndTwoSE$ and $\warmEndRndThree\pm\warmEndRndThreeSE$ after $10{,}000$ steps. Fine-tuning from a trained start can thus overshoot without low-rank adapters and in a binary model; the excess over training from scratch is modest, and the size of LLaDA's overshoot is not reproduced (\cref{app:llada}).

\section{LLaDA-8B: details}\label{app:llada}

\subsection{Protocol}

\paragraph{Model, data and sampler.} LLaDA-8B-Instruct \citep{nie2025llada} in bf16. GSM8K (main configuration): a fixed set of $\llmTrainPrompts$ training questions, the same in every round and chain, and a fixed set of $\llmEvalPrompts$ test questions; the prompt is the question in the chat template, without few-shot examples. Answers are generated in a window of $\llmGenLength$ tokens with blocks of $\llmBlockLength$ and $\llmTokensPerStep$ tokens per step ($64$ steps), temperature $1$ and no classifier-free guidance, one sequence at a time, with random remasking or low-confidence remasking. The answer of a sequence is the last number in its decoded text.
\paragraph{Rounds.} In round $t$ the model $m_{t-1}$ answers each training question once with the sampler of its chain; the chains that keep only correct answers (as in \citealp{chen2025dparallel}) filter this set. The model $m_t$ continues training the LoRA adapter of $m_{t-1}$ (rank $\llmLoraRank$, $\alpha=\llmLoraAlpha$, dropout $0.05$, on the seven linear layers of every transformer block; the first round starts from a fresh adapter) with AdamW at learning rate $10^{-4}$ after a $10$-step warm-up, no weight decay and gradient clipping at $1$, for $\llmLoraSteps$ steps of $\llmLoraBatch$ sequences (micro-batches of $2$, accumulation $8$); the training set is cycled in random order, so the filtered chains see their smaller sets several times. The loss is the supervised fine-tuning loss of LLaDA: the prompt is not masked, the $\llmGenLength$ generated answer tokens, including the end tokens the model wrote, are masked at a random ratio, and answers are right-padded with the end-of-sequence token within a micro-batch. Sampling uses the unmerged fp32 adapter.
\paragraph{Readouts.} Every model is evaluated with its own sampler on the $\llmEvalPrompts$ test questions with $\llmEvalSamples$ samples each. \emph{Scoring.} pass@1 decodes the whole answer window with special tokens removed and takes the last number, so content after the first end token counts; the chains that keep only correct answers filter with the same rule. The truncated accuracy applies the rule to the text before the first end token. Self-BLEU \citep{zhu2018texygen} is the token-level BLEU of each answer against the other answers to its question, on sequences truncated at the first end token; in the random chains the truncated text shrinks with $\E[f]$, so we do not interpret their Self-BLEU. \emph{Dependence proxy.} We fix $\depStates$ partially decoded states $a$ (the first two of eight blocks decoded) made by the base model with random remasking (used for the random chains) and, separately, $\depStates$ states made with low-confidence remasking (used for the low-confidence chains). For each model we sample the first $\depWindow$ positions $W$ of the third block from left to right and average $\log q_{\mathrm{chain}}(x_W\mid a)-\sum_{i\in W}\log q(x_i\mid a)$, where $q_{\mathrm{chain}}$ is the joint law given by the model's left-to-right conditionals and $q(x_i\mid a)$ its conditional of one position given $a$. The expectation is $\KL\bigl(q_{\mathrm{chain}}\,\big\|\,\prod_iq(x_i\mid a)\bigr)$, the total correlation of $q_{\mathrm{chain}}$ when the model's conditionals are consistent, and otherwise this plus their inconsistency.
\paragraph{Replicates.} The first round was trained three times for the random chain and twice for the low-confidence chain. The original trainings use random stream A. Stream C changes the initialization of the adapter, the order of the training data and the random draws during training (masking ratios, masks and dropout); stream B, used only in the random chain, keeps the initialization and the order of A and changes only the draws during training. B is therefore less independent of A than C is; retraining with the same stream is not bitwise reproducible on the GPU. The training with stream C starts the second chain of each rule, which then continues for three rounds with fresh streams. The filtered chains and the chain with real data have one chain each.
\paragraph{Vocabulary padding.} LLaDA's output layer has $\llVocabRows$ rows and its tokenizer $\llVocabTok$ tokens. We did not mask the $\llVocabPad$ padding rows when sampling; in a scan of $\llPadScanSamples$ generations from our LLaDA experiments, $\llPadScanHits$ contain a padding id, all of them with random remasking, and none of the $\llPadScanLcMain$ low-confidence generations of the main experiment does.

\subsection{Chains}

\Cref{tab:llada-chains} lists the readouts of the LLaDA chains without real data (with real data: \cref{tab:real-data}). In the random chain that keeps only correct answers, pass@1 rises from round to round while the truncated accuracy falls: the filter selects answers whose last number is correct, and that number increasingly stands after the first end token. We did not run a chain that filters on the truncated text.

\begin{table}[h]
\centering
\caption{LLaDA chains: pass@1 ($\pm$ standard error), truncated accuracy (not computed for the second chains), Self-BLEU, mean position of the first end token $\E[f]$, fraction of answers with a tail, mean numbers of end tokens and of tail tokens (all over $\llmEvalPrompts\times\llmEvalSamples$ answers), and the dependence proxy on the $\depStates$ fixed states of the chain's own remasking rule ($\pm$ standard error over states). Filtered chains keep only correct answers. The second chains start from $m'_0=m_0$; $m'_1$ is one of the first-round retrainings.}
\label{tab:llada-chains}
\scriptsize
\begin{tabular}{@{}llcccccccc@{}}
\toprule
Chain & model & pass@1 & truncated & Self-BLEU & $\E[f]$ & tail & $\E[n_E]$ & $\E[n_{\mathrm{tail}}]$ & dependence proxy\\
\midrule
Random & $m_0$ & 0.135\,$\pm$\,0.012 & 0.129 & 0.186 &  199.4 & 0.273 &  \phantom{0}45.7 &  \phantom{0}10.9 &  9.99\,$\pm$\,0.97\\
 & $m_1$ & 0.041\,$\pm$\,0.007 & 0.029 & 0.125 &  104.5 & 0.862 &  \phantom{0}92.0 &  \phantom{0}59.5 &  8.55\,$\pm$\,0.59\\
 & $m_2$ & 0.033\,$\pm$\,0.007 & 0.023 & 0.112 &  \phantom{0}77.4 & 0.977 &  \phantom{0}84.8 &  \phantom{0}93.8 &  7.37\,$\pm$\,0.45\\
 & $m_3$ & 0.040\,$\pm$\,0.007 & 0.019 & 0.103 &  \phantom{0}56.0 & 1.000 &  \phantom{0}65.2 &  134.8 &  5.01\,$\pm$\,0.33\\
 & $m_4$ & 0.030\,$\pm$\,0.007 & 0.024 & 0.097 &  \phantom{0}47.0 & 1.000 &  \phantom{0}66.0 &  143.0 &  2.55\,$\pm$\,0.20\\
\addlinespace[2pt]
Random, second chain & $m'_1$ & 0.028\,$\pm$\,0.006 & --- & 0.092 &  \phantom{0}89.0 & 0.818 &  120.2 &  \phantom{0}46.8 &  9.33\,$\pm$\,0.60\\
 & $m'_2$ & 0.024\,$\pm$\,0.005 & --- & 0.082 &  \phantom{0}51.9 & 0.968 &  130.2 &  \phantom{0}74.0 &  8.49\,$\pm$\,0.42\\
 & $m'_3$ & 0.021\,$\pm$\,0.005 & --- & 0.079 &  \phantom{0}38.5 & 0.999 &  128.1 &  \phantom{0}89.4 &  5.08\,$\pm$\,0.37\\
 & $m'_4$ & 0.025\,$\pm$\,0.006 & --- & 0.079 &  \phantom{0}31.9 & 1.000 &  130.0 &  \phantom{0}94.1 &  3.53\,$\pm$\,0.26\\
\addlinespace[2pt]
Random, filtered & $m_1$ & 0.129\,$\pm$\,0.014 & 0.068 & 0.133 &  127.5 & 0.712 &  \phantom{0}69.1 &  \phantom{0}59.4 &  9.10\,$\pm$\,0.65\\
 & $m_2$ & 0.133\,$\pm$\,0.013 & 0.035 & 0.136 &  \phantom{0}88.0 & 0.930 &  \phantom{0}64.1 &  103.9 &  6.11\,$\pm$\,0.43\\
 & $m_3$ & 0.173\,$\pm$\,0.017 & 0.022 & 0.115 &  \phantom{0}55.1 & 0.995 &  \phantom{0}41.8 &  159.1 &  4.24\,$\pm$\,0.30\\
 & $m_4$ & 0.217\,$\pm$\,0.019 & 0.016 & 0.128 &  \phantom{0}48.0 & 0.998 &  \phantom{0}29.2 &  178.8 &  2.89\,$\pm$\,0.22\\
\addlinespace[2pt]
Low-confidence & $m_0$ & 0.612\,$\pm$\,0.022 & 0.609 & 0.624 &  235.9 & 0.016 &  \phantom{0}19.6 &  \phantom{0}\phantom{0}0.6 &  7.12\,$\pm$\,0.86\\
 & $m_1$ & 0.570\,$\pm$\,0.023 & 0.570 & 0.633 &  233.8 & 0.034 &  \phantom{0}21.3 &  \phantom{0}\phantom{0}0.9 &  7.81\,$\pm$\,0.85\\
 & $m_2$ & 0.546\,$\pm$\,0.024 & 0.547 & 0.682 &  229.4 & 0.073 &  \phantom{0}25.4 &  \phantom{0}\phantom{0}1.2 &  5.98\,$\pm$\,0.63\\
 & $m_3$ & 0.522\,$\pm$\,0.024 & 0.519 & 0.722 &  223.1 & 0.053 &  \phantom{0}31.9 &  \phantom{0}\phantom{0}1.0 &  6.14\,$\pm$\,0.65\\
 & $m_4$ & 0.501\,$\pm$\,0.025 & 0.498 & 0.734 &  224.1 & 0.053 &  \phantom{0}30.7 &  \phantom{0}\phantom{0}1.2 &  5.56\,$\pm$\,0.62\\
\addlinespace[2pt]
Low-confidence, second chain & $m'_1$ & 0.557\,$\pm$\,0.024 & --- & 0.646 &  237.5 & 0.027 &  \phantom{0}17.7 &  \phantom{0}\phantom{0}0.8 &  7.62\,$\pm$\,0.85\\
 & $m'_2$ & 0.473\,$\pm$\,0.024 & --- & 0.669 &  239.6 & 0.052 &  \phantom{0}15.2 &  \phantom{0}\phantom{0}1.2 &  6.14\,$\pm$\,0.70\\
 & $m'_3$ & 0.483\,$\pm$\,0.025 & --- & 0.724 &  240.3 & 0.032 &  \phantom{0}14.5 &  \phantom{0}\phantom{0}1.3 &  6.24\,$\pm$\,0.75\\
 & $m'_4$ & 0.395\,$\pm$\,0.025 & --- & 0.722 &  238.5 & 0.047 &  \phantom{0}16.9 &  \phantom{0}\phantom{0}0.6 &  5.37\,$\pm$\,0.63\\
\addlinespace[2pt]
Low-confidence, filtered & $m_1$ & 0.665\,$\pm$\,0.022 & 0.665 & 0.683 &  233.1 & 0.017 &  \phantom{0}22.8 &  \phantom{0}\phantom{0}0.1 &  7.13\,$\pm$\,0.64\\
 & $m_2$ & 0.666\,$\pm$\,0.023 & 0.663 & 0.742 &  235.2 & 0.022 &  \phantom{0}20.0 &  \phantom{0}\phantom{0}0.8 &  7.29\,$\pm$\,0.75\\
 & $m_3$ & 0.666\,$\pm$\,0.024 & 0.664 & 0.772 &  229.7 & 0.021 &  \phantom{0}25.5 &  \phantom{0}\phantom{0}0.8 &  6.21\,$\pm$\,0.68\\
 & $m_4$ & 0.652\,$\pm$\,0.025 & 0.647 & 0.800 &  232.4 & 0.022 &  \phantom{0}22.5 &  \phantom{0}\phantom{0}1.0 &  6.34\,$\pm$\,0.67\\
\bottomrule
\end{tabular}

\end{table}

\paragraph{Dependence proxy.} On the fixed random-remasking states, the proxy of the random chain changes by $\llDepDiff\pm\llDepDiffSE$ from $m_0$ to $m_4$ (paired by state; $\llRndRepDepDiff$ in the second chain), and its first-round change lies between $\llDepROneLo$ and $\llDepROneHi$ across the three trainings. A flatter model also has a lower proxy, and the models do become flatter: the mean conditional entropy in the window rises from $\llHWBase$ to $\llHWFour$. The proxy therefore shows the direction of \cref{sec:contraction} without separating it from this flattening. In the random chains the proxy also falls on the fixed low-confidence states over four rounds, after a rise in the first. On the fixed low-confidence states, the dependence proxy after four rounds differs from the base model's by $\llLcDepDiff\pm\llLcDepDiffSE$ and $\llLcRepDepDiff\pm\llLcRepDepDiffSE$ in the two low-confidence chains and by $\llLcFiltDepDiff\pm\llLcFiltDepDiffSE$ in the filtered chain (paired by state); re-evaluating the base model itself with other sampling seeds moves it by $\llBaseReevalDep\pm\llBaseReevalDepSE$. Unlike the small end-token chain (\cref{app:small-models}), LLaDA's low-confidence chains fill the window less often\posthoc: on the $\lcFullWindowQuestions$ questions for which two answers of the base model fill it, the probability of filling it falls from $\lcFullWindowBase$ to $\lcFullWindowFour$ after four rounds ($\lcFullWindowRep$ in the second chain).

\subsection{Real data in every round}\label{app:llada-real}

This chain repeats the random chain with one change. In every round, half of the $\llmTrainPrompts$ training answers are sequential answers of the base model to training questions $0$--$\lamRealQHi$ (one each, the same $\lamRealN$ answers in all four rounds; mean first end token at $\lamRealEf$, tail fraction $\lamRealTail$, accuracy $\lamRealAcc$), and the other half are parallel answers of the previous round's model to questions $\lamGenQLo$--$\lamGenQHi$. In the map of \cref{sec:setup} this is $\lambda=\lamLambda$ pooled over questions (no question has both kinds of answer), with the base model's own distribution as the data. Training and evaluation follow the protocol above, with the test questions and evaluation seeds of the random chain. \Cref{tab:real-data} lists the chain next to the two chains without real data.

\begin{table}[h]
\centering
\caption{LLaDA with real data in every round ($\lambda=\lamLambda$, one chain) and the two random chains without real data: mean position of the first end token and fraction of answers with a tail (over $\llmEvalPrompts\times\llmEvalSamples$ answers), dependence proxy on the $\depStates$ fixed random-remasking states ($\pm$ standard error over states), and $\E[f]$ of the generated half of each round's training set (the real half has $\lamRealEf$ in every round).}
\label{tab:real-data}
\scriptsize
\begin{tabular}{@{}c ccc ccc cc c@{}}
\toprule
 & \multicolumn{3}{c}{$\E[f]$} & \multicolumn{3}{c}{fraction with a tail} & \multicolumn{2}{c}{dependence proxy} & training set\\
\cmidrule(lr){2-4}\cmidrule(lr){5-7}\cmidrule(lr){8-9}\cmidrule(l){10-10}
round & real data & none & none (2nd) & real data & none & none (2nd) & real data & none & $\E[f]$, generated half\\
\midrule
0 & \multicolumn{3}{c}{199.4} & \multicolumn{3}{c}{0.273} & \multicolumn{2}{c}{9.99\,$\pm$\,0.97} & ---\\
1 &  117.2 &  104.5 &  89.0 & 0.784 & 0.862 & 0.818 & 8.97\,$\pm$\,0.68 & 8.55\,$\pm$\,0.59 &  198.0\\
2 &  \phantom{0}77.8 &  \phantom{0}77.4 &  51.9 & 0.939 & 0.977 & 0.968 & 9.24\,$\pm$\,0.64 & 7.37\,$\pm$\,0.45 &  118.8\\
3 &  \phantom{0}90.5 &  \phantom{0}56.0 &  38.5 & 0.975 & 1.000 & 0.999 & 8.83\,$\pm$\,0.66 & 5.01\,$\pm$\,0.33 &  \phantom{0}79.4\\
4 &  \phantom{0}60.8 &  \phantom{0}47.0 &  31.9 & 0.997 & 1.000 & 1.000 & 7.30\,$\pm$\,0.54 & 2.55\,$\pm$\,0.20 &  \phantom{0}92.0\\
\bottomrule
\end{tabular}

\end{table}

\paragraph{Pre-specified readouts.} Three readouts were fixed before the run, all for the fourth round and paired by question (\cref{app:prespecified}). The first end token is later than in both chains without real data: the differences are $\lamEfDiffOrig$ $\lamEfDiffOrigCI$ and $\lamEfDiffRep$ $\lamEfDiffRepCI$ tokens. Fewer answers with a tail than in both chains: not met, contrary to the expectation recorded with the protocol; the difference is $\lamTailDiff$ $\lamTailDiffCI$ against either chain, in which every answer has a tail. The change of $\E[f]$ from round three to round four is $\lamStepFour$ $\lamStepFourCI$, in the bin ``still shortening'', after a rise by $\lamStepThree$ in round three that also shows in the training answers generated by the same model ($\lamGenEfFour$ after $\lamGenEfThree$). The intervals cover evaluation noise only: the two chains without real data differ from each other by $\lamChainsGapFour$ tokens at round four, and at round two the chain with real data is level with the first of them ($\lamEfDiffOrigTwo$ $\lamEfDiffOrigTwoCI$). The real answers are the same in every round; memorization may lower their effective weight.

\paragraph{Dependence and flattening\posthoc.} On the fixed states the proxy of this chain lies between $\lamDepMinEarly$ and $\lamDepMaxEarly$ in the first three rounds and is $\lamDepFour$ after the fourth, $\lamDepDiffFour\pm\lamDepDiffFourSE$ above the random chain on the same states; the second chain without real data differs from the first by only $\lamChainsDepGapFour\pm\lamChainsDepGapFourSE$. The mean conditional entropy in the window is $\lamHWFour$ after four rounds, against $\llHWFour$ in the random chain (difference $\lamHWDiffFour$ $\lamHWDiffFourCI$; base model $\llHWBase$). Real data thus keep both the proxy and the sharpness of the conditionals closer to the base model; the proxy does not separate the two. These quantities were planned, without a decision rule.

\paragraph{Training sets and model outputs.} The training set of round four has its first end token at $\lamTrainEfFour$ on average and a tail in $\lamTrainTailFour$ of its answers; the model trained on it answers the test questions with the first end token at $\lamEfFour$ and a tail in $\lamTailFour$ of the answers; its pass@1 is $\lamPassFour$ (truncated accuracy $\lamTruncFour$), although $\lamRealAcc$ of the real answers are correct. We drew no sequential samples of the models of this chain, so its learning error and its parallel step are not separated.

\subsection{The first round and the speed limit}

Write $\smp{Q}_0$ and $\smp{Q}_1$ for sequential decoding (one token per step) of $m_0$ and $m_1$, $\smp{D}_0$ for the parallel samples of $m_0$ (the training data of round $1$) and $\smp{D}_1$ for those of $m_1$, all on training questions, and $P_X(e)$ for the fraction with $f\ge e$. In the pooled form, the whole round has the log margin $\log P_{\smp{D}_1}-4\log P_{\smp{D}_0}$, which splits exactly into the parallel step of $m_1$, $\log P_{\smp{D}_1}-4\log P_{\smp{Q}_1}$, and the learning error $4(\log P_{\smp{Q}_1}-\log P_{\smp{D}_0})$; the parallel step of the base model is $\log P_{\smp{D}_0}-4\log P_{\smp{Q}_0}$. The per-question form replaces each $P_X^4$ by $\E_z[P_{X,z}^4]$, estimated without bias by averaging $\binom{K_z}4/\binom r4$ over questions, where $K_z$ of the $r$ samples of question $z$ have $f\ge e$ (\cref{app:proofs-termination}); it uses training questions $0$--$\speedQHi$ with $r=\speedSamples$ samples per question and decoding mode, one of which comes from the earlier experiments (with the four new ones alone, every term at every block end keeps its classification). \Cref{tab:round1} gives all terms; intervals resample questions.

\begin{table}[h]
\centering
\caption{First-round split of the log margin to the speed limit ($n=4$) at three block ends; 95\% intervals from resampling questions (for the pooled forms, the resampling fixed for the comparison with Dream). Negative values exceed the limit.}
\label{tab:round1}
\scriptsize
\begin{tabular}{@{}llccc@{}}
\toprule
Model, form & term & $e=64$ & $e=96$ & $e=128$\\
\midrule
LLaDA, per question (questions 0--99) & whole round & $-$0.163 [$-$0.258, $-$0.063] & $-$0.326 [$-$0.507, $-$0.135] & $-$0.483 [$-$0.738, $-$0.220]\\
 & learning error & $-$0.489 [$-$0.684, $-$0.314] & $-$1.025 [$-$1.403, $-$0.689] & $-$1.408 [$-$2.035, $-$0.895]\\
 & parallel step of $m_1$ & $+$0.326 [$+$0.159, $+$0.514] & $+$0.698 [$+$0.411, $+$1.034] & $+$0.925 [$+$0.480, $+$1.504]\\
 & parallel step of $m_0$ & $+$0.006 [$-$0.041, $+$0.061] & $-$0.012 [$-$0.093, $+$0.073] & $+$0.040 [$-$0.068, $+$0.160]\\
\addlinespace[2pt]
Dream, per question (questions 0--99) & whole round & $+$0.211 [$+$0.077, $+$0.366] & $+$0.411 [$+$0.192, $+$0.664] & $+$0.594 [$+$0.297, $+$0.981]\\
 & learning error & $-$0.109 [$-$0.304, $+$0.084] & $-$0.526 [$-$0.902, $-$0.183] & $-$0.854 [$-$1.488, $-$0.312]\\
 & parallel step of $m_1$ & $+$0.319 [$+$0.185, $+$0.481] & $+$0.937 [$+$0.664, $+$1.288] & $+$1.448 [$+$1.021, $+$2.059]\\
 & parallel step of $m_0$ & $+$0.098 [$+$0.016, $+$0.193] & $+$0.191 [$+$0.057, $+$0.340] & $+$0.494 [$+$0.291, $+$0.736]\\
\addlinespace[2pt]
LLaDA, pooled (questions 0--199) & whole round & $-$0.281 [$-$0.426, $-$0.130] & $-$0.347 [$-$0.622, $-$0.062] & $-$0.328 [$-$0.718, $+$0.066]\\
 & learning error & $-$0.750 [$-$1.056, $-$0.466] & $-$1.228 [$-$1.723, $-$0.772] & $-$1.753 [$-$2.425, $-$1.151]\\
 & parallel step of $m_1$ & $+$0.469 [$+$0.188, $+$0.777] & $+$0.881 [$+$0.459, $+$1.342] & $+$1.426 [$+$0.850, $+$2.057]\\
 & parallel step of $m_0$ & $-$0.016 [$-$0.057, $+$0.035] & $-$0.079 [$-$0.163, $+$0.015] & $-$0.133 [$-$0.254, $-$0.005]\\
\addlinespace[2pt]
Dream, pooled (questions 0--199) & whole round & $+$0.163 [$-$0.038, $+$0.380] & $+$0.670 [$+$0.314, $+$1.055] & $+$1.089 [$+$0.657, $+$1.575]\\
 & learning error & $-$0.136 [$-$0.429, $+$0.136] & $-$0.055 [$-$0.485, $+$0.376] & $-$0.633 [$-$1.309, $+$0.034]\\
 & parallel step of $m_1$ & $+$0.299 [$+$0.085, $+$0.542] & $+$0.725 [$+$0.368, $+$1.104] & $+$1.722 [$+$1.199, $+$2.302]\\
 & parallel step of $m_0$ & $+$0.094 [$-$0.032, $+$0.234] & $+$0.219 [$+$0.006, $+$0.447] & $+$0.912 [$+$0.562, $+$1.299]\\
\bottomrule
\end{tabular}

\end{table}

\subsection{Fixed groupings}\label{app:llada-fixed}

\paragraph{Design.} If the training data are decoded with a fixed grouping, perfect learning puts the trained model into the fixed-point family of that projection, and decoding it in parallel with the same grouping reproduces its sequential samples (\cref{sec:split}). A grouping is a permutation of the $\llmBlockLength$ positions of a block, cut in its order into eight groups of four, used in all blocks and samples; $G$ and $G'$ are the first two permutations of a random generator whose seed was fixed in advance. For each grouping the base model answers the $\fxTrainN$ training questions once, and a model is trained on these answers with the recipe of the chains. On training questions $0$--$\fxQHi$ the models are decoded once per question in parallel with the decoders of \cref{tab:fixed-grouping} and sequentially, in a uniformly random order and in the order of the grouping. The model trained on randomly remasked data is $m_1$ of the random chain. Readouts and decision rules were fixed before these samples were generated (\cref{app:prespecified}). Each grouping has one training (a planned second one with $G$ was not run); the intervals cover the sampling of questions only.
\begin{table}[h]
\centering
\caption{One round of LLaDA on data decoded with a fixed grouping ($G$ or $G'$) or with random remasking (the first round of the random chain). Differences of $\E[f]$ are paired by question on training questions $0$--$\fxQHi$, with one answer per question and decoding mode; the dependence proxy uses $\fxStates$ fixed states; 95\% intervals. Under perfect learning the parallel step of a model decoded with the grouping of its training data is zero, and so is its dependence proxy on the positions of one group. The decoder contrasts of the base model are post hoc.}
\label{tab:fixed-grouping}
\scriptsize
\begin{tabular}{@{}lcccc@{}}
\toprule
& & \multicolumn{3}{c}{model trained for one round on data decoded with}\\
\cmidrule(l){3-5}
& base model & random remasking & $G$ & $G'$\\
\midrule
\multicolumn{5}{@{}l}{\emph{Parallel step in $\E[f]$ against sequential decoding in random order, by decoder}}\\
decoded with random remasking & --- & $-$45.9 [$-$55.6, $-$36.1] & $-$33.8 [$-$43.9, $-$23.9] & ---\\
decoded with $G$ & --- & $-$31.6 [$-$41.5, $-$21.5] & $-$29.3 [$-$38.9, $-$19.3] & $-$32.1 [$-$42.0, $-$22.0]\\
decoded with $G'$ & --- & --- & $-$40.2 [$-$49.9, $-$30.4] & $-$36.8 [$-$46.5, $-$27.0]\\
\addlinespace[2pt]
\multicolumn{5}{@{}l}{\emph{The same against sequential decoding in the order of the grouping}}\\
decoded with its own grouping ($G$ for the base model) & $-$14.4 [$-$21.6, $-$7.1] & --- & $-$36.5 [$-$45.9, $-$26.8] & $-$32.9 [$-$42.2, $-$23.2]\\
\addlinespace[2pt]
\multicolumn{5}{@{}l}{\emph{Decoder contrasts on one model, $\E[f]$}}\\
$G$ minus $G'$ & $+$18.6 [$+$9.8, $+$27.5] & --- & $+$10.9 [$+$1.3, $+$20.6] & $+$4.6 [$-$4.8, $+$14.1]\\
$G$ minus random remasking & $+$13.0 [$+$4.7, $+$21.4] & $+$14.3 [$+$5.5, $+$23.1] & $+$4.6 [$-$5.1, $+$14.4] & ---\\
\addlinespace[2pt]
\multicolumn{5}{@{}l}{\emph{The round, $\E[f]$}}\\
learning error & --- & $-$56.6 [$-$66.6, $-$46.6] & $-$31.1 [$-$40.1, $-$21.9] & $-$33.1 [$-$43.2, $-$22.7]\\
whole round & --- & $-$102.5 [$-$111.2, $-$93.7] & $-$60.4 [$-$68.9, $-$51.8] & $-$69.8 [$-$79.1, $-$60.4]\\
\addlinespace[2pt]
\multicolumn{5}{@{}l}{\emph{Dependence proxy relative to the base model}}\\
four positions of one group of $G$ & 1 & 0.641 [0.546, 0.757] & 0.686 [0.587, 0.805] & ---\\
four positions of four groups of $G$ & 1 & 0.605 [0.511, 0.720] & 0.693 [0.592, 0.811] & ---\\
four positions of one group of $G'$ & 1 & 0.510 [0.417, 0.610] & --- & 0.593 [0.485, 0.714]\\
four positions of four groups of $G'$ & 1 & 0.627 [0.523, 0.748] & --- & 0.523 [0.421, 0.642]\\
\bottomrule
\end{tabular}

\end{table}

\paragraph{The parallel step does not vanish.} All four pre-specified readouts have intervals below zero (\cref{tab:fixed-grouping}): with $G$ against sequential decoding in random order the step cannot be separated from half the step of the random chain, and the other three are not reduced. The pre-specified comparison with the step of the random chain finds a smaller step in $\fxSmallerCount$ of the $\fxReadouts$ readouts. That one difference is exactly the sum of two terms\posthoc: the difference between the two models under the same decoder $G$, $\fxSameDecoderDiff$ $\fxSameDecoderDiffCI$, and the effect of decoding the model of the random chain with $G$ instead of random remasking, $\fxDecRndG$ $\fxDecRndGCI$; it is a property of the decoder.
\paragraph{The trained models do not recognize their grouping.} The steps of a trained model under its own decoder are not closer to zero than its steps under another decoder (\cref{tab:fixed-grouping}). The two pre-specified conditions on one model under two decoders both fail, and the decoders themselves differ already on the base model\posthoc{} (decoder contrasts in \cref{tab:fixed-grouping}), so the difference between $G$ and $G'$ on the model trained with $G$ need not reflect the training. The match contrast removes the difference between the decoders: averaged over the two trained models, $\E[f]$ under the own grouping minus under the other is $\fxMatch$ $\fxMatchCI$\posthoc. It equals the mean matched step minus the mean mismatched step, and would be $\fxMatchPerfect$ if the matched steps were zero and the mismatched steps as measured.

\paragraph{Dependence within a group.} The proxy is taken for the four positions of the first group in the third block, given the first two blocks of training answers decoded with the grouping, and for four control positions from four different groups. In the training data the positions of one group are conditionally independent given the state, so under perfect learning the proxy of the trained model vanishes there. Its ratio to the base model's proxy (\cref{tab:fixed-grouping}) is in the band ``less than half removed'' for $G$ and cannot be separated from one half for $G'$. The pre-specified contrasts with the positions from different groups and with the model of the random chain contain zero for both groupings, so the decrease is not specific to the grouping. As in the chains, the window entropy rises with training, so the decrease need not be dependence removed by the grouping.

\paragraph{Learning error.} The learning error of the two rounds with fixed groupings is $\fxLearnG$ and $\fxLearnAlt$ tokens of $\E[f]$, against $\fxLearnRnd$ in the random round\posthoc; the training data differ as well (mean first end token $\fxDataEfG$, $\fxDataEfAlt$ and $\fxDataEfRnd$), and with one training per grouping we do not conclude that a fixed grouping reduces the learning error.
\paragraph{Reading.} The prediction that a fixed schedule stops after one round fails in LLaDA under this recipe, and the premise that fails is perfect learning: after $\llmLoraSteps$ LoRA steps the model has not acquired the conditional independence of its training data. A possible reason (untested) is the loss: it masks the answer at a random ratio, independently over positions, so contexts in which all earlier groups are decoded and a whole group is masked are essentially never presented, and the independence within a group is visible only in such contexts. The test was not run on Dream.

\subsection{Tokens per step on the base model}

\Cref{tab:par} shows the base model with random remasking and $n=1,\dots,32$ tokens per step on the evaluation questions ($\parAnswersPerN$ answers per $n$), and \cref{tab:base-margins} the per-question margins of \eqref{eq:per-question} on questions $\parQLo$--$\parQHi$ (the pre-specified question set: the data of questions $0$--$49$ were in the repository before the readout was fixed). The data for $n=4$ are the base model's evaluation in the main chains, so the $\parFourCount$ cells for $n=4$ are descriptive on both question sets, and the pre-specified cells are the $\parPreCount$ with $n=2,8,16,32$ ($e\le128$ for $n\le8$, $e\le96$ for $n=16,32$). On all $250$ questions, the cell $n=4$, $e=128$ and the cell $n=2$, $e=96$ also have intervals below zero ($\parAllFourOTE$ $\parAllFourOTECI$ and $\parAllTwoNS$ $\parAllTwoNSCI$). From a monotone start, more tokens per step can only end earlier (\cref{prop:monotone-in-n}); $\E[f]$ in \cref{tab:par} falls with $n$, but the answers at $n=1$ already have some tails, so this readout is only descriptive. \Cref{tab:block-content} gives the content term referred to in \cref{sec:exp-split}: the summed end-token belief at the start of a block after prefixes decoded with $n$ tokens per step, minus the same after sequentially decoded prefixes, paired by question. The values for $n=2,8,16$ were added after the data were seen, and those for $n=4$ come from a separate run; all are descriptive.

\begin{table}[h]
\centering
\caption{Base LLaDA with random remasking and $n$ tokens per step ($250$ questions $\times$ $4$ answers).}
\label{tab:par}
\scriptsize
\begin{tabular}{@{}rcccccc@{}}
\toprule
$n$ & steps & pass@1 & $\E[f]$ & tail & $\E[n_E]$ & $\E[n_{\mathrm{tail}}]$\\
\midrule
1 &  256 & 0.497\,$\pm$\,0.022 &  233.4 & 0.012 &  21.5 &  \phantom{0}\phantom{0}1.1\\
2 &  128 & 0.324\,$\pm$\,0.020 &  222.5 & 0.070 &  31.9 &  \phantom{0}\phantom{0}1.7\\
4 &  \phantom{0}64 & 0.135\,$\pm$\,0.012 &  199.4 & 0.273 &  45.7 &  \phantom{0}10.9\\
8 &  \phantom{0}32 & 0.051\,$\pm$\,0.008 &  153.9 & 0.626 &  58.7 &  \phantom{0}43.4\\
16 &  \phantom{0}16 & 0.026\,$\pm$\,0.006 &  112.1 & 0.899 &  44.2 &  \phantom{0}99.7\\
32 &  \phantom{0}\phantom{0}8 & 0.036\,$\pm$\,0.008 &  \phantom{0}80.0 & 0.990 &  20.9 &  155.0\\
\bottomrule
\end{tabular}

\end{table}

\begin{table}[h]
\centering
\caption{Per-question log margins $\log P_n(e)-\frac n{\bar n}\log\E_z[P_{1,z}(e)^{\bar n}]$, $\bar n=\min(n,4)$, of the base model to the speed limit: for $n\le4$ the bound is the middle term of \eqref{eq:per-question}, for $n>4$ the weaker bound $(\E_z[P_{1,z}(e)^4])^{n/4}$ (\cref{app:proofs-termination}) (questions $\parQLo$--$\parQHi$; 95\% intervals). Bold: pre-specified cells ($n=2,8$ at $e\le128$; $n=16,32$ at $e\le96$) with the interval below zero; italics: cells read only descriptively, among them all cells for $n=4$.}
\label{tab:base-margins}
\scriptsize
\begin{tabular}{@{}rccccc@{}}
\toprule
$e$ & $n=2$ & $n=4$ & $n=8$ & $n=16$ & $n=32$\\
\midrule
64 & $+$0.006 [$-$0.014, $+$0.028] & \textit{$+$0.026 [$-$0.009, $+$0.069]} & $+$0.014 [$-$0.058, $+$0.101] & $+$0.014 [$-$0.123, $+$0.182] & $+$0.037 [$-$0.233, $+$0.371]\\
96 & $-$0.027 [$-$0.054, $+$0.003] & \textit{$-$0.027 [$-$0.071, $+$0.028]} & \textbf{\boldmath $-$0.127 [$-$0.219, $-$0.022]} & \textbf{\boldmath $-$0.348 [$-$0.524, $-$0.149]} & \textbf{\boldmath $-$0.792 [$-$1.134, $-$0.395]}\\
128 & \textbf{\boldmath $-$0.042 [$-$0.075, $-$0.007]} & \textit{$-$0.060 [$-$0.123, $+$0.012]} & \textbf{\boldmath $-$0.249 [$-$0.370, $-$0.107]} & \textit{$-$0.585 [$-$0.822, $-$0.306]} & \textit{$-$1.264 [$-$1.789, $-$0.694]}\\
160 & \textit{$+$0.019 [$-$0.026, $+$0.067]} & \textit{$+$0.007 [$-$0.087, $+$0.111]} & \textit{$-$0.116 [$-$0.292, $+$0.093]} & \textit{$-$0.253 [$-$0.623, $+$0.145]} & \textit{$-$0.308 [$-$1.122, $+$0.523]}\\
192 & \textit{$+$0.067 [$+$0.011, $+$0.124]} & \textit{$+$0.033 [$-$0.090, $+$0.161]} & \textit{$-$0.061 [$-$0.282, $+$0.194]} & \textit{$-$0.041 [$-$0.507, $+$0.458]} & \textit{$+$0.370 [$-$0.673, $+$1.420]}\\
224 & \textit{$+$0.124 [$+$0.050, $+$0.201]} & \textit{$+$0.145 [$-$0.002, $+$0.307]} & \textit{$+$0.195 [$-$0.096, $+$0.504]} & \textit{$+$0.471 [$-$0.105, $+$1.090]} & \textit{$+$1.585 [$+$0.233, $+$2.843]}\\
\bottomrule
\end{tabular}

\end{table}

\begin{table}[h]
\centering
\caption{Content term $C_n$ at the start of blocks 1--3, the second to fourth block (blocks are numbered from $0$; questions $\parQLo$--$\parQHi$, paired by question; 95\% intervals).}
\label{tab:block-content}
\scriptsize
\begin{tabular}{@{}rccc@{}}
\toprule
$n$ & block 1 & block 2 & block 3\\
\midrule
2 & $+$0.073 [$-$0.014, $+$0.152] & $+$0.439 [$+$0.236, $+$0.668] & $+$0.310 [$+$0.105, $+$0.530]\\
4 & $+$0.230 [$+$0.123, $+$0.337] & $+$0.553 [$+$0.403, $+$0.721] & $+$0.540 [$+$0.328, $+$0.753]\\
8 & $+$0.460 [$+$0.344, $+$0.581] & $+$1.088 [$+$0.889, $+$1.304] & $+$1.188 [$+$0.903, $+$1.567]\\
16 & $+$0.663 [$+$0.519, $+$0.812] & $+$1.393 [$+$1.213, $+$1.577] & $+$1.508 [$+$1.236, $+$1.799]\\
32 & $+$0.671 [$+$0.553, $+$0.792] & $+$1.516 [$+$1.338, $+$1.704] & $+$1.075 [$+$0.850, $+$1.292]\\
\bottomrule
\end{tabular}

\end{table}

\emph{The decomposition at $n=32$.} With $n=32$ a whole block is decoded in one step, so its survival can be compared with the record formula computed from sequential decoding. Summed up to $e=128$ on questions $\parQLo$--$\parQHi$, the shortfall of the log survival, $\parDecompR$, splits exactly into the content term $C=\parDecompC$ (the summed end-token belief at the block start after parallel prefixes minus that after sequential prefixes), a term $I=\parDecompD$ that vanishes in expectation for consistent conditionals (post hoc interval $\parDecompDCI$), and two terms that compare $-\log$ of a survival with a mean number of end tokens ($\parDecompNA$ under $n=32$ and $\parDecompNB$ under sequential decoding): the content term carries the violation (the pooled $C$ includes a reweighting of questions, estimated at $\parDecompW$, which the paired terms of \cref{tab:block-content} do not contain). Two pre-specified count tests check the recorded beliefs against the observed end tokens (passed) and, at $n=1$, a necessary condition for consistent conditionals (not rejected, which does not show consistency; in block $1$ alone $I=\parDecompDOne$ $\parDecompDOneCI$, a weak post hoc indication against it).

\subsection{Where the learning error comes from}\label{app:llada-learning-error}

\paragraph{Calibration.} On parallel samples of the base model for the evaluation questions, which are not used in training, take contexts in blocks $2$--$6$ whose last position is masked and that show no end token, and compare, over their masked positions, the summed end-token probability with the number of end tokens actually at those positions. For $m_1$ the ratio is $\calBlockMOne$ when the context masks the text as in block-by-block decoding and $\calTrainMOne$ when it masks the answer at random as in training (the latter driven by a few questions); for the base model it is $\calBlockMZero$ and $\calTrainMZero$. In the four trainings of the padding design below, after $250$ steps, the two ratios are $\calBlockFourLo$--$\calBlockFourHi$ and $\calTrainFourLo$--$\calTrainFourHi$.
\paragraph{Padding.} Four further trainings of round $1$ form a $2\times2$ design: with or without the end tokens that pad the answers, for two random streams. Removing the padding changes the end-token excess on contexts without a visible end token by $\padDATF$ $\padDATFCI$ and $\padDBTF$ $\padDBTFCI$ in the two streams, so the pre-specified test gives opposite signs and no conclusion; changing only the random stream moves the same readout by $\padStreamLo$--$\padStreamHi$ (steps $125$ and $250$); after $125$ steps both streams give a decrease, $\padShareLo$--$\padShareHi$ of the excess. Both trainings without padding still end earlier than their training data when decoded sequentially (overshoot ratios of $\E[f]$, defined as in \cref{app:warm-start} with $\smp{Q}_0$ and $\smp{D}_0$ in place of the starting model and $q_1$: $\padOvershootNopad$ and $\padOvershootNopadAlt$, against $\padOvershootCkpt$ and $\padOvershootPadAlt$ with padding). Padding is therefore at most a minor contributor.

\section{Dream-7B: details}\label{app:dream}

\paragraph{Protocol.} Dream-v0-Instruct-7B \citep{ye2025dream} follows the LLaDA protocol of \cref{app:llada}: the same training and test questions, answer window, blocks of $\llmBlockLength$ with $\llmTokensPerStep$ tokens per step, temperature, LoRA recipe and readouts, with a random chain and a low-confidence chain of four rounds. Dream was trained once per round. Dream predicts each token from the output at the preceding position (it is initialized from an autoregressive model; \citealp[Sec.~4.1]{ye2025dream}); our sampler applies this shift, which we checked on reference answers. The end tokens are the two ids of end-of-text and end-of-turn; the end-of-turn id, the mask token and padding ids occur in none of the generations, so $f$ is the same whether E means end-of-text only or either id. The output layer has $\dmVocabRows$ rows and the tokenizer $\dmVocabTok$ tokens; for Dream we mask the $\dmVocabPad$ padding rows when sampling values (their logits are set to $-\infty$), and compute all other probabilities over the full vocabulary, except the dependence proxy, which uses the masked distribution that is actually sampled. LLaDA's padding rows were not masked (\cref{app:llada}).

\paragraph{Pre-specified tests.} Before any Dream result was read, we fixed five tests with their decision rules (timing: \cref{app:prespecified}). All five came out as predicted: in the random chain, from $m_0$ to $m_4$, $\E[f]$ changed by $\dmTestEf$ $\dmTestEfCI$, the tail fraction by $\dmTestTail$ $\dmTestTailCI$ and $\E[n_{\mathrm{tail}}]$ by $\dmTestNtail$ $\dmTestNtailCI$; the dependence proxy on the fixed random-remasking states changed by $\dmTestDep$ $\dmTestDepCI$ relative to the base model; and in the low-confidence chain Self-BLEU changed by $\dmTestBleu$ $\dmTestBleuCI$ (intervals pair questions or states). The protocol also fixed, without expected directions, $\dmOpenCells$ cells on three open questions (\cref{tab:round1}) and $\dmCompCells$ comparisons of directions with LLaDA's chains, of which $\dmCompSame$ agree. The tests concern the total change over four rounds; which part of it is learning error is identified only in the first round, below. As for LLaDA, the model also flattens (the window entropy on the fixed random-remasking states rises from $\dmHWBase$ to $\dmHWFour$), and the proxy also decreases on the other chain's states.

\begin{table}[h]
\centering
\caption{Dream chains; columns as in \cref{tab:llada-chains}. Dream was trained once per round.}
\label{tab:dream-chains}
\scriptsize
\begin{tabular}{@{}llcccccccc@{}}
\toprule
Chain & model & pass@1 & truncated & Self-BLEU & $\E[f]$ & tail & $\E[n_E]$ & $\E[n_{\mathrm{tail}}]$ & dependence proxy\\
\midrule
Random & $m_0$ & 0.058\,$\pm$\,0.009 & 0.049 & 0.092 &  169.6 & 0.452 &  \phantom{0}68.5 &  17.9 &  10.23\,$\pm$\,0.73\\
 & $m_1$ & 0.018\,$\pm$\,0.004 & 0.025 & 0.064 &  114.8 & 0.923 &  \phantom{0}77.6 &  63.6 &  \phantom{0}8.30\,$\pm$\,0.56\\
 & $m_2$ & 0.026\,$\pm$\,0.006 & 0.016 & 0.067 &  \phantom{0}64.2 & 0.994 &  123.0 &  68.7 &  \phantom{0}7.08\,$\pm$\,0.46\\
 & $m_3$ & 0.017\,$\pm$\,0.004 & 0.020 & 0.058 &  \phantom{0}68.0 & 0.996 &  109.2 &  78.9 &  \phantom{0}5.12\,$\pm$\,0.32\\
 & $m_4$ & 0.023\,$\pm$\,0.005 & 0.021 & 0.053 &  \phantom{0}52.9 & 1.000 &  116.4 &  86.6 &  \phantom{0}3.10\,$\pm$\,0.25\\
\addlinespace[2pt]
Low-confidence & $m_0$ & 0.415\,$\pm$\,0.020 & 0.415 & 0.429 &  150.4 & 0.099 &  104.8 &  \phantom{0}0.8 &  \phantom{0}8.37\,$\pm$\,0.93\\
 & $m_1$ & 0.305\,$\pm$\,0.018 & 0.313 & 0.471 &  136.2 & 0.286 &  116.5 &  \phantom{0}3.3 &  \phantom{0}7.05\,$\pm$\,0.82\\
 & $m_2$ & 0.280\,$\pm$\,0.018 & 0.291 & 0.488 &  159.3 & 0.336 &  \phantom{0}92.0 &  \phantom{0}4.7 &  \phantom{0}6.34\,$\pm$\,0.70\\
 & $m_3$ & 0.325\,$\pm$\,0.021 & 0.323 & 0.613 &  147.4 & 0.262 &  106.0 &  \phantom{0}2.6 &  \phantom{0}5.44\,$\pm$\,0.64\\
 & $m_4$ & 0.303\,$\pm$\,0.022 & 0.312 & 0.624 &  168.2 & 0.284 &  \phantom{0}80.8 &  \phantom{0}7.1 &  \phantom{0}5.57\,$\pm$\,0.70\\
\bottomrule
\end{tabular}

\end{table}

\paragraph{The first round.} \Cref{tab:round1} gives the first-round split of the log margin for both models. The per-question form for Dream uses training questions $0$--$\dqQHi$ with $\dqSamples$ answers per question and decoding mode, one of them from the chain experiment, and the decision rules of the LLaDA experiment, fixed before the new samples were generated. The parallel step of $m_1$ is within the limit at the three block ends, and so is the whole round; the learning-error term is below zero at $e=96$ and $128$ and undecided at $e=64$. The base model's own parallel step is within the limit, whereas LLaDA's is undecided. With the four new samples alone every classification is the same except the base step at $e=64$, which becomes undecided.

In tokens, with one answer per question on training questions $0$--$\poolQHi$, the learning error of $\E[f]$ is $\dmEfLearn$ $\dmEfLearnCI$ and the parallel step of $m_1$ $\dmEfStepOne$ $\dmEfStepOneCI$ for Dream, against $\llEfLearn$ $\llEfLearnCI$ and $\llEfStepOne$ $\llEfStepOneCI$ for LLaDA. On the per-question samples\posthoc{} (means of the five answers per question) the two terms are $\efqDmLearn$ $\efqDmLearnCI$ and $\efqDmStep$ $\efqDmStepCI$ for Dream and $\efqLlLearn$ $\efqLlLearnCI$ and $\efqLlStep$ $\efqLlStepCI$ for LLaDA. In both models $m_1$, decoding sequentially, ends earlier than its training data, with more end tokens and more tails.

\paragraph{The low-confidence chain.} Self-BLEU rises and pass@1 falls, as for LLaDA, and on the fixed low-confidence states both the dependence proxy and the mean conditional entropy in the window decrease. The tail fraction rises more than for LLaDA ($\dmLcTailBase$ to $\dmLcTailFour$). Dream's base model already writes tails when decoding sequentially ($\dmSeqTailCount$ of $\dmSeqTailOf$ answers), so the source of these tails cannot be identified. $\E[f]$ and $\E[n_E]$ alternate in sign from round to round; whether these changes exceed the variation between trainings cannot be decided with one training per round.

\section{Pre-specified readouts and post hoc analyses}\label{app:prespecified}

For most experiments, the quantities to read, the statistics and what would count as support or as a contradiction were written into the experiment's protocol, or into a committed script, before the data were read. ``Pre-specified'' means this; nothing was registered on an external platform. \Cref{tab:prespecified} lists the main ones with their outcome under the stated rule. Everything else in the paper is descriptive or post hoc.

{\footnotesize
\begin{longtable}{@{}>{\raggedright\arraybackslash}p{0.2\textwidth}>{\raggedright\arraybackslash}p{0.41\textwidth}>{\raggedright\arraybackslash}p{0.32\textwidth}@{}}
\caption{Pre-specified readouts and their outcomes.}\label{tab:prespecified}\\
\toprule
Experiment & Fixed before the data were read & Outcome \\
\midrule
\endfirsthead
\multicolumn{3}{@{}l}{\emph{Table \thetable, continued}}\\
\toprule
Experiment & Fixed before the data were read & Outcome \\
\midrule
\endhead
\bottomrule
\endfoot
\multicolumn{3}{@{}l}{\emph{LLaDA}}\\*\addlinespace[2pt]
Main chains and follow-up & expected directions, without decision rules: the dependence proxy falls (random chains); Self-BLEU rises and the window entropy falls (low-confidence chains); directions of the end-token structure & as expected, except the window entropy, lowest only after round $4$; the proxy could not be judged on the chains' own states and was judged on fixed states in the follow-up (its decrease on the low-confidence states is not monotone) \\\addlinespace[2pt]
Round-1 retrainings & the retrained first rounds lie beyond the midpoint between the base model and the original first round & random remasking: all three readouts; low-confidence: pass@1 and Self-BLEU \\\addlinespace[2pt]
Second chains & after four rounds, pass@1 and Self-BLEU (low-confidence) and the changes of $\E[f]$ and of the tail fraction (random) reproduce the original chains & all reproduced \\\addlinespace[2pt]
Tokens per step, base model & the per-question speed limit in $\parPreCount$ cells with $n=2,8,16,32$; two count tests (recorded beliefs; a necessary condition for consistent conditionals); monotonicity in $n$ & $\parViolCount$ of $\parPreCount$ cells violated; beliefs pass; condition not rejected; monotonicity in $n$ only descriptive (its premise, no tails at $n=1$, fails) \\\addlinespace[2pt]
Per-question round 1 & signs of the parallel step of $m_1$, the learning-error term, the whole round and the base step at $e=64,96,128$ & $m_1$'s step within the limit; learning error and round beyond it; base step undecided \\\addlinespace[2pt]
Learning-error diagnosis and end-token padding & whether end tokens are more frequent after an earlier end token ($r$) than as the first one ($h$); in which contexts $m_1$ over-predicts end tokens; sign of the padding effect at step $250$ in two random streams & $r=\learnRHRatio\,h$ (post hoc: too weak to explain the size); over-prediction only block by block (with training-style masking the interval contains $1$; the split by visible end tokens in \cref{app:llada-learning-error} is post hoc); padding: opposite signs, no conclusion \\\addlinespace[2pt]
Fixed groupings & for each of two groupings: bands for the parallel step relative to zero and to half the step of the random chain; control conditions; bands for the ratio of the dependence proxy within a group to the base model's & no step in the bands ``stopped'' or ``reduced''; smaller than the random chain's step in $\fxSmallerCount$ of $\fxReadouts$; the two conditions on one model under two decoders fail; ratios $\fxDepG$ and $\fxDepAlt$, not specific to the grouping \\\addlinespace[2pt]
Real data in every round & after four rounds, against both random chains: the first end token is later; fewer answers have a tail; the change of $\E[f]$ from round three to four & later: met; fewer tails: not met; still shortening \\\addlinespace[4pt]
\multicolumn{3}{@{}l}{\emph{Dream}}\\\addlinespace[2pt]
Chains & five tests on the four rounds of both chains; comparisons of directions with LLaDA's chains, without expected directions & all five as predicted; $\dmCompSame$ of $\dmCompCells$ direction comparisons with LLaDA agree \\\addlinespace[2pt]
Per-question round 1 & as for LLaDA & $m_1$'s step, the whole round and the base step within the limit; learning-error term below zero at $e=96,128$, undecided at $e=64$ \\\addlinespace[4pt]
\multicolumn{3}{@{}l}{\emph{Small models}}\\\addlinespace[2pt]
Confidence rules, $\lambda=0$ & learned chains stay near a model of table errors; three tie-breaking variants on a chain with ties & drifted much further than predicted; two variants as expected, the index-tie variant did not stop \\\addlinespace[2pt]
Confidence rules, $\lambda>0$ & top-$2$, $\lambda=0.5$: four bins after four rounds, then stop or drift over rounds $5$--$8$; threshold rule: a threshold on the difference from a replay of the run's marginal errors, set during the run in place of the protocol's criterion; top-$2$, $\lambda=0.2$: bins over rounds $5$--$8$ and $9$--$12$ & top-$2$, $\lambda=0.5$: undecided at round $4$, then ``stopped'' (point estimate); threshold rule: met (low power); top-$2$, $\lambda=0.2$: undecided, then compatible with a lower balance and with a stop \\\addlinespace[2pt]
Longer training & bins for the extra dependence that the learned top-$2$ rule removes each round beyond the true rule (its sign, whether it is specific to the rule) and for its change with three times longer training & no consistent sign (weak); specific to the rule; neither bin for longer training \\\addlinespace[2pt]
Fine-tuning from a trained start & overshoot ratio at step $250$ above training from scratch in all six (arm, seed) pairs & met, with a small margin \\
\end{longtable}}

The Ising and end-token chains under state-independent schedules and the copy-plus-rare-trigger experiment were compared with predictions written before the runs, without decision rules. For the tokens-per-step experiment the script was committed after the data of questions $0$--$49$ were in the repository, so the readout uses questions $50$--$249$; the data for $n=4$ come from the main chains, so those cells are descriptive. For Dream the GPU job started before the tests were committed; results were not pushed or reported until they were.

\paragraph{Post hoc.} The main analyses added after the data were seen are marked \posthoc{} in \cref{sec:experiments}. They are: the three-term split of the copy-plus-rare-trigger experiment and the estimates of $\bar\theta$; the comparison of the dependence proxy and the window entropy of the LLaDA chain with real data with the random chain (planned, without a decision rule); the first-order ratchet (\cref{prop:ratchet}), derived after the confidence-rule runs, as a reading of their drift, and the accounting of the balance of the confidence rules at $\lambda>0$; the speed-limit cells for $n=4$, whose data predate the readout; the content terms $C_n$ for $n=2,8,16$, the description of $C_4$, and the intervals of the term $I$ in the decomposition at $n=32$; the split of LLaDA's end-token calibration by visible end tokens; the first round in tokens on the per-question samples; how often LLaDA's low-confidence chains fill the window; and, for the fixed groupings, the decomposition of the one step that is smaller than the random chain's, the decoder contrasts on the base model, the match contrast and the comparison of the learning errors.

\section{Compute and reproducibility}\label{app:compute}

All experiments ran on one NVIDIA RTX PRO 4500 Blackwell GPU (32\,GB). The recorded GPU wall-clock time is $\gpuTotal$ hours: $\gpuLLaDA$ for LLaDA (the main chains $\gpuLladaMain$, the second chains $\gpuLladaLcrep$ and $\gpuLladaRndrep$, the chain with real data $\gpuLladaLam$, the fixed groupings $\gpuLladaFixsched$ and $\gpuLladaFixalt$, and the first-round and base-model experiments), $\gpuDream$ for Dream (main chains $\gpuDreamMain$) and $\gpuSmall$ for the small models, which ran up to four training processes at once. Earlier pilot runs are not included.

Every result reported in this paper is produced by scripts from the tables of the experiments and the outputs of the theory scripts, which contain the exact enumerations (counterexamples, closed forms and perfect-learning chains). \codeplaceSentence{} the code of all experiments, the tables behind every reported number, the theory scripts with their outputs, and the scripts that generate the numbers, tables and figures of this paper; model adapters and generated text are not included because of their size. A link to the supplementary material will be added in a subsequent version of this preprint.

\end{document}